\documentclass{new_tlp}
\usepackage{mathptmx}

\usepackage{adjustbox}
  \title[Forgetting in Answer Set Programming -- A Survey]
        {Forgetting in Answer Set Programming -- A Survey}

  \author[R. Gon\c calves and M. Knorr and J. Leite]
         {RICARDO GON\c CALVES and MATTHIAS KNORR and JO\~AO LEITE\\
         NOVA LINCS, Departamento de Inform\'atica, Faculdade de Ci\^encias e Tecnologia, Universidade Nova de Lisboa, Portugal\\
         \email{{rjrg|mkn|jleite}@fct.unl.pt}}

\jdate{March 2003}
\pubyear{2003}
\pagerange{\pageref{firstpage}--\pageref{lastpage}}
\doi{S1471068401001193}

\usepackage[disable]{todonotes}

\usepackage{comment}
\DeclareMathAlphabet{\mathcal}{OMS}{cmsy}{m}{n}
\usepackage{enumitem}
\usepackage{url}

\usepackage[figuresright]{rotating}
\usepackage{lscape}
\usepackage{amsmath}
\newtheorem{theorem}{Theorem}
\newtheorem{definition}{Definition}

\newtheorem{lemma}{Lemma}
\newtheorem{corollary}{Corollary}
\newtheorem{example}{Example}

\usepackage[plain,lined,ruled,linesnumbered,noresetcount]{algorithm2e}

\usepackage{color}
\definecolor{mygreen}{rgb}{0.0, 0.6, 0.0}
\definecolor{myyellow}{rgb}{1.0, 0.65, 0.0}
\usepackage{ifthen}
\newcommand{\brifnotempty}[1]{\ifthenelse{\equal{#1}{}}{}{ \br{#1}}}
\newenvironment{lemma*}[2][]
	{\pagebreak[2] \par \noindent \textbf{Lemma~\ref{#2}\brifnotempty{#1}.}\it}{\par}
\newenvironment{theorem*}[2][]
	{\pagebreak[2] \par \noindent \textbf{Theorem~\ref{#2}\brifnotempty{#1}.}\it}{\par}
\newenvironment{proposition*}[2][]
	{\pagebreak[2] \par \noindent \textbf{Proposition~\ref{#2}\brifnotempty{#1}.}\it}{\par}
\newenvironment{corollary*}[2][]
	{\pagebreak[2] \par \noindent \textbf{Corollary~\ref{#2}\brifnotempty{#1}.}\it}{\par}

\newcommand{\lang}{\mathcal{L}}
\newcommand{\sign}{\mathcal{A}}

\newcommand{\setm}{\text{\textbackslash}}

\newcommand{\htmodels}{\models_{\sf HT}}
\newcommand{\htequiv}{\equiv_{\sf HT}}
\newcommand{\Sequiv}{\equiv_{\sf HT}}

\newcommand{\HT}{\mathcal{HT}}

\newcommand{\classP}{\mathcal{C}}
\newcommand{\la}{\leftarrow}
\newcommand{\nf}{not\,}

\newcommand{\head}[1]{\ensuremath{\mathit{A}}}
\newcommand{\pbody}[1]{\ensuremath{\mathit{B}}}
\newcommand{\nbody}[1]{\ensuremath{\mathit{C}}}
\newcommand{\nnbody}[1]{\ensuremath{\mathit{D}}}

\newcommand{\rhead}[1]{\ensuremath{\mathit{head}(#1)}}
\newcommand{\rbody}[1]{\ensuremath{\mathit{body}(#1)}}

\newcommand{\vsim}{\mathrel{\scalebox{1}[1.5]{$\shortmid$}\mkern-3.1mu\raisebox{0.15ex}{$\sim$}}}

\newcommand{\hor}{H}
\newcommand{\nor}{n}
\newcommand{\dis}{d}

\newcommand{\ex}{e}

\newcommand{\tuple}[1]{\ensuremath{\langle#1\rangle}}

\newcommand{\pE}[1]{{\bf (E$_{#1}$)}}
\newcommand{\pIR}{{\bf (IR)}}
\newcommand{\pW}{{\bf (W)}}
\newcommand{\pPP}{{\bf (PP)}}
\newcommand{\pNP}{{\bf (NP)}}
\newcommand{\pSE}{{\bf (SE)}}
\newcommand{\pCP}{{\bf (CP)}}
\newcommand{\pSP}{{\bf (SP)}}

\newcommand{\pSI}{{\bf (SI)}}
\newcommand{\pSIu}{{\bf (SI_u)}}
\newcommand{\psC}{{\bf (sC)}}
\newcommand{\pwC}{{\bf (wC)}}
\newcommand{\pwE}{{\bf (wE)}}
\newcommand{\pwSP}{{\bf (wSP)}}
\newcommand{\psSP}{{\bf (sSP)}}

\newcommand{\pUP}{{\bf (UP)}}

\newcommand{\pUI}{{\bf (UI)}}

\newcommand{\pSC}{{\bf (SC)}}
\newcommand{\pRC}{{\bf (RC)}}
\newcommand{\pNC}{{\bf (NC)}}
\newcommand{\pPI}{{\bf (PI)}}

\newcommand{\spP}{{\bf P}}

\newcommand{\spE}[1]{{\bf E$_{#1}$}}
\newcommand{\spIR}{{\bf IR}}
\newcommand{\spW}{{\bf W}}
\newcommand{\spPP}{{\bf PP}}

\newcommand{\spSE}{{\bf SE}}
\newcommand{\spCP}{{\bf CP}}
\newcommand{\spSP}{{\bf SP}}

\newcommand{\spSI}{{\bf SI}}
\newcommand{\spSIu}{{\bf SI_u}}
\newcommand{\spsC}{{\bf sC}}
\newcommand{\spwC}{{\bf wC}}
\newcommand{\spwE}{{\bf wE}}

\newcommand{\spwSP}{{\bf wSP}}
\newcommand{\spsSP}{{\bf sSP}}

\newcommand{\spUP}{{\bf UP}}

\newcommand{\spUI}{{\bf UI}}

\newcommand{\spSC}{{\bf SC}}
\newcommand{\spRC}{{\bf RC}}
\newcommand{\spNC}{{\bf NC}}
\newcommand{\spPI}{{\bf PI}}

\newcommand{\f}[2]{\ensuremath{\mathsf{f}(#1,#2)}}
\newcommand{\fgt}{\ensuremath{\mathsf{f}}}
\newcommand{\classF}{\ensuremath{\mathsf{F}}}

\newcommand{\as}[1]{\ensuremath{\mathcal{AS}(#1)}}

\newcommand{\Min}[1]{\ensuremath{\mathcal{MIN}(#1)}}

\newcommand{\SRel}{\ensuremath{\mathcal{R}_{\inst}}}

\newcommand{\Rel}{\ensuremath{Rel}}
\newcommand{\RA}{\ensuremath{R}}

\newcommand{\inst}{\ensuremath{\tuple{P,V}}}

\newcommand{\strong}{strong}
\newcommand{\weak}{weak}
\newcommand{\sem}{sem}
\newcommand{\FS}{S}
\newcommand{\FW}{W}
\newcommand{\htF}{{\sf HT}}
\newcommand{\smF}{{\sf SM}}
\newcommand{\Sas}{ Sas}
\newcommand{\SE}{\sf SE}

\newcommand{\AltNaive}{\sf R}
\newcommand{\rF}{\sf R}

\newcommand{\TAltNaive}{\sf M}
\newcommand{\mF}{\sf M}
\newcommand{\spF}{\sf SP}
\newcommand{\up}{\sf UP}

\newcommand{\Cn}{Cn}

\newcommand{\op}{{\mathsf{f}}}

\usepackage{amssymb}
\definecolor{myGray}{gray}{0.80}
\newcommand{\cM}{\checkmark}
\newcommand{\tM}{{\scalebox{0.8}{$\times$}}}

\newcommand{\coNP}{\mbox{coNP}}
\newcommand{\SigmaP}[1]{{ {\Sigma}_{#1}^{P}}}
\newcommand{\PiP}[1]{{{\Pi}_{#1}^{P}}}

\begin{document}

\label{firstpage}

\maketitle

  \begin{abstract}
Forgetting -- or variable elimination -- is an operation that allows the removal, from a knowledge base, of \emph{middle} variables no longer deemed relevant.
In recent years, many different approaches for forgetting in Answer Set Programming have been proposed, in the form of specific operators, or classes of such operators, commonly following different principles and obeying different properties.
Each such approach was developed to address some particular view on forgetting, aimed at obeying a specific set of properties deemed desirable in such view, but a comprehensive and uniform overview of all the existing operators and properties is missing. 
In this paper, we thoroughly examine existing properties and (classes of) operators for forgetting in Answer Set Programming, drawing a complete picture of the landscape of these classes of forgetting operators, which includes many novel results on relations between properties and operators, including considerations on concrete operators to compute results of forgetting and computational complexity. 
Our goal is to provide guidance to help users in choosing the operator most adequate for their application requirements.
  \end{abstract}

  \begin{keywords}
  Forgetting, Answer Set Programming, Logic Programs
  \end{keywords}


\section{Introduction}\label{sec:intro}

Forgetting is part of the human nature, often with negative connotations, and investigation for its causes started in Cognitive Psychology in the late 19th century \cite{Ebbinghaus1885}. 
Studies show that forgetting can be caused, among others, by the interference in between new and previously learned information, by decay over time, or because certain links, called cues, used to retrieve a specific memory are not (any longer) available 
\cite{Shrestha17}.
In fact, recent experiments in Neuroscience reveal that part of this forgetting appears to be done actively, in the sense that the human brain seems to be able to remove information no longer deemed necessary \cite{DavisZ17}.
This enables humans to distinguish important memories among the huge amount of memories they acquire, while abstracting irrelevant details, which allows for making better decisions under varying conditions \cite{RichardsF17}.

In organizations, \emph{intentional forgetting} plays a similar role \cite{KlugeG18}. 
As increasing amounts of information may become difficult to process, forgetting allows one to simplify the interpretation of the acquired knowledge which contributes to future success of the organization.
The methods of forgetting employed build on ideas of forgetting in Cognitive Psychology, for example, rather than unlearning organizational routines, the retrieval cues are forgotten \cite{KlugeG18}.

In the same vein, huge problems appear in Computer Science, where the vastly increasing amounts of data challenge the limits of space in terms of physical storage and processing speed, in particular in situations where the exponential worst-case complexity becomes prohibitive.
At the same time, forgetting has become increasingly important to properly deal with legal and privacy issues, such as, for example, the elimination of illegally acquired information to implement a court order, or enforcing the \emph{right to be forgotten}, i.e., the right of the eliminination of private data on a person's request, following the EU General Data Protection Regulation \cite{eu:gdpr}.

Thus, forgetting has received increasing attention in Computer Science aiming at the elimination of irrelevant information for improved decision processes and tools \cite{BeierleT19}.
In Artificial Intelligence and Knowledge Representation and Reasoning in particular, the ideas of forgetting can be traced back to Boole's variable marginalization \shortcite{Boole54}, also called the eliminination of middle terms.
These notions raised interest when Lin and Reiter \shortcite{LR94} considered forgetting about a fact or a relation in a first-order logic for Cognitive Robotics.
This subsequently triggered research into forgetting and closely related notion such as uniform interpolation \cite{Visser96}, variable elimination \cite{LLM03} or ignorance \cite{BaralZ05}, being extended for example to Description Logics \cite{GhilardiLW06,WangWTP10,LutzW11}, Planning \cite{ErdemF07}, and Modal Logic \cite{ZZ-AI09}.
Commonly two kinds of forgetting can be found in the literature \cite{EiterK19}.
One of them eliminates some formula from a given knowledge base, in a way that closely resembles the operation of contraction in belief revision.
The other, more common one, views forgetting as an operation that omits part of the vocabulary of the knowledge base, including possible adjustments on formulas to accommodate some notion of preservation of information (for the remaining vocabulary).
This is the kind of forgetting we consider here and particularly useful when we want to eliminate elements representing auxiliary concepts, with the aim to simplify a knowledge base or improve its declarativity, or related to data protection issues.
The importance of forgetting in Knowledge Representation and Reasoning is also witnessed by applications in cognitive robotics \cite{LinR97,LiuW11,RajaratnamLPT14}, for resolving conflicts \cite{LLM03,ZhangF06,EiterW08,LangM10,DelgrandeW15}, and ontology abstraction and comparison \cite{WangWTP10,KontchakovWZ10,KonevL0W12,KonevL0W13}.

While forgetting has been extensively studied in the context of classical logic \cite{BledsoeH80,LLM03,Larrosa00,LarrosaMN05,MiddeldorpOI96,Moinard07,Weber86,Gabbay:2008:SOQ:1816959}, it only recently gathered wider attention in Answer Set Programming (cf.\ the overview by Eiter and Kern-Isberner \shortcite{EiterK19}).
Answer Set Programming (ASP) \cite{Gelfond1988,GL91} provides a declarative rule-based language for knowledge representation and reasoning accompanied with efficient implementations such as CLASP \cite{GebserKKOSS11}, DLV \cite{LeonePFEGPS06,AlvianoCDFLPRVZ17}, and Smodels \cite{SimonsNS02}. 
Its non-monotonic rule-based nature required the development of specific methods and techniques -- similar to what happened with other belief change operations such as revision and update \cite{Alferes2000,EiterFST02,Sakama2003,Slota2012,Slota2012a,DelgrandeSTW13,SlotaL14}. 
The result is a significant number of different approaches on forgetting \cite{ZhangF06,EiterW08,Wong09,WangZZZ12,WangWZ13,KnorrA14,WangZZZ14,DelgrandeW15,GoncalvesKL-ECAI16,GoncalvesKLW17,GoncalvesJKLW19,BertholdGKL19,BertholdGKL19b,GoncalvesJKL21}, some presenting single operators, others semantic characterizations of classes of operators, usually with different sets of properties deemed desirable, some adapted from the literature on \emph{classical }forgetting \cite{ZZ-AI09,WangZZZ12,WangZZZ14}, others introduced for the case of ASP \cite{EiterW08,Wong09,WangZZZ12,WangWZ13,KnorrA14,DelgrandeW15,GoncalvesKLW17,GoncalvesJKLW19}, 
and often defined for different classes of answer set programs (details on the individual motivations for introducing these approaches and their distinct characteristics can be found in Section~\ref{sec:ops}). 

The result is a \emph{complex} landscape filled with classes of operators and properties, with very little effort put into drawing a map that could help us to better understand the relationships between properties and operators.  
Whereas, in principle, having an operator that obeys \emph{all} of the properties would be desirable, it turns out that any such operator cannot be defined for any class that includes the standard class of normal logic programs \cite{JWY-ijcai15}.
In fact, one of the properties arguably best captures the essence of forgetting, however, there cannot exist an operator that always satisfies it \cite{GoncalvesKL-ECAI16,GoncalvesKLW20}.
This strengthens the idea that there cannot be a one-size-fits-all forgetting operator for ASP, but rather a variety of approaches, each obeying a specific set of properties. The choice of operator will then depend on which properties are deemed more important for the specific application in hand, for which it is important to understand: a)  the relationships between different properties, b) which properties are obeyed by which (classes of) operators, and even c) whether some sets of properties make more sense than others.
To this end, we present a systematic, comprehensive and thorough survey of \emph{forgetting in ASP}, including many novel results and insights that help answering the raised questions and provide guidelines to users which operators are most suited in which applications. 

After a brief section with preliminaries and a section where the common notion of forgetting in ASP is defined, the paper is divided into three main sections. The first one contains a comprehensive account of the properties found in the literature, together with an investigation into the relationships between them, including several novel results. The subsequent section is devoted to describing the operators defined in the literature, and establishing some results on their relationships, including an equivalence between two of these operators, as well as considerations on proposed concrete operators, and the computational complexity of problems related to forgetting.  Then, we devote our final major section to present a comprehensive account of which properties are satisfied by which operators, some of the results to be found scattered in the literature, but more than half being novel. 
This section also provides a detailed discussion of the suitability of these classes with respect to their characteristics, and establishes precisely the relationship to uniform interpolation. Finally, the suitability of the classes of operators w.r.t.\ several applications considered in the context of forgetting is discussed.

This paper is a considerable extension of a previous conference publication \cite{GoncalvesKL16}.
The material has been extended to incorporate novel approaches in the literature and revised to accommodate these new results and provide a more complete picture of the existing properties and classes of forgetting operators, and their relations.
The new detailed material on concrete forgetting operators, computational complexity, the relation to uniform interpolation, and suitability of classes of operators for certain applications help shed further light on these relations, and provide more guidance for the reader. 
This is complemented with an appendix, available as supplementary material to this paper, that contains all the proofs of the novel results, as well as those in the previous publication, and pointers for all those results spread out over the literature.

\section{Answer Set Programming}\label{sec:asp}

In this section, we recall the necessary notions and notation on logic programs under the answer set semantics.

We assume a \emph{propositional signature} $\sign$, i.e., a finite set of (propositional) atoms, also termed propositional variables synonymously. 
An \emph{(extended) logic program} $P$ over $\sign$ is a finite set of \emph{(extended) rules} of the form
\begin{align}
a_1 \vee \ldots \vee a_k  \la b_1,..., b_l, \nf c_{1},..., \nf c_m, \nf \nf d_1,..., \nf \nf d_n \; , \label{l:rule}
\end{align}
where all $a_1,\ldots,a_k,b_1,\ldots, b_l,c_{1},\ldots, c_m$, and $d_{1},\ldots, d_n$ are atoms of $\sign$. 
Such rules $r$ are also commonly written in a more succinct way as 
\begin{equation}
A \la B, \nf C, \nf \nf D \; , \label{l:shortRule}
\end{equation}
where $\head{r} = \{a_1,\ldots,a_k\}$, $\pbody{r}=\{b_1,\ldots, b_l\}$, $\nbody{r}=\{c_{1},\ldots, c_m\}$, and $\nnbody{r}=\{d_{1},\ldots, d_n\}$,
and we use both forms interchangeably. 
For each rule $r$, we distinguish the \emph{head}, $\rhead{r}=A$, and its \emph{body}, $\rbody{r}=\pbody{r}\cup\nf \nbody{r}\cup \nf \nf \nnbody{r}$, where 
$\nf C$ and $\nf\nf D$ represent $\{\nf p\mid p\in C\}$ and $\{\nf\nf p\mid p\in D\}$, respectively.
We write the set of atoms appearing in $P$ as $\sign(P)$ and the class of (extended) logic programs as $\classP_{\ex}$. 
Such extended logic programs are actually equivalent to an even more expressive syntax \cite{LifschitzTT99}, but the more concise syntax used here suffices. 
Also note that double negation is commonly required in the context of forgetting in ASP and commonly supported by answer set solvers such as clingo \cite{GebserKKLOORSSW18}.

The class of extended programs includes a number of special kinds of rules $r$: if $n=0$, then we call $r$ \emph{disjunctive}; if, in addition, $k\leq 1$, then $r$ is \emph{normal}; if on top of that $m=0$, then we call $r$ \emph{Horn}; if moreover $l=0$, then we call $r$ a \emph{fact}.
We also admit \emph{constraints}, which are (extended) rules where $k=0$.
The classes of \emph{disjunctive}, \emph{normal}, and \emph{Horn programs}, $\classP_{\dis}$, $\classP_{\nor}$, and $\classP_{\hor}$,
are defined as a finite set of disjunctive, normal, and Horn rules,
respectively. 
We have $\classP_{\hor}\subset \classP_{\nor}\subset \classP_{\dis}\subset \classP_{\ex}$. 

We now define the \emph{answer sets} \cite{GL91} of a program, i.e., its models, and we recall this notion based on HT-models as defined in the context of the logic of here-and-there \cite{Heyting30}, the monotonic logic underpinning ASP \cite{Pearce99}. We start with the notion of \emph{reduct}~\cite{GL91} of a program $P$ with respect to a set $I$ of atoms:
\[
P^I = \{\head{r}\la \pbody{r} : r \text{ of the form (\ref{l:shortRule}) in } P \text{ such that } \nbody{r}\cap I=\emptyset \text{ and } \nnbody{r}\subseteq I\}.
\]

An \emph{HT-interpretation} is a pair $\langle X,Y\rangle$ s.t.\ $X\subseteq Y \subseteq \sign$.
Note that we follow a common convention and usually abbreviate sets in HT-interpretations such as $\{a,b\}$ with the sequence of its elements, $ab$.
Given a program $P$, an HT-interpretation $\langle X,Y\rangle$ is an \emph{HT-model of $P$} if $Y\models P$ and $X\models P^{Y}$, where $\models$ is the standard consequence relation for classical logic and where programs are interpreted as conjunctions of classical implications 
\[
b_1 \wedge...\wedge b_l\wedge \neg c_{1}\wedge...\wedge \neg c_m\wedge \neg\neg d_1\wedge...\wedge \neg\neg d_n \rightarrow  a_1 \vee \ldots \vee a_k  
\]
corresponding to its rules $r$ of the form~(\ref{l:rule}) where $\neg$, $\wedge$, and $\vee$ denote classical negation, conjunction and disjunction, respectively.
We occasionally admit for the sake of readability that the HT-models of a program $P$ are restricted to $\sign(P)$ even if $\sign(P)\subset \sign$.
The set of \emph{all HT-models of $P$} is written $\HT(P)$. 
Given two programs $P_1$ and $P_2$, if $\HT(P_1)\subseteq \HT(P_2)$, then $P_1$ \emph{entails} $P_2$ in HT, written $P_1\htmodels P_2$.
Also, $P_1$ and $P_2$ are \emph{HT-equivalent}, written $P_1\htequiv P_2$, if $\HT(P_1)=\HT(P_2)$.

A set of atoms $Y$ is an \emph{answer set} of $P$ if $\tuple{Y,Y}\in\HT(P)$, and there is no $X\subset Y$ such that $\tuple{X,Y}\in\HT(P)$.
We denote by $\as{P}$ the set of all answer sets of $P$.
For $\classP_{\dis}$ and its subclasses, all $I\in \as{P}$ are pairwise incomparable.
A program $P$ is \emph{consistent} if $\as{P}$ is not empty. 

\begin{example}\label{exampleAnswerSets}
Consider the following program $P$:
\begin{align*}
a &\la \nf b &  b & \la \nf c & e & \la d & d &\la a
\end{align*}
We can show that $\langle b,bde\rangle$ is an HT-model of $P$ because $\{b,d,e\}\models P$ and $\{b\}\models P^{\{b,d,e\}}$ where $P^{\{b,d,e\}}$ is as follows:
\begin{align*}
b & \la & e & \la d & d &\la a
\end{align*}
It is then easy to see that $\{b,d,e\}$ is not an answer set of $P$ (given that $\langle b,bde\rangle$ is an HT-model).
Similarly, $\{b,d\}$ is not an answer set of $P$ because $\{b,d\}\not\models P$.
In fact, we can verify that $\{b\}$ is the only answer set of $P$, and that $P$ is therefore consistent.
\end{example}

Different notions of equivalence between programs have been established \cite{EiterFW07}, essentially based on comparing their answer sets in different ways.
Namely, we say that two programs $P_1, P_2$ are \emph{equivalent}, written $P_1\equiv P_2$, if $\as{P_1}=\as{P_2}$, i.e., their answer sets coincide. 
Uniform equivalence strengthens this condition by imposing that the equality of answer sets should hold even if a set of facts is added to both programs. 
Formally, two programs $P_1, P_2$ are \emph{uniformly equivalent}, written $P_1\equiv_u P_2$, if $\as{P_1\cup R}=\as{P_2\cup R}$ for every set $R$ of facts. 
Strong equivalence further strengthens the condition by imposing equality of answer sets under the addition of any set of rules. 
Formally, two programs $P_1, P_2$ are \emph{strongly equivalent} if $\as{P_1\cup R}=\as{P_2\cup R}$ for every $R\in \classP_{\ex}$. 
It is well-known that two programs $P_1, P_2$ are strongly equivalent exactly when $P_1\htequiv P_2$~\cite{LPV01}, and we therefore use $\htequiv$ to denote both equivalences.
Relativized equivalence relaxes the condition of strong equivalence by allowing to vary the language of the additional programs. 
Formally, two programs $P_1, P_2$ are \emph{relativized equivalent} w.r.t.\ $V\subseteq \sign$, written $P_1\equiv_V P_2$, if $\as{P_1\cup R}=\as{P_2\cup R}$ for every $R\in \classP_{\ex}$ s.t.\ $\sign(R)\subseteq \sign\text{\textbackslash} V$.
Thus, strong equivalence and equivalence can be considered special cases of relativized equivalence, where the considered set $V$ is empty and identical to $\sign$, respectively.

\begin{example}
Consider $P_1=\{b\la\}$. 
Then $P_1$ and $P$ from Ex.~\ref{exampleAnswerSets} are equivalent, but neither strongly nor uniformly equivalent, e.g., because adding $R=\{c\la\}$ to both programs yields different answer sets, namely $\{b,c\}$ and $\{a,c,d,e\}$, respectively.
Also, neither is an HT-consequence of the other, since $\langle\emptyset,bc\rangle\in \HT(P)$, but not in $\HT(P_1)$, and $\langle b,bd\rangle\in\HT(P_1)$, but not in $\HT(P)$.

On the other hand, $P_2=\{a\vee b\}$ and $P_3=\{a\la\nf b; b\la \nf a\}$, are well-known to be uniformly equivalent \cite{EiterF03}, but not strongly equivalent, e.g., for $R=\{a\la b; b\la a\}$.
We have that $P_2\htmodels P_3$, but not $P_3\htmodels P_2$, because of, e.g., $\langle \emptyset,ab\rangle$.
\end{example}

Occasionally, we want to omit certain elements of the signature from sets of interpretations and HT-models.
Given a set of atoms $V$, the \emph{$V$-exclusion} of a set of answer sets (resp.\ a set of HT-interpretations) $\mathcal{M}$, written $\mathcal{M}_{\parallel V}$, is $\{X\text{\textbackslash} V\mid X\in\mathcal{M}\}$ (resp.\ $\{\tuple{X\text{\textbackslash} V,Y\text{\textbackslash} V}\mid \tuple{X,Y}\in\mathcal{M}\}$). 
Also, given two sets of atoms $X,X'\subseteq \sign$, we write $X\sim_V X'$ whenever $X\text{\textbackslash} V=X'\text{\textbackslash} V$. For $HT$-interpretations $\langle H,T\rangle$ and $\langle X,Y\rangle$, $\langle H,T\rangle\sim_V \langle X,Y\rangle$ denotes that $H\sim_V X$ and $T\sim_V Y$.
Then, for a set $\mathcal{M}$ of $HT$-interpretations, $\mathcal{M}_{\dagger V}$ denotes the set $\{\langle X,Y\rangle\mid \langle H,T\rangle\in\mathcal{M}$ and $\langle X,Y\rangle\sim_V \langle H,T\rangle\}$.

\section{Forgetting}\label{sec:forget}

In this section, we formally introduce the notion of forgetting in answer set programming.
More precisely, we define operators of forgetting and classes of these in a general way that aligns with all the different approaches presented in the literature.
We note that our account is on forgetting propositional atoms, just as all the literature on forgetting in answer set programming. 
This also allows us to capture forgetting atoms from ground programs obtained from programs built over predicate symbols, constants, and variables, simply by considering a one-to-one mapping of such ground atoms to a propositional alphabet.
To keep the exposition simple, we here consider forgetting in the propositional setting. 

The principal idea of forgetting in ASP is to remove certain atoms from a given program, while preserving its semantics for the remaining atoms.

\begin{example}\label{ex:basic}
Consider program $P$ from Ex.~\ref{exampleAnswerSets}:
\begin{align*}
a &\la \nf b &  b & \la \nf c & e & \la d & d &\la a
\end{align*}
If we want to forget about some atom, then we expect all rules that do not mention this atom to persist, while rules that mention it to no longer occur.
For example, when forgetting about $d$ from $P$, the first two rules should be contained in a result of forgetting, while the latter two should not.
At the same time, implicit dependencies, such as $e$ depending on $a$ via $d$, should be preserved.
Hence, we would expect the following result:
\begin{align*}
a &\la \nf b &  b & \la \nf c &  e &\la a
\end{align*}

In addition, if the atom to be forgotten does not appear at the same time in some rule body and some rule head, usually no dependencies need to be preserved. 
Alternatively, consider forgetting about $c$ from $P$. 
Then, since $c$ only appears negated in the body of the rule with head $b$, $c$ is false. 
Thus, $b$ becomes unconditionally true when forgetting, and the expected result would be:
\begin{align*}
a &\la \nf b &  b & \la & e & \la d & d &\la a
\end{align*}
Of course, with a fact for $b$ present, the body of the first rule can never be true and, alternatively, we may consider the following program as result of forgetting:
\begin{align*}
b & \la & e & \la d & d &\la a
\end{align*}
It can be verified that both programs are in fact strongly equivalent, i.e., both equally preserve the semantics for the remaining atoms.
\end{example}

Thus, forgetting can be viewed as returning a set of programs, which are equivalent in some way, e.g., according to one of the notions presented in the previous section, that only mention the remaining atoms and preserve the semantics of the given program over these remaining atoms.
In the literature, concrete operators have been defined that, similar to a function, provide one unique such representative for each program $P$ and set of atoms $V$ to be forgotten.
We formalize this central idea with the notion of a forgetting operator.

\begin{definition}\label{def:forgOp}
Given a class of logic programs $\classP$ over $\sign$, a \emph{forgetting operator (over $\classP$)} is defined as a function $\op:\classP\times 2^{\sign}\to \classP$ where, for each $P\in \classP$ and $V\subseteq \sign$, $\f{P}{V}$, the \emph{result of forgetting about $V$ from $P$},
\begin{itemize}
	\item is a program over $\sign(P)\text{\textbackslash} V$; and
	\item preserves the semantic relations between atoms in $\sign(P)\text{\textbackslash} V$ from $P$. 
\end{itemize} 
We denote the domain of $\fgt$ by $\classP(\fgt)$.
A forgetting operator $\op$ is called \emph{closed} for $\classP'\subseteq\classP(\fgt)$ if, for every $P\in \classP'$ and $V\subseteq \sign$, we have $\f{P}{V}\in \classP'$.
\end{definition}
\noindent

Our definition establishes that forgetting can be understood as a reduction of the language preserving the semantic relations for the remaining atoms in $P$.
The latter is in line with the argument in Example~\ref{ex:basic} and requires that these semantic relations be established based on some semantic notion such as answer sets or one of the established equivalence notions, though, for the sake of generality, no concrete notion is fixed.  
This condition allows us to exclude nonsensical functions, such as, for example simply always deleting the entire program or replacing it by arbitrary rules over the remaining atoms, but we do not specify precisely how these semantic relations are established.
In the literature, more precise notions for such semantic relations have been defined.
For example, Delgrande \shortcite{Delgrande17} defined for arbitrary logics that the result of forgetting corresponds to all the consequences of the given formulas over the remaining language.
As we will see in Sect.~\ref{sec:ops}, this aligns with some of the approaches in the literature of forgetting in ASP, but many others rather rely on for example preserving some notion of models (answer sets or HT-models) in some way, which in fact often turns out to be closely connected to the properties of forgetting in ASP we present in Sect.~\ref{sec:props}.
For this reason, we abstain here from specifying how these semantic relations are determined specifically and refer to Sect.~\ref{sec:ops} for the details for each of the existing approaches.

We point out that the notion of closed operators allows us to indicate whether an operator, when applied to a (sub-)class of programs for which it is defined, does return a program in that same class.
This is important, since based on this we are able to answer the question whether such operator can be iterated on such (sub-)class.
Naturally, by Def.~\ref{def:forgOp}, any forgetting operator is closed for the most general class for which it is defined.

It is worth noting that some notions of forgetting do not explicitly require that atoms to be forgotten be absent from the result of forgetting, but instead that they be \emph{irrelevant}, i.e., the result of forgetting is strongly equivalent to a program that does not mention the atoms to be forgotten.
In our view, this is not aligned with the conceptual idea of forgetting itself.
However, since such (irrelevant) occurrences of atoms in a result of forgetting are commonly assumed to be not occurring in the result, the required strong equivalence naturally holds. 
Hence, requiring that forgetting operators yield programs without the atoms to be forgotten does not prevent coverage of these particular approaches.

Now, as Ex.~\ref{ex:basic} indicates, preserving the semantics for the remaining atoms is not necessarily tied to one unique program.
In fact, in the literature, usually, a representative up to some notion of equivalence between programs is considered, which is also why, in this case, we often refer to \emph{a} result of forgetting (in indefinite terms) as opposed to \emph{the} result.
In this sense, many notions of forgetting for logic programs are defined semantically, i.e., they introduce a class of operators that satisfy a certain semantic characterization.
To capture this, we introduce the notion of a class of forgetting operators.

\begin{definition}\label{def:forgClass}
A \emph{class $\classF$ of forgetting operators (over $\classP$)} is a set of forgetting operators $\fgt$,  with $\classP(\fgt)\subseteq \classP$, that satisfy the semantic characterization of that class. 
\end{definition}

\noindent
In this sense, the notion of a class of operators is used as an easy way of referring to all concrete operators that satisfy its semantic characterization, which is also useful in the cases in the literature where only the semantic characterization is presented and no concrete operator is defined. 
To remain as general and uniform as possible, in this paper, we focus on classes of operators.
Whenever a notion of forgetting in the literature is defined through a concrete forgetting operator only, we consider the class containing that single operator.

The subset inclusion for the domain of operators in the previous definition is justified by the fact that there are classes of operators in the literature that include concrete operators only defined for a subclass of the programs considered by the class of operators.

Finally, with respect to uniform interpolation, we note that it is indeed closely connected to the concept of forgetting \cite{GabbayPV11}.
However, it does not exactly correspond to the notion of forgetting in ASP in the broad sense as considered here.
We will discuss this in more detail in Sect.~\ref{subsec:interpol} and establish a precise relationship after we have properly presented properties of forgetting, classes of forgetting operators and their relations.

\section{Properties of Forgetting}\label{sec:props}

In the literature of forgetting in answer set programming, commonly one central focus has been the investigation of guiding principles that would provide desirable characteristics of classes of operators of forgetting, often called properties of forgetting.
In this section, we recall these properties found in the literature and investigate existing relations between them.

In the course of this presentation, we opt for following a historical order without any considerations on their importance or preference among each other.
In terms of technical notation, unless otherwise stated, $\classF$ represents a class of forgetting operators, $\classP(\fgt)$ the class of programs over $\sign$ of a given $\fgt\in \classF$, and whenever we write that a single operator $\op$ obeys some property, we mean that the singleton class composed of that operator, $\{\op\}$, obeys such property.

The first three properties, named strengthened Consequence, weak Equivalence, and Strong Equivalence, were proposed by Eiter and Wang \shortcite{EiterW08}, though not formally introduced as such. 
The first two were in fact guiding principles for defining their notion of forgetting, and formalized later \cite{GoncalvesKL16}, while the third was frequently considered in the literature in terms of formal results, but only formalized as a property by Wang et al.~\shortcite{WangWZ13}.

Strengthened Consequence requires that the answer sets of a result of forgetting be answer sets of the original program, ignoring the atoms to be forgotten.
\begin{itemize}
\item[\psC] $\classF$ satisfies \emph{strengthened Consequence} if, for each $\fgt\in \classF$, $P\in \classP(\fgt)$ and $V\subseteq \sign$, we have $\as{\f{P}{V}}\subseteq \as{P}_{\parallel V}$.
\end{itemize}
In other words, forgetting does not admit the introduction of new answer sets, it may only remove some in the course of forgetting.

The other two properties focus on the preservation of some notion of equivalence during forgetting, i.e., if two programs are equivalent (w.r.t.\ some notion of equivalence of programs), then the respective results are as well.
\begin{itemize}
\item[\pwE] $\classF$ satisfies \emph{weak Equivalence} if, for each $\fgt\in \classF$, $P, P'\in \classP(\fgt)$ and $V\subseteq \sign$: if $P\equiv P'$, then $\f{P}{V}\equiv \f{P'}{V}$.
\item[\pSE] $\classF$ satisfies \emph{Strong Equivalence} if, for each $\fgt\in \classF$, $P, P'\in \classP(\fgt)$ and $V\subseteq \sign$: if $P\Sequiv P'$, then $\f{P}{V}\Sequiv \f{P'}{V}$.
\end{itemize}
Weak Equivalence and Strong Equivalence require that forgetting preserves equivalence and strong equivalence of programs, respectively.
For other notions of equivalence, no corresponding property has been considered in the literature.

The next four properties, called Irrelevance, Weakening, Positive Persistence, and Negative Persistence, were introduced by Zhang and Zhou \shortcite{ZZ-AI09} as postulates of knowledge forgetting in the context of modal logics, and later adopted by Wang et al.~\shortcite{WangZZZ12,WangZZZ14} for forgetting in ASP.

Irrelevance requires that a result of forgetting be strongly equivalent to a program that does not mention the atoms to be forgotten.

\begin{itemize}
 \item[\pIR] $\classF$ satisfies \emph{Irrelevance} if, for each $\fgt\in \classF$, $P\in \classP(\fgt)$ and $V\subseteq \sign$, $\f{P}{V}\Sequiv P'$ for some $P'$ not containing any $v\in V$.
\end{itemize}
This property corresponds to the concept of being irrelevant discussed in the end of the previous section. Thus, satisfaction of this property is an integral part of the definitions of forgetting operators and classes (cf.\ Defs.~\ref{def:forgOp} and \ref{def:forgClass}).

The other three properties focus on HT-consequences.
Namely, Weakening requires that the HT-models of $P$ also be HT-models of a result of forgetting.

\begin{itemize}
\item[\pW] $\classF$ satisfies \emph{Weakening} if, for each $\fgt\in \classF$, $P\in \classP(\fgt)$ and $V\subseteq \sign$, we have $P\htmodels\f{P}{V}$. 
\end{itemize}
This means that a result of forgetting, $\f{P}{V}$, has at most the same consequences as the program $P$ itself.

Positive and negative persistence concern preserving the HT-consequences of $P$ and not introducing new ones, respectively.

\begin{itemize}
\item[\pPP] $\classF$ satisfies \emph{Positive Persistence} if, for each $\fgt\in \classF$, $P\in \classP(\fgt)$ and $V\subseteq \sign$: if $P\htmodels P'$, with $P'\in \classP(\fgt)$ and $\sign(P')\subseteq \sign\setm V$, then $\f{P}{V}\htmodels P'$. 
\item[\pNP] $\classF$ satisfies \emph{Negative Persistence} if, for each $\fgt\in \classF$, $P\in \classP(\fgt)$ and $V\subseteq \sign$: if $P\not\htmodels P'$, with $P'\in \classP(\fgt)$ and $\sign(P')\subseteq \sign\setminus V$, then $\f{P}{V}\not\htmodels P'$.
\end{itemize}
Thus, Positive Persistence requires that the HT-consequences of $P$ not containing atoms to be forgotten be preserved in a result of forgetting, while Negative Persistence requires that a program not containing atoms to be forgotten not be an HT-consequence of $\f{P}{V}$, unless it was already a HT-consequence of $P$.

Essentially in parallel to the appearance of the previous set of properties, Wong introduced a set of properties in his PhD dissertation~\shortcite{Wong09}.
They were defined for forgetting a single atom from a given disjunctive program, and did not gather much attention in the literature, possibly due to the form of publication.
These properties were generalized to extended programs and to forgetting sets of atoms and assigned a more descriptive name (as Wong simply used alphanumeric identifiers) \cite{GoncalvesKL16,GoncalvesKLJELIA16,GKL17}.
In the course of this generalization, it turned out that two of the resulting properties would precisely coincide with \pSE\ and \pPP.
Thus, in the following, we will present those generalizations of Wong's properties that are distinct from the previous ones, using the descriptive names for the ease of readability.
Namely, we recall, Strong Invariance, Strong Consequence, Rule Consequence, Non-contradictory Consequence and Permutation Invariance.

Strong Invariance requires that it be (strongly) equivalent to add a program without the atoms to be forgotten before or after forgetting.

\begin{itemize}
 \item[\pSI] $\classF$ satisfies \emph{Strong (addition) Invariance} if, for each $\fgt\in \classF$, $P\in \classP(\fgt)$ and $V\subseteq \sign$, we have $\f{P}{V}\cup R \Sequiv \f{P\cup R}{V}$ for all programs $R\in \classP(\fgt)$ with $\sign(R)\subseteq \sign\setm V$.
\end{itemize}
In particular, this means that when computing a forgetting result using an operator from a class that satisfies this property, we can ignore the rules that do not mention the atoms to be forgotten while forgetting, and only add them in the end to this result.

The next three properties are related to HT-consequences. 
Namely, Strong Consequence requires that HT-consequences be preserved when forgetting.
\begin{itemize}
\item[\pSC] $\classF$ satisfies \emph{Strong Consequence} if, for each $\fgt\in \classF$, $P, P'\in \classP(\fgt)$ and $V\subseteq \sign$,  if $P\htmodels P'$, then $\f{P}{V}\htmodels \f{P'}{V}$.
\end{itemize}
This is similar in spirit to properties \pwE\ and \pSE, only here the HT-consequence is preserved while forgetting.

Rule Consequence requires that any rule which is a consequence of a result of forgetting about $V$ from $P$ be a consequence of a result of forgetting about $V$ from a single rule among the HT-consequences of $P$.
\begin{itemize}
\item[\pRC] $\classF$ satisfies \emph{Rule Consequence} if, for each $\fgt\in \classF$, $P\in \classP(\fgt)$ and $V\subseteq \sign$, if $\f{P}{V}\htmodels r$, then $\f{\{r'\}}{V}\htmodels r$ for some rule $r'$ such that $\{r'\}\in \classP(\fgt)$, and $P\htmodels r'$. 
\end{itemize}
In particular, since any rule $r$ in a result of forgetting is an HT-consequence of it, for each such rule, there is a rule $r'$ in the original program that gives rise to $r$ (in terms of HT-consequence).

Non-contradictory Consequence requires that whenever a rule $r$ is an HT-consequence of a forgetting result $\f{P}{V}$, then the rule obtained by adding the default negation of the atoms of $V$ to the body of $r$ should be an HT-consequence of $P$.
\begin{itemize}
\item[\pNC] $\classF$ satisfies \emph{Non-contradictory Consequence} if, for each $\fgt\in \classF$, $P\in \classP(\fgt)$ and $V\subseteq \sign$, if $\f{P}{V}\htmodels \head{r}\la\pbody{r}\cup\nf \nbody{r}\cup \nf \nf \nnbody{r}$, then $P\htmodels \head{r}\la\pbody{r}\cup\nf \nbody{r}\cup \nf V\cup \nf \nf \nnbody{r}$.
\end{itemize}
Again, since any rule $r$ in a result of forgetting is an HT-consequence of it, the original program $P$ has such a non-contradictory HT-consequence.

The final property by Wong, Permutation Invariance, requires that the order not be relevant when sequentially forgetting atoms.

\begin{itemize}
\item[\pPI] $\classF$ satisfies \emph{Permutation Invariance} if, for each $\fgt\in \classF$, $P\in \classP(\fgt)$, and $V\subseteq \sign$, we have that  $\f{P}{V}\Sequiv \f{...\f{P}{V_{1}}}{..., V_{n}}$ for every partition $\{V_{1},...,V_{n}\}$ of $V$.
\end{itemize}
In fact, such iteration of forgetting had been considered before in a similar manner in a result by Eiter and Wang \citeyear{EiterW08}, but based on an already given sequence of atoms to be forgotten instead of a set.
Thus, \pPI\ is slightly more general \cite{GKL17}.
Also, independently, a variant of \pPI\ was introduced by Wang et al.~\citeyear{WangWZ13}, but shown to be equivalent \cite{GKL17}.
Therefore, in the following, we only use \pPI\ as a representative of these properties of invariance for different orders of forgetting a set of atoms.

The next property, called Existence, was first discussed by Wang et al.~\shortcite{WangZZZ12} and formalized by Wang et al.~\shortcite{WangWZ13}.
It requires that a result of forgetting for $P$ in $\classP$ be again in the class $\classP$.
This is important to determine if class of forgetting operators is suitably well-defined for the class of programs it is intended for, as well for iteration on subclasses of programs for which it is defined.
We follow the notation introduced in \cite{GoncalvesKL16} which formalizes this property s.t.\ it be explicitly tied to a class $\classP$, thus allowing to speak about a class of forgetting operators $\classF$ being closed for different classes $\classP$. 
\begin{itemize}
\item[\pE{\classP}] $\classF$ satisfies \emph{Existence for $\classP$}, i.e., $\classF$ is \emph{closed for a class of programs $\classP$} if there exists $\op\in \classF$ s.t.\ $\op$ is closed for $\classP$.
\end{itemize}
Thus, class $\classF$ being closed for some $\classP$ requires that there exist some ``witness in favor of it''.
This also means that if a class $\classF$ of operators does not satisfy this property for some class $\classP$ of programs, no operator of $\classF$ can be found that is closed for that class of programs.

The next property, called Consequence Persistence, was introduced by Wang et al.~\shortcite{WangWZ13} building on the ideas behind \psC\ by Eiter and Wang \shortcite{EiterW08}.
Consequence persistence requires that the answer sets of a result of forgetting correspond exactly to the answer sets of the original program, ignoring the atoms to be forgotten.

\begin{itemize}
\item[\pCP] $\classF$ satisfies \emph{Consequence Persistence} if, for each $\fgt\in \classF$, $P\in \classP(\fgt)$ and $V\subseteq \sign$, we have $\as{\f{P}{V}}=\as{P}_{\parallel V}$.
\end{itemize}
In other words, forgetting cannot introduce new answer sets nor remove existing ones.

The following property, called Strong Persistence, was introduced by Knorr and Alferes \shortcite{KnorrA14} with the aim of imposing the preservation of all dependencies contained in the original program building on ideas of strong equivalence between the original program and a result of forgetting (modulo the atoms to be forgotten).

\begin{itemize}
\item[\pSP] $\classF$ satisfies \emph{Strong Persistence} if, for each $\fgt\in \classF$, $P\in \classP(\fgt)$ and $V\subseteq \sign$, we have $\as{\f{P}{V}\cup R}=\as{P\cup R}_{\parallel V}$, for all programs $R\in \classP(\fgt)$ with $\sign(R)\subseteq \sign\setm V$. 
\end{itemize}
This strengthens \pCP\ considerably by imposing that the correspondence between answer sets of the result of forgetting and those of the original program be preserved in the presence of any additional set of rules not containing the atoms to be forgotten.

A further property, Weakened Consequence, is due to results by Delgrande and Wang \shortcite{DelgrandeW15}.
It requires that the answer sets of the original program be preserved while forgetting, ignoring the atoms to be forgotten.

\begin{itemize}
\item[\pwC] $\classF$ satisfies \emph{weakened Consequence} if, for each $\fgt\in \classF$, $P\in \classP(\fgt)$ and $V\subseteq \sign$, we have $\as{P}_{\parallel V}\subseteq \as{\f{P}{V}}$. 
\end{itemize}
This property can thus also be understood as being the counterpart to \psC: one does prevent the introduction of new answer sets, the other the loss of existing ones.
One can also observe that they correspond to the two inclusions of \pCP.

The following two properties, weakened and strengthened Strong Persistence, are similiar in spirit, as they are generalizations of \pwC\ and \psC\ that correspond to the two inclusions of \pSP\ \cite{GoncalvesKLW17}.

\begin{itemize}
\item[\pwSP] $\classF$ satisfies \emph{weakened Strong Persistence} if, for each $\fgt\in \classF$, $P\in \classP(\fgt)$ and $V\subseteq \sign$, we have $\as{P\cup R}_{\parallel V}\subseteq \as{\f{P}{V}\cup R}$, for all $R\in \classP(\fgt)$ with $\sign(R)\subseteq \sign\text{\textbackslash} V$.
\item[\psSP] $\classF$ satisfies \emph{strengthened Strong Persistence} if, for each $\fgt\in \classF$, $P\in \classP(\fgt)$ and $V\subseteq \sign$, we have $\as{\f{P}{V}\cup R}\subseteq\as{P\cup R}_{\parallel V}$, for all $R\in \classP(\fgt)$ with $\sign(R)\subseteq \sign\text{\textbackslash} V$.
\end{itemize}

Weakened Strong Persistence guarantees that all answer sets of $P$ are preserved when forgetting, no matter which rules $R$ over $\sign\text{\textbackslash} V$ are added to $P$, while strengthened Strong Persistence ensures that all answer sets of a result of forgetting indeed correspond to answer sets of $P$, independently of the added set of rules $R$.

Finally, Uniform Persistence was introduced by Gon\c{c}alves et al.~\shortcite{GoncalvesJKLW19} 
in the context of forgetting in modular answer set programming.
Uniform Persistence requires that the correspondence between answer sets of a result of forgetting and those of the original program be preserved in the presence of any additional set of facts not containing the atoms to be forgotten.

\begin{itemize}
\item[\pUP] $\classF$ satisfies \emph{Uniform Persistence} if, for each $\fgt\in \classF$, $P\in \classP(\fgt)$ and $V\subseteq \sign$, we have $\as{\f{P}{V}\cup R}=\as{P\cup R}_{\parallel V}$, for all sets of facts $R$ with $\sign(R)\subseteq \sign\setm V$. 
\end{itemize}
Hence, \pUP\ can be seen as a variant of \pSP\ under uniform equivalence.
Since this is the only property in the literature related to uniform equivalence, we complement with one further property to complete the picture in that regard.
It is a variant of a property already presented, namely \pSI, but directed towards uniform equivalence. 

\begin{itemize}
 \item[\pUI] $\classF$ satisfies \emph{Uniform (addition) Invariance} if, for each $\fgt\in \classF$, $P\in \classP(\fgt)$ and $V\subseteq \sign$, we have $\f{P}{V}\cup R \Sequiv \f{P\cup R}{V}$ for all sets of facts $R$ with $\sign(R)\subseteq \sign\setm V$.
\end{itemize}

Thus, Uniform Invariance requires that it be (strongly) equivalent to add a set of facts without the atoms to be forgotten before or after forgetting.

Recently, a different relaxation has of \pSI\ has been introduced by Gon\c calves et al. \shortcite{GoncalvesJKL21} when investigating syntactic operators under uniform equivalence.

\begin{itemize}[align=left]
	\item[$\pSIu$] $\classF$ satisfies \emph{Strong Invariance with respect to uniform equivalence} if, for each $\fgt\in \classF$, $P\in \classP(\fgt)$ and $V\subseteq \sign$, we have $\f{P}{V}\cup R\equiv_u \f{P\cup R}{V}$, for all programs $R\in \classP(\fgt)$ with $\sign(R)\subseteq \sign\setm V$. 
\end{itemize}

This property relaxes strong invariance by allowing that rules not mentioning the atoms to be forgotten can be ignored and be added to the result still preserving uniform equivalence (and not strong equivalence as \pSI).
The difference to \pUI\ is that \pUI\ allows us to ignore facts over the remaining language while forgetting (under strong equivalence) whereas $\pSIu$ allows us to ignore rules over the remaining language under uniform equivalence.
The latter is arguably more useful when forgetting from a program that contains general rules not mentioning the atoms to be forgotten. 

All these properties are not orthogonal to one another, and in what follows we join the results established in the literature on the relations that exist between them.
In particular, we opt for presenting these results in a concise way, trying to avoid repetition of results that are implicitly obtained from others.

To ease the reading, for a property {\bf (P)}, we represent with ``{\bf (P)}'' that ``$\classF$ satisfies {\bf (P)}''.

\begin{theorem}\label{prop:relations}
The following relations hold for all $\classF$:
 
\begin{enumerate}

\item \pW\ is equivalent to \pNP;\label{i3}

\item \pSP\ implies \pSE;\label{i5}

\item \pCP\ and \pSI\ together are equivalent to \pSP; \label{i7}

\item  \psC\ and \pwC\ together are equivalent to \pCP;\label{i8}

\item \pCP\ implies \pwE;\label{i9}

\item \pSE\ and \pSI\ together imply \pPP;\label{i10}

\item \pwSP\ and \psSP\ together are equivalent to \pSP;\label{i16}

\item \psC\ and \pSI\ together imply \psSP;\label{i17}

\item \pwC\ and \pSI\ together imply \pwSP;\label{i18}

\item  \pW\ and \pPP\ together imply \pSC;\label{i11}

\item \pSC\ implies \pSE;\label{i12}

\item \pW\ implies \pNC;\label{i13}

\item \pwC\ is incompatible with \pW\ for $\classF$ over $\classP$ such that $\classP_{\nor}\subseteq \classP$;\label{i24}

\item \pwC\ and \pUI\ together are incompatible with \pRC\ for $\classF$ over $\classP$ such that $\classP_{\nor}\subseteq \classP$.\label{i25}

\item \pSI\ implies \pUI;\label{i22}

\item \pCP\ and \pUI\ together are equivalent to \pUP;\label{i23}

\item \pSI\ implies $\pSIu$;\label{i27}

\item \pUP\ is incompatible with $\pSIu$.\label{i26}

\end{enumerate}

\end{theorem}

Note first, that \ref{i3}., as proven in by Ji et al.\shortcite{JWY-ijcai15}, also relies on \pIR\ in its original formulation, in particular, \pW\ is equivalent to \pNP\ and \pIR.
However, as \pIR\ is an integral part of our definition of forgetting operators, this reliance is ensured implicitly. 
This means that, by \ref{i3}., the four properties, originally proposed by Zhang and Zhou \shortcite{ZZ-AI09}, in the context of forgetting in ASP actually reduce to two distinct ones, namely \pW\ and \pPP.

The following results, \ref{i5}.--\ref{i18}., show that \pSP\ is an expressive property, where \ref{i5}.\ was shown by Knorr and Alferes \shortcite{KnorrA14} \ref{i7}.--\ref{i10}.\ by Gon\c calves et al.~ \shortcite{GoncalvesKL16}, and \ref{i16}.--\ref{i18}.\ are new.
In fact, \ref{i7}. provides a non-trivial decomposition of \pSP\ into \pSI\ and \pCP.
These two are themselves expressive, as witnessed by other results.
Namely, \ref{i8}.\ shows that \pCP\ in turn is the combination of \psC\ and \pwC, and \ref{i9}.\ that it implies preservation of equivalence, while \ref{i10}.\ provides the non-trivial result that Strong Equivalence and Strong Invariance imply Positive Persistence.
Then, \ref{i16}.\ provides another decomposition of \pSP\ into \pwSP\ and \psSP\ which in turn, by \ref{i17}.\ and \ref{i18}., are implied by properties used in the decompositions \ref{i7}.\ and \ref{i8}., respectively.
Thus, \pSP\ implies \pSE, \pCP, \pSI, \psC, \pwC, \pwE, \pPP, \pwSP\, and \psSP, where the result for \pPP\ has been shown directly Ji et al.~\shortcite{JWY-ijcai15}.

The next three results, \ref{i11}.-\ref{i13}., were shown by Gon\c calves et al.~\shortcite{GKL17} and clarify the relation between the properties originally proposed by Zhang and Zhou \shortcite{ZZ-AI09} and those by Wong \shortcite{Wong09}.
Namely, the two distinct properties \pW\ and \pPP\ introduced by Zhang and Zhou~\shortcite{ZZ-AI09} together imply \pSC\ which in turn implies \pSE, where the combination of these two results, i.e., that \pW\ and \pPP\ together imply \pSE, has been shown directly by Gon\c calves et al.~\shortcite{GoncalvesKL16}.
In addition, \pW\ alone implies \pNC, which indicates that among Wong's properties, \pSC\ is stronger than \pNC.

The following two results establish incompatibility results for classes of operators defined over (at least) normal programs.
Both results are new, though \ref{i24}.\ is a revised result by Wang et al.~\shortcite{WangWZ13} where incompatibility of \pW\ with \pCP\ is established.
In this sense, this new result makes the incompatibility more precise.
In comparison to \ref{i25}., we can also observe that, though no formal relation exists between \pW\ and \pRC, \pW\ is stronger than \pRC\ in the sense that it is incompatible with \pwC, whereas, in the case of \pRC, an additional property is necessary, namely \pUI, to establish incompatibility with \pwC. 

The two new results \ref{i22}.--\ref{i23}., establish relations between the new properties introduced here w.r.t.\ uniform equivalence and their correspondents based on strong equivalence.
In fact, \ref{i22}.\ establishes that Strong Invariance implies Uniform Invariance.
The second result provides a decomposition of \pUP, and it is interesting as it, together with \ref{i7}., allows us to trace the difference between \pUP\ and \pSP\ to the different considered form of invariance. 
This strengthens previous results by Gon\c calves et al.~\shortcite{GoncalvesJKLW19} that positioned \pUP\ in between \pCP\ and \pSP.
However, unlike \pSP, \pUP\ is incompatible with any property that allows to ignore the remaining arbitrary rules (beyond facts).
This is naturally the case for \pSI, as \pSI\ relies on strong equivalence whereas \pUP\ relies on uniform equivalence, but, by \ref{i26}., even the relaxation of \pSI\ to uniform equivalence, $\pSIu$ (\ref{i27}.), is incompatible with \pUP\ \cite{GoncalvesJKL21}.

\begin{figure}
  \includegraphics[width=0.7\linewidth]{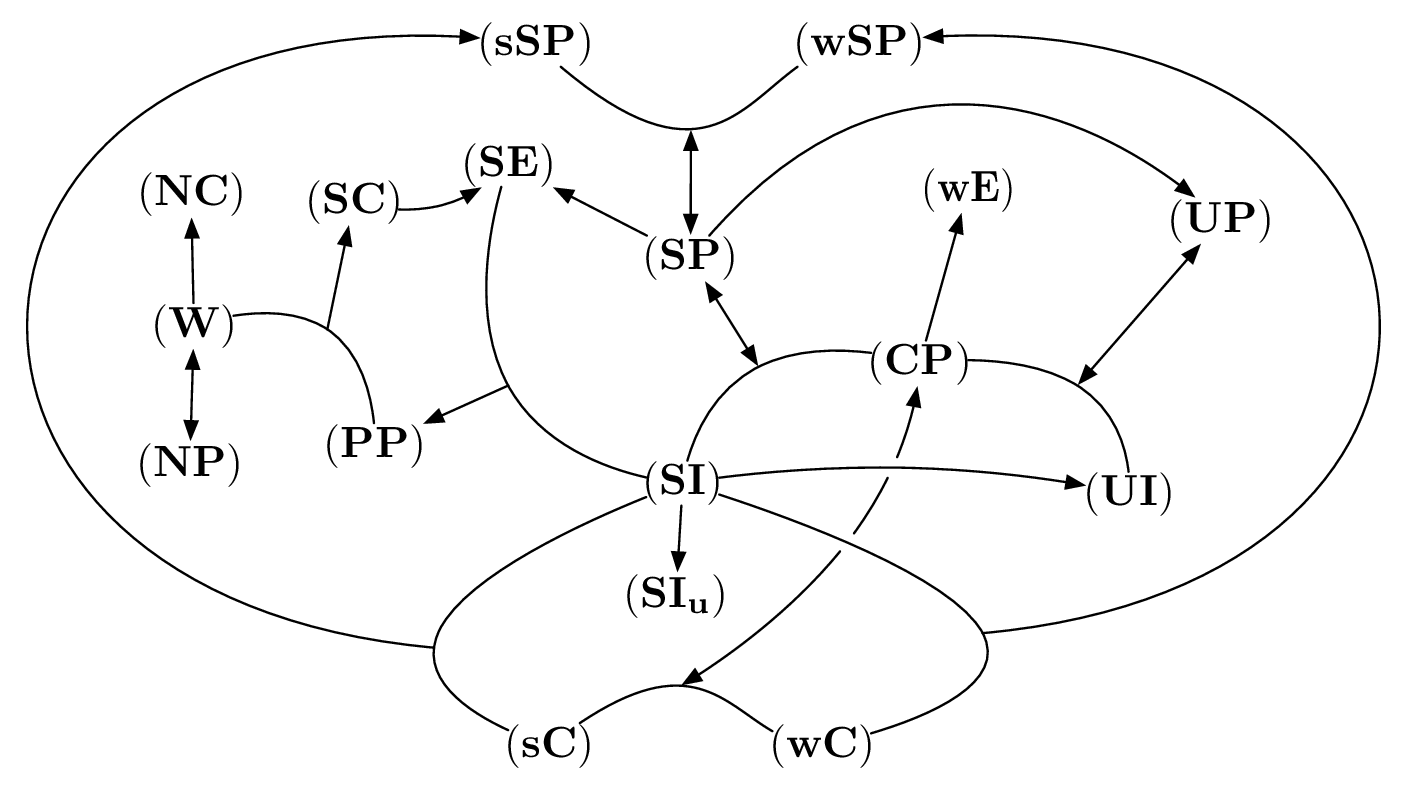}
  \caption{Relations between properties according to Thm.\ref{prop:relations}: curved lines without arrows join properties and arcs indicate directions of implication.}
  \label{fig:thm1}
\end{figure}

To gain a better overview on the results in Thm.~\ref{prop:relations}, Fig.~\ref{fig:thm1} summarizes the positive results (all but the incompatibility results) in graphical form, representing equivalences and implications between (compositions of) properties.
We can observe that \pSP\ is indeed important as the vast majority of properties is implied by it, further strengthening our view that this property is central as it arguably best captures preserving all the relations between the remaining atoms.
Though \pSP\ can in general not be satisfied for classes of programs containing normal programs \cite{GoncalvesKL-ECAI16,GoncalvesKLW20}, in our view, it ideally corresponds to what forgetting in ASP should amount to whenever possible. 
Now, among the pairs of properties equivalent to \pSP, \pSI\ and \pCP\ are arguably more important: the former allows one to focus on the rules containing the atoms to be forgotten when forgetting, which is beneficial for computation and amenable to syntactic forgetting, and the latter captures (the baseline in comparison to \pSP) that the answer sets be preserved.
While \pSP\ is in general not satisfiable, its relaxation to uniform equivalence, \pUP, is \cite{GoncalvesJKLW19}.
In fact, \pUP\ is the strongest relaxation of \pSP\ w.r.t. programs $R$ such that there is a forgetting operator over a class including normal programs that satisfies it \cite{GoncalvesJKL21}.
This makes this property an important alternative, well-suited in the setting of ASP, where problem solutions are often encoded as rules with varying sets of facts.
There are also two properties not present in Fig.~\ref{fig:thm1} that we deem important, \pPI\ and \pE{C}.
The former allows forgetting atoms in any order, which facilitates the usage of forgetting and its implementation in concrete operators that may be defined forgetting one atom at a time.
The latter is crucial for guaranteeing that operators (and classes of these) are indeed well-defined as well as permitting the iteration of these.
Finally, we note that among properties that are variants of each other with respect to equivalence and strong equivalence, such as \pwE\ and \pSE, we consider those based on strong equivalence more important as it is well-known that equivalence does not preserve the structure of rules, and this applies also in the context of forgetting.
We revisit these observations in more detail taking into consideration also the existing approaches in the literature and which properties these satisfy (as presented in the following sections). 

\section{Operators of Forgetting}\label{sec:ops}

The different properties presented in the literature have often been the driving motivation for the definition of a variety of different approaches on forgetting in answer set programming.
We now turn our attention to these classes of operators of forgetting, by first reviewing the approaches found in the literature and  then establishing non-trivial relations between them.
In the course of the exhibition, we follow a chronological order, similar in spirit to the presentation of the properties in the previous section.

\paragraph{Strong and Weak Forgetting}
The first proposals are due to Zhang et al.~\shortcite{ZhangF05,ZhangFW05,ZhangF06} with the objective to apply forgetting for conflict resolution in the case of inconsistencies as well as to capture and characterize updates of answer set programs.
The authors introduced two syntactic operators for normal logic programs, termed Strong and Weak Forgetting.
Both start with computing a reduction corresponding to the well-known weak partial evaluation (WGPPE) \cite{BrassD99}, defined as follows: for a normal logic program $P$ and $a\in\sign$, $R(P,a)$ is the set of all rules of the form $\rhead{r_1}\la \rbody{r_1}\setminus\{a\}\cup \rbody{r_2}$ such that there are $r_1,r_2\in P$ with $a\in\rbody{r_1}$ and $\rhead{r_2}=a$.
Then, the two operators differ on how they subsequently remove rules containing $a$, the atom to be forgotten.
In Strong Forgetting, all rules containing $a$ are simply removed:
\[
\fgt_{\strong}(P,a) = \{r\in R(P,a)\mid a\not\in \sign(r)\}
\] 
In Weak Forgetting, rules with occurrences of $\nf a$ in the body are kept, after $\nf a$ is removed.
\begin{align*}
\fgt_{\weak}(P,a) & = \{\rhead{r}\la\rbody{r}\setm\{\nf a\}\mid r\in R(P,a), a\not\in \rhead{r}\cup\rbody{r}\}
\end{align*}
The motivation for this difference is whether such $\nf a$ is seen as support for the rule head (Strong) or not (Weak).
In both cases, the actual operator for a set of atoms $V$ is defined by the sequential application of the respective operator to each $a\in V$. 
Both operators are shown to be closed for $\classP_{\nor}$ and thus well-defined. 
The corresponding singleton classes are defined as follows.
\[
\classF_{\strong}=\{\fgt_{\strong}\} \qquad \classF_{\weak}=\{\fgt_{\weak}\}
\]
No semantic characterization of the operators was provided, but a simplified representative was created which has the same answer sets as the corresponding forgetting results.
The authors also introduced a framework for conflict-solving in answer set programs based on strong and weak forgetting and showed that different approaches of logic program updates can be represented. 
It was also shown that forgetting does not increase the complexity of determining inferences of logic programs under answer set semantics.

\paragraph{Semantic Forgetting}
Building on ideas first exposed by Wang et al \shortcite{WangSS05}, Eiter and Wang \shortcite{EiterW08} proposed Semantic Forgetting to improve on some of the shortcomings of the two purely syntax-based operators $\fgt_{\strong}$ and $\fgt_{\weak}$.
Semantic Forgetting introduces a class of operators for consistent disjunctive programs\footnote{Actually, classical negation can occur in scope of $\nf$, but due to the restriction to consistent programs, this difference is of no effect \cite{GL91}, so we ignore it here.} defined as follows:
\[
\classF_{\sem} = \{\fgt\mid \as{\f{P}{V}} = \Min{\as{P}_{\parallel V}} \text{ for all } P\in \classP(\fgt) \text{ and } V\subseteq\sign\}
\]
The basic idea is to characterize a result of forgetting just by its answer sets, obtained by considering only the minimal sets among the answer sets of $P$ ignoring $V$.
Thus, preserving the semantic relations between the remaining atoms in the sense of Def.~\ref{def:forgOp} is based on preserving answer sets, i.e., certain atoms occur in the same answer sets. 
The authors showed that their approach can be characterized by forgetting in classical logic (using the minimal models).
Three concrete algorithms were presented, two based on semantic considerations and one syntactic.
Unlike the former two, the latter is not closed for classes $\classP_{\dis}^+$ and $\classP_{\nor}^+$ (where the superscript $^+$ denotes the restriction to consistent programs), since double negation is required in general.
Hence, it is not a forgetting operator according to our definition.
A detailed complexity analysis was provided discussing model checking as well as credulous and skeptical reasoning under forgetting. 
The authors also presented a framework for resolving conflicts in multi-agent systems which is similar in spirit to that defined by Zhang and Foo \shortcite{ZhangF06} though arguably more general in the sense that it can be adapted more easily than the former. The authors also characterized inheritance logic programs \cite{BuccafurriFL02} and update logic programs \cite{EiterFST02} using their forgetting approach.

\paragraph{Semantic Strong and Weak Forgetting}
Wong \shortcite{Wong09} argued that forgetting in ASP should be characterized by a set of properties, similar as proposed by Eiter and Wang \shortcite{EiterW08}, but rather rely on strong equivalence, as answer sets alone do not contain all the information present in a program nor do they preserve all the information that is unrelated to the forgotten atoms.
He defined two classes of forgetting operators for disjunctive programs, building on HT-models.\footnote{Wong \shortcite{Wong09} considered SE-models \cite{Turner03}. Without loss of generality, we consider the more general HT-models.}
First, given a program $P$,  we define $\Cn(P)=\{r\mid r \text{ disjunctive}, P\htmodels r$, $\sign(r)\subseteq \sign(P)\}$, the set of all consequences of $P$.
We obtain $P_\FS(P,a)$ and $P_\FW(P,a)$, the results of strongly and weakly forgetting a single atom $a$ from $P$, as follows:
\begin{enumerate}
\item Consider $P_1=\Cn(P)$.
\item Obtain $P_2$ by removing from $P_1$: (i) $r$ with $a\in \rbody{r}$, (ii) $a$ from the head of each $r$ with $\nf a\in \rbody{r}$.
\item Given $P_2$, obtain  $P_\FS(P,a)$ and $P_\FW(P,a)$ by transforming certain rules $r$ in $P_2$ as follows:
\begin{center}
\begin{tabular}{c|c|c}
& $r$ with $\nf a$ in body & $r$ with $a$ in head\\
\cline{1-3}
$\FS$ & (remove)& (remove)\\
$\FW$ & remove only $\nf a$& remove only $a$ \\
\end{tabular}
\end{center}
\end{enumerate}
The generalization to sets of atoms $V$, i.e., $P_\FS(P,V)$ and $P_\FW(P,V)$, can be obtained by simply sequentially forgetting each $a\in V$, yielding the following classes of operators.
\begin{align*}
\classF_{\FS} & = \{\fgt\mid \f{P}{V}\htequiv P_\FS(P,V) \text{ for all } P\in \classP(\fgt) \text{ and } V\subseteq\sign\}\\
\classF_{\FW} & = \{\fgt\mid \f{P}{V}\htequiv P_\FW(P,V) \text{ for all } P\in \classP(\fgt) \text{ and } V\subseteq\sign\}
\end{align*}
While steps 2. and 3. are syntactic, different strongly equivalent representations of $\Cn(P)$ exist, thus providing different instances.
Wong \shortcite{Wong09} defined one construction based on inference rules for HT-equivalence, closed for $\classP_{\dis}$.
He also introduced T-equivalence which can be characterized as a weaker form of equivalence by considering only a subset of the inference rules for HT-equivalence, and showed that these correspond to strong and weak forgetting.
Thus, $\classF_{\FS}$ and $\classF_{\FW}$ as well as strong and weak forgetting rely on preserving the consequences from $P$ over the remaining atoms, in the spirit of Delgrande's general approach \shortcite{Delgrande17}, though for a varying notion of equivalence/consequence.
Finally, computational complexity of $\classF_{\FS}$ and $\classF_{\FW}$ was not considered by Wong.

\paragraph{HT-Forgetting}
Wang et al.\ \shortcite{WangZZZ12,WangZZZ14} introduced HT-Forgetting, also termed Knowledge Forgetting \cite{WangZZZ14},  building on work by Zhang and Zhou \shortcite{ZZ-AI09} in the context of modal logics that proposed certain desirable characteristics of forgetting, which were shown to precisely characterize forgetting in classical propositional logic and modal logic S5.
The authors showed that no existing notion would correspond to these ideas, and proposed  
HT-Forgetting which was defined for extended programs and used representations of sets of HT-models directly.
\[
\classF_{\htF} = \{\fgt\mid \HT(\f{P}{V}) = \HT(P)_{\dagger V} \text{ for all } P\in \classP(\fgt) \text{ and } V\subseteq\sign\}
\]
Thus, in this case, the semantic relations between the remaining atoms are preserved based on the equivalence of HT-models.
A concrete operator was presented \cite{WangZZZ14} that was shown to be closed for $\classP_{\ex}$ and $\classP_{\hor}$, and it was also shown that no operator exists that is closed for either $\classP_{\dis}$ or $\classP_{\nor}$.
In addition, FLP-forgetting \cite{WangZZZ14} was considered under the FLP-stable model semantics \cite{Truszczynski10}, but as this semantics differs from the answer set semantics and the results are essentially identical for both variants considered, we only focus on the answer set semantics.
The authors also established results on computational complexity and discussed conflict solving using HT-forgetting based on a framework similar to that used by Eiter and Wang \shortcite{EiterW08}.

\paragraph{SM-Forgetting}
Wang et al.\ \shortcite{WangWZ13} defined a modification of HT-Forgetting, SM-Forgetting, for extended programs, with the objective of preserving the answer sets of the original program (modulo the forgotten atoms). 
\begin{align*}
\classF_{\smF} \! = \! \{\fgt  \mid \HT(\f{P}{V}) \text{ is a maximal subset of } & \HT(P)_{\dagger V} \text{ s.t.\ } \as{\f{P}{V}}= \as{P}_{\parallel V} \\
& \text{ for all } P\in \classP(\fgt) \text{ and } V\subseteq\sign\}
\end{align*}
Hence, in this case, though the definition is a variation of the one for $\classF_{\htF}$, the semantic relations between the remaining atoms are preserved based on the equivalence of answer sets (utilizing the correspondence of HT-models).
A concrete operator was provided that builds on the notion of countermodels in HT-logic \cite{CabalarF07}, a technique, in fact, also used by Wang et al.~\shortcite{WangZZZ14} for HT-forgetting, and subsequently in the literature.
This class, similar to $\classF_{\htF}$, was shown to be closed for $\classP_{\ex}$ and $\classP_{\hor}$, and it was also shown that no operator exists that is closed for either $\classP_{\dis}$ or $\classP_{\nor}$. 
The authors also discussed relations to forgetting in propositional logic and uniform interpolation and considered the computational complexity of model checking, credulous and skeptical inference.

\paragraph{Strong AS-Forgetting}
Knorr and Alferes \shortcite{KnorrA14} introduced Strong AS-Forgetting with the aim of preserving not only the answer sets of $P$ itself but also those of $P\cup R$ for any $R$ over the signature without the atoms to be forgotten.
The notion was defined abstractly for classes of programs $\mathcal{C}$.
\begin{align*}
\classF_{\Sas} \! = \! \{\fgt\mid \as{\f{P}{V}\cup R}=\as{P\cup R}_{\parallel V} \text{ for all} \text{ programs } & R\in \mathcal{C} \text{ with } \sign(R)\subseteq \sign(P)\setminus V, \\ & \text{ for all } P\in \classP(\fgt) \text{ and } V\subseteq\sign \}
\end{align*}
The definition of this class is indeed closely aligned with property \pSP\ and naturally, the semantic relations between the remaining atoms are preserved based on the equivalence of answer sets (for varying programs).  
A concrete operator was presented for a non-standard class of programs (extended programs without disjunction), but not closed for $\classP_{\nor}$ and only defined for certain programs with double negation.
It is therefore not an operator for the class of normal programs according to Def.~\ref{def:forgOp}.
In parallel, Strong WF-Forgetting for normal programs under the well-founded semantics was considered with favorable results (closed and defined for the whole class), but is not further considered here.
The computational complexity was studied, but only in terms of computing a model.

\paragraph{SE-Forgetting}
Delgrande and Wang \shortcite{DelgrandeW15}  introduced SE-Forgetting based on the idea that forgetting an atom from program $P$ is characterized by the set of those SE-consequences, i.e., HT-consequences, of $P$ that do not mention the atoms to be forgotten.
The notion was defined for disjunctive programs building on an inference system by Wong \shortcite{Wong08} that preserves strong equivalence.
Given that $\vdash_s$ is the consequence relation of this system, $\Cn_\sign(P)$ is $\{r\in\lang_\sign\mid r \text{ disjunctive}, P\vdash_s r\}$.
The class is defined by:
\begin{align*}
\classF_{\SE} = \{\fgt\mid \f{P}{V} \htequiv \Cn_\sign(P)\cap \lang_{\sign(P)\setminus V} \text{ for all } P\in \classP(\fgt) \text{ and } V\subseteq\sign\}
\end{align*}
This notion is clearly aligned with the general notion of forgetting by Delgrande \shortcite{Delgrande17} relying on preserving the logical consequences over the remaining language.
An operator was provided, which is closed for $\classP_{\dis}$, based on computations using resolution, and a prototype implementation was made available.
It was observed that, though forgetting a single atom results only in a quadratic blow-up in the size of the program, forgetting several atoms yields an exponential blow-up of the resulting program (in the worst case).
Conflict solving was also revisited based on the framework presented by Eiter and Wang \shortcite{EiterW08}, aiming to provide more intuitive/better solutions, mainly due to the fact that HT-consequences were used in opposite to Semantic Forgetting that merely relies on preserving answer sets.  

The following three classes are all based on the same idea introduced by Knorr and Alferes \shortcite{KnorrA14}, i.e., they aim at preserving all the dependencies between atoms not being forgotten in the sense of property \pSP, but taking into consideration that it is not always possible to forget and satisfy \pSP\ \cite{GoncalvesKL-ECAI16}.
All three approaches rely on the manipulation of HT-models to ensure that the semantic relations between the remaining atoms are preserved, oriented by the idea to maintain all answer sets from the original program whenever this is possible.

To ease the reading and to keep the material self-contained, we recall here the necessary notions, adapted from criterion $\Omega$ \cite{GoncalvesKL-ECAI16}, which allows us to determine whether it is possible to forget a set of atoms while satisfying \pSP.

Let $P$ be a program over $\sign$, $V\subseteq \sign$, and $Y\subseteq \sign\setm V$. 
Consider the following.
\begin{align*}
\Rel_{\tuple{P,V}}^Y  & =\{A\subseteq V\mid \tuple{Y\cup A,Y\cup A}\in \HT(P) \text{ and }  \nexists A'\subset A \text{ such that }\tuple{Y\cup A',Y\cup A}\in \HT(P)\}\\
\RA^{Y,A}_{\tuple{P,V}} & =\{X\setm V\mid \tuple{X,Y\cup A}\in \HT(P)\}\\
\SRel^Y & =\{\RA_{\tuple{P,V}}^{Y,A}\mid A\in \Rel_{\tuple{P,V}}^Y\}
\end{align*} 
The set $\Rel_{\tuple{P,V}}^Y$ identifies those $A\subseteq V$ such that $Y\cup A$ is a potential answer set of $P$, which are therefore relevant for $Y$ being an answer set of the result of forgetting. 
For each such $A$, the set $\RA^{Y,A}_{\tuple{P,V}}$ collects the $V$-reduct of the left component of the HT-models of $P$ with right component $Y\cup A$.
For each $Y$, the set $\SRel^Y$ collects all such sets.

\paragraph{SP-Forgetting}
Gon\c{c}alves et al.~\shortcite{GoncalvesKL-ECAI16,GoncalvesKLW20} introduced SP-Forgetting with the aim to satisfy \pSP\ whenever that is possible. 
The notion was defined for extended logic programs, and, following criterion $\Omega$, tied to the existence of a least element in the set $\SRel^Y$, which, if it exists, coincides with the intersections over $\SRel^Y$. 
\begin{align*}
\classF_{\spF} =  \{\fgt\mid \HT(\f{P}{V}) \! = \! \{\tuple{X,Y}\mid Y\subseteq \sign(P)\text{\textbackslash} V \wedge X\!\in \bigcap\SRel^Y \} \text{ for all } P\in \classP(\fgt) \text{ and } V\subseteq\sign\}
\end{align*}

A concrete (semantic) operator was defined by Gon\c calves et al.~\shortcite{GoncalvesKLW20} based on countermodels, as well as a syntactic operator \cite{BertholdGKL19}. 
Both operators and the entire class are closed for $\classP_{\ex}$ and $\classP_{\hor}$, and it was shown that no operator exists that is closed for either $\classP_{\dis}$ or $\classP_{\nor}$. 

\paragraph{Relativized Forgetting}
Gon\c{c}alves et al.~\shortcite{GoncalvesKLW17,GoncalvesKLW20} introduced Relativized Forgetting as a solution to the fact that, in general, the result of forgetting according to SP-Forgetting may have answer sets that do not correspond to answer sets in the original program $P$.
Relativized Forgetting was originally defined using $V$-HT-models, an extension of HT-models closed related with relativized equivalence~\cite{EiterFW07}.
An alternative characterization \cite{GoncalvesKLW20}, which helps clarify its relation with SP-Forgetting, is used here.
\begin{align*}
\classF_{\AltNaive} =
\{\fgt\mid \HT(\f{P}{V}) \! = \! \{\tuple{X,Y}\mid Y\subseteq \sign\text{\textbackslash} V \wedge X\!\in \bigcup\SRel^Y \}\text{ for all } P\in \classP(\fgt) \text{ and } V\subseteq\sign\}.
\end{align*}
Thus, in comparison to SP-Forgetting, a union of the elements in the sets $\SRel^Y$ is used.

\paragraph{$F_M$-Forgetting} Gon\c{c}alves et al.~\shortcite{GoncalvesKLW17,GoncalvesKLW20} introduced $F_M$-Forgetting as an alternative to both SP-Forgetting and Relativized Forgetting, based on the fact that SP-Forgetting may introduce answer sets that do not correspond to those of the original program, while Relativized Forgetting may remove answer sets of the original program, even in cases where it would be possible to forget and satisfy \pSP. 
The difference between $\classF_{\spF}$ and $\classF_{\AltNaive}$ lies in the usage of intersection and union in their respective definitions.
Whenever $\SRel^Y$ has more than one element union and intersection will not coincide. Based on this, $F_M$-Forgetting was formally defined as follows based on a case distinction.
\begin{align*}
\classF_{\TAltNaive}=\{\fgt\mid \HT(\f{P}{V}) \! = \! \{\tuple{X,Y}\mid\ & Y\subseteq \sign\text{\textbackslash} V \text{ and }
   X\!\in \bigcup\SRel^Y \text{, if }  \SRel^Y \text{ has no least element,}\\
 &\text{or } X\!\in \bigcap\SRel^Y \text{, otherwise}\}\text{ for all } P\in \classP(\fgt) \text{ and } V\subseteq\sign\}
\end{align*}

For both of the latter classes, a concrete operator was defined based on countermodels, and, in both cases, the operator and the class are closed for $\classP_{\ex}$ and $\classP_{\hor}$. 
Moreover, no operator of the class exists that is closed for either $\classP_{\dis}$ or $\classP_{\nor}$ \cite{GoncalvesKLW20}.
In addition, a comparison between the three previous classes is presented, discussing when to prefer which class of operators, as well as an analysis of the computational complexity \cite{GoncalvesKLW20}.

\paragraph{Uniform Forgetting} Uniform Forgetting was introduced by Gon\c{c}alves et al.~\shortcite{GoncalvesJKLW19} in the context of forgetting in modular answer set programming. 
In this setting the programs are fixed, and only the input, i.e., sets of facts, varies, which is closely related with the notion of uniform equivalence. 
Uniform Forgetting aims at preserving the answer sets of $P$ no matter what input not containing the atoms to be forgotten is added to $P$. 
This implies a careful choice of the HT-models of a result of forgetting, and the formal definition requires additional technical notions which we recall here for self-containedness (in a resumed manner). 
\begin{align*}
Sel^Y_{\tuple{P,V}}&=\{A\subseteq V\mid \tuple{Y\cup A, Y\cup A}\in \HT(P)\}\\
T_{\tuple{P,V}}&=\{Y\subseteq\sign\setm V \mid~\text{there exists } A\in Sel^Y_{\tuple{P,V}} \text{ s.t. }  \tuple{Y\cup A', Y\cup A}\notin HT(P)\text{ for every } A'\subset A\}\\
N^{Y,A}_{\tuple{P,V}}&=\{X\setminus V\mid \tuple{X,Y\cup A}\in \HT(P) \text{ and } X\neq Y\cup A\}
\end{align*}
The sets $Sel^Y_{\tuple{P,V}}$ characterize all the different total HT-models of $P$ for each $Y\subseteq \sign\setm V$.
Among these, the ones that give rise to total HT-models of the result of forgetting are given by the set $T_{\tuple{P,V}}$.
For the non-total HT-models of the result of forgetting, we consider the set  $N^{Y,A}_{\tuple{P,V}}$ and the indexed family of such sets $\mathcal{S}^{Y}_{\tuple{P,V}}=\{N^{Y,i}\}_{i\in I}$ where $I=Sel^Y_{\tuple{P,V}}$.
For each tuple $(X_i)_{i\in I}$ such that $X_i\in N^{Y,i}$, the intersection of its sets is denoted as $\bigcap_{i\in I} X_i$, and $SInt^Y_{\tuple{P,V}}$ is the set of all such intersections. Formally, Uniform Forgetting combines the total models from $T_{\tuple{P,V}}$ and the non-total from $SInt^Y_{\tuple{P,V}}$ as folows:
\begin{align*}
\classF_{\up}=\{\fgt\mid  \HT(\f{P}{V}) = (\{\tuple{Y,Y}\mid Y\in T_{\tuple{P,V}}\}\  \cup\  \{\tuple{X,Y}\mid\  & Y\in T_{\tuple{P,V}} \text{ and }X \in SInt^Y_{\tuple{P,V}}\}),\\ 
 &\text{ for all } P\in \classP(\fgt) \text{ and } V\subseteq \sign\}.
\end{align*}
Again, a concrete operator was defined based on countermodels, which, just like $\classF_{\up}$ itself, is closed for $\classP_{\ex}$ and $\classP_{\hor}$. 
A further operator was defined which combines the former with a syntactic one whenever this is possible \cite{GoncalvesJKL21}.
Moreover, no operator of the class exists that is closed for either $\classP_{\dis}$ or $\classP_{\nor}$. 
In addition, the effects of applying forgetting to answer set programming modules were studied, as well as the computational complexity.
	
While all these classes were introduced with differing motivations, they coincide under certain conditions, e.g., when restricted to specific classes of programs. 
This is the case for Horn programs, where the result of forgetting according to most of the classes of operators presented above are strongly equivalent.

\begin{theorem}\label{prop:coincideHorn}
For all Horn programs $P$, every $V\subseteq \sign(P)$, and all forgetting operators $\fgt_1, \fgt_2$ in the classes $\classF_{\strong}$, $\classF_{\weak}$, $\classF_{\FS}$, $\classF_{\htF}$, $\classF_{\smF}$, $\classF_{\Sas}$, $\classF_{\SE}$, $\classF_{\spF}$, $\classF_{\AltNaive}$, $\classF_{\TAltNaive}$, and $\classF_{\up}$,  it holds that $\fgt_1(P,V)\htequiv\fgt_2(P,V)$.
\end{theorem}

Notably, of the classes of operators presented in this section, $\classF_{\sem}$ and $\classF_{\FW}$ are the only ones that do not coincide with all others when restricted to Horn programs.

\begin{example}
Consider the following Horn program $P=\{a\la e,\ e\la b,\ b\la,\ c\la d\}$.
Then, for any $\fgt$ in any of the classes mentioned in Thm.~\ref{prop:coincideHorn}, we have that $\f{P}{\{e\}}$  is strongly equivalent to the program $\{a\la b,\ b\la,\ c\la d\}$.
None of the three concrete operators defined for $\classF_{\sem}$ actually satisfies this condition, because the two semantic operators do not consider $c\la d$ as it is not relevant for the answer sets, while the syntactic one discards this rule in its pre-processing.
However, a modification of the syntactic operator is possible such that it coincides with the result (by omitting part of the pre-processing), i.e., though $\classF_{\sem}$ does not align with Thm.~\ref{prop:coincideHorn}, at least one corresponding operator exists.
For $\classF_{\FW}$, the result completely differs since any operator in $\classF_{\FW}$ must include $\la b$ in its result.
\end{example}

Interestingly, Wang et al. \shortcite{WangZZZ12,WangZZZ14} additionally showed that, for $\classP_{\hor}$, the result of $\classF_{\htF}$ is strongly equivalent to that of classical forgetting. 
We thus obtain as a corollary that this holds for all classes of forgetting operators mentioned in Thm.~\ref{prop:coincideHorn}.  

Besides the coincidence when restricted to the class of Horn programs, there are two classes of operators that turn out to coincide. 
\begin{theorem}\label{prop:EqualityFSFSE}
Consider the class of disjunctive programs.
Then, $\classF_{\FS}$ and $\classF_{\SE}$ coincide.
\end{theorem}
\noindent
This coincidence can be traced back to the fact that the inference system used for $\classF_{\SE}$ is the same as that used to define the example operator for $\classF_{\FS}$. 
Since the two classes coincide, in what follows, we often use $\classF_{\FS}$ to refer to both.
This correspondence can be extended to $\classF_{\htF}$ when the result of forgetting is still in the class of disjunctive programs.

\begin{theorem}\label{prop:SEFequalsHTF}
Let $P$ be a disjunctive program, $V\subseteq \sign(P)$, $\fgt_{\FS}\in\classF_{\FS}$, $\fgt_{\htF}\in\classF_{\htF}$, and $\fgt_{\SE}\in\classF_{\SE}$.
Then, $\fgt_{\FS}(P,V)\htequiv\fgt_{\htF}(P,V)\htequiv\fgt_{\SE}(P,V)$ whenever $\fgt_{\htF}(P,V)$ is strongly equivalent to a disjunctive program.
\end{theorem}
\noindent
This does not hold in general though, as the next example shows.
\begin{example}\label{exampleHTvsFS}
Given $P=\{a\la \nf b$, $b\la \nf a$,  $\la a,b\}$, consider forgetting about $b$ from $P$.
For any $\fgt_{\htF}\in\classF_{\htF}$, it is easy to see that $\fgt_{\htF}(P,\{b\})$ is strongly equivalent to $a\la\nf\nf a$, which is not strongly equivalent to any disjunctive program. In the case of $\classF_{\SE}$ and $\classF_{\FS}$ the result of forgetting is, by definition, always a disjunctive program.
\end{example}
This also means that item 1.\ in Prop.~2 \cite{DelgrandeW15}, which semantically characterizes $\classF_{\SE}$ by asserting that it coincides with the set of HT-models restricted to the remaining atoms, i.e., that it coincides with $\classF_{\htF}$ on the class of disjunctive programs, actually does not hold. 

\paragraph{Forgetting Operators}
Concrete forgetting operators have been considered in all the presented approaches, either in the form of a syntax-based operator, namely for $\classF_{\strong}$, $\classF_{\weak}$, $\classF_{\sem}$, $\classF_{\Sas}$, and $\classF_{\spF}$, or based on a semantic characterization, namely for $\classF_{\sem}$, $\classF_{\htF}$, $\classF_{\smF}$, $\classF_{\spF}$, $\classF_{\rF}$, $\classF_{\mF}$, and $\classF_{\up}$, or one combining the latter two principles, namely for $\classF_{\up}$, or based on a consequence relation, namely $\classF_{\FS}$, $\classF_{\FW}$, and $\classF_{\SE}$.

On closer inspection, the syntactic operators share a common basis in the form of the principle of weak partial evaluation (WGPPE) \cite{BrassD99}, which is present in the construction of each of them, even in the realization of the consequence relation used in the case of $\classF_{\SE}$.
Hence, positive occurrences in the rule bodies of atoms to be forgotten are generally treated in the same way.
The difference lies in the treatment of negation and of the rules not mentioning the atoms to be forgotten, as well as the applicability of the operator in question. 
In the case of $\classF_{\strong}$ and $\classF_{\weak}$, the treatment of negation is not founded on a semantic principle, which causes problems as argued by Eiter and Wang \shortcite{EiterW08}.
The other three operators defined for the classes $\classF_{\sem}$, $\classF_{\Sas}$, and $\classF_{\spF}$ as well as the syntactic part of the operator defined for $\classF_{\up}$, even share the basic ideas for treatment of negation, inspired by the first such approach for $\classF_{\sem}$.
As argued by Berthold et al.~\shortcite{BertholdGKL19}, they then differ in how rules not mentioning the atoms to be forgotten are treated.
Namely, the operator in $\classF_{\sem}$ often simplifies these away, as its semantic notion is only relying on answer sets.
On the other hand the operator defined for $\classF_{\Sas}$ is only applicable in a very restricted setting.

For the semantic operators, two major approaches exist.
The operators defined for $\classF_{\sem}$ rely on computing the answer sets, applying the definition of forgetting and finding a canonical program that represents the resulting set of answer sets.
All the other approaches rely on the construction based on countermodels \cite{CabalarF07}, as first used by Wang et al.~\shortcite{WangWZ13}.
The benefit is that one can simply determine the set of HT-models of a result of forgetting, and then provide a corresponding program based on countermodels.
The difference between these operators resides then only in the characterization of the desired HT-models themselves.
While this is an elegant way to obtain a result of forgetting, that can in fact be applied to any approach based on HT-models, the resulting program is often not in a minimal form, containing many unnecessary rules, which requires further non-trivial considerations on minimal programs \cite{CabalarPV07}.
This also impacts on the similarity between the original program and a result of forgetting, i.e., while the semantic characterization is precisely matched, syntactically they may be completely different, even for rules that do not mention the atoms to be forgotten.
The latter can be avoided in certain cases, as argued by Gon\c calves et al.~\shortcite{GoncalvesKLW20}, namely if property \pSI\ is satisfied, as this allows us to exclude the rules not mentioning atoms to be forgotten from the forgetting process, and simply pass them to the result.
Still, for the rules involving the atoms to be forgotten, the result may bear no resemblance.
The approaches based on a consequence relation do not suffer from this problem, i.e., the rules of the program that are not removed while forgetting do persist, they are, however, accompanied by a huge number of additional rules that do not add anything to the program which would require additional processing and simplification.
This is what makes providing syntactic operators for classes with a well-defined semantic characterization an important approach.
It is not an easy one though.

\paragraph{Computational Complexity}
We finish the section with considerations on the computational complexity of forgetting.
To begin with, all approaches showed or mentioned that computing a concrete result of forgetting with one particular operator is in general in EXP.
In addition, many approaches provided arguments and results that show that forgetting is a computationally expensive task, but also argued that this is not surprising given the computational complexity of problems such as model existence in answer set programming \cite{DantsinEGV01}.
A general comparison is however not straightforward as several approaches are quite distinct, which impacts on the kind of complexity results that are presented.
In the following, we therefore focus on the results presented in the literature, in particular on two problems whose complexity is considered in several approaches, as this makes them suitable for such a comparison.
In what follows, we assume a basic understanding of standard complexity classes as well as the polynomial hierarchy.

\begin{table}
\renewcommand*{\arraystretch}{1.4}
\begin{tabular}{l|ccc}
& $\classP_{\nor}$ & $\classP_{\dis}$ & $\classP_{\ex}$\\
\cline{1-4}
$\classF_{\strong}$ & $\coNP$ & $-$ & $-$ \\
$\classF_{\weak}$ & $\coNP$ & $-$ & $-$\\
$\classF_{\sem}$ & $\coNP$ & $\PiP{2}$ & $-$\\
$\classF_{\htF}$ & $\tM$ & $\tM$ & $\coNP$\\
$\classF_{\smF}$ & $\tM$ & $\tM$ & $\PiP{2}$
\end{tabular}
\caption{Known complexity results for skeptical reasoning under forgetting. All results are completeness results. For class $\classF$ (of operators) and class $\classP$ (of programs), '\protect\tM' represents that $\classP$ has not been considered for this problem for $\classF$, and '-' that $\classF$ is not defined for $\classP$.
}
\label{fig:complexitySkept}	
\end{table}

The first problem is skeptical reasoning under forgetting, i.e., is some atom true in all answer sets of a forgetting result $\f{P}{V}$.
Formally, given $\fgt\in\classF$, $P\in \classP(\fgt)$, $V\subseteq \sign$, and $a\in \sign\setminus V$, we determine $\f{P}{V}\models_s a$, where $\models_s$ denotes skeptical inference, i.e., truth in all answer sets.
This problem has been considered for $\classF_{\strong}$, $\classF_{\weak}$, $\classF_{\sem}$, $\classF_{\htF}$ and $\classF_{\smF}$ for different classes of programs.
In fact, for $\classF_{\sem}$, $a$ can be a literal, and for $\classF_{\htF}$ a formula, but we simplified this aspect here for the sake of the comparison.
Fig.~\ref{fig:complexitySkept} summarizes the results.
We can observe that, for normal programs, $\classF_{\strong}$, $\classF_{\weak}$, which are only defined for this class of programs, and $\classF_{\sem}$ coincide. 
In fact, though it has not been considered explicitly for normal programs, the result for $\classF_{\htF}$ does also coincide (for the larger class of programs) due to hardness of skeptical reasoning for ASP even without forgetting.
For $\classF_{\smF}$, this is not likely due to the maximization check.
In general, for classes $\classP_{\dis}$ and $\classP_{\ex}$ for which skeptical reasoning without forgetting is $\PiP{2}$-complete, we observe that $\classF_{\sem}$ and $\classF_{\smF}$ are computationally more expensive than the other classes due the additional minimization/maximization which is part of their respective definitions. 
We note that only $\classF_{\sem}$ and $\classF_{\smF}$ do also consider credulous reasoning.
In the former case, the result raises to $\SigmaP{3}$, while, in the latter, it remains on the same level of the hierarchy, i.e., $\SigmaP{2}$.

Thus, requiring that answer sets be preserved comes at a cost in terms of computational complexity.
When comparing $\classF_{\sem}$ with both $\classF_{\strong}$ and $\classF_{\weak}$ with their lack of semantic  grounds (for part of their construction), arguably the additional cost is preferrable.
When comparing $\classF_{\htF}$ and $\classF_{\smF}$, this is less straightforward, as $\classF_{\htF}$ clearly is based on semantic grounds.
Such considerations therefore depend on the intended usage and whether preserving answer sets justifies the additional cost.

The second problem considered is determining whether a given program is indeed a result of forgetting.
Formally, given $\fgt\in\classF$, $P,P'\in \classP(\fgt)$, $V\subseteq \sign$, we determine $P'\htequiv\f{P}{V}$.
This problem has been considered for $\classF_{\htF}$, $\classF_{\smF}$, $\classF_{\spF}$, $\classF_{\rF}$, $\classF_{\mF}$, and $\classF_{\up}$ and Fig.~\ref{fig:complexityResult} presents the results.
It can be observed that the results for $\classF_{\htF}$ and $\classF_{\smF}$ do coincide (unlike the first problem), and that there is an increase in terms of complexity for the classes that relate to property \pSP.
This is probably not surprising, as $\classF_{\spF}$, $\classF_{\rF}$, and $\classF_{\mF}$ all relate to criterion $\Omega$ that establishes whether it is possible to forget and preserve $\pSP$ for the concrete combination of program and atoms to be forgotten, which has been shown to be $\SigmaP{3}$-complete \cite{GoncalvesKLW20}.
The result for class $\classF_{\up}$ also points into this direction, but hardness and completeness remain to be shown.
Still, the known complexity results on uniform equivalence \cite{EiterFW07} do indicate that it is not likely that an improvement can be achieved in comparison to the other three classes.

Here, the main conclusion is that trying to preserve the answer sets for arbitrary programs (or sets of facts) to be added increases the computational complexity.
In our view, this is preferable unless the intended usage does not require it.
We will revisit this question in the following section. 

\begin{table}
\renewcommand*{\arraystretch}{1.4}
\begin{tabular}{cccccc}
$\classF_{\htF}$ & $\classF_{\smF}$ & $\classF_{\spF}$ & $\classF_{\rF}$ & $\classF_{\mF}$ & $\classF_{\up}$\\
\cline{1-6}
$\PiP{2}$ & $\PiP{2}$ & $\PiP{3}$ & $\PiP{3}$ & $\PiP{3}$ & $\PiP{3}$
\end{tabular}
\caption{Known complexity results for determining whether a given program is a result of forgetting. All results are completeness results but $\classF_{\up}$, which is only an inclusion.
}
\label{fig:complexityResult}
\end{table}

\section{On the Properties of Existing Operators}\label{sec:results}

Having presented the properties and classes of operators introduced in the literature for forgetting in ASP, in this section, we provide a detailed comparison of these classes with respect to their characteristics.
In more detail, we draw a precise picture on the relations between classes of operators and the properties they satisfy.
We then discuss the suitability of these classes based on their characteristics, establish concisely the relationship to uniform interpolation, and discuss their suitability w.r.t. several applications considered in the context of forgetting.

\subsection{Specific Properties}\label{subsec:specific}

Putting aside for a moment the considerations on the suitability of properties at the end of Sect.~\ref{sec:props}, the \emph{desirability} of these properties is, to some extent, in the eye of the beholder. Often, a particular novel approach to forgetting is justified by the fact that previous approaches did not obey some new property deemed crucial, neglecting however that this novel approach actually ended up failing to satisfy other properties, themselves deemed crucial by those who introduced them. Whereas the introduction of most known approaches to forgetting was accompanied by a study of some properties they each enjoyed, there are many missing gaps, some because some properties were only introduced later, others because they were simply neglected. 
Despite the discussion and potential controversy around the adequacy of the properties, which may ultimately depend on the application at hand, the first and perhaps most important step is to draw an exhaustive picture regarding which properties are obeyed by which classes of operators. This takes us to the central theorem of our paper, illustrated in one easy-to-read table.
 
 \renewcommand{\arraystretch}{1.15}
 
\begin{table}[h]
\centering
\begin{adjustbox}{width=1\textwidth,center=\textwidth}
$\begin{array}{rccccccccccccc}
&\classF_{\strong} &  \classF_{\weak}  &  \classF_{\sem}  &  \classF_{\FS}  &  \classF_{\FW}  &  \classF_{\htF}  &  \classF_{\smF}  &  \classF_{\Sas}  &  \classF_{\SE}  &  \classF_{\spF}  &  \classF_{\rF}  	&  \classF_{\mF}  &  \classF_{\up} \\\cline{1-14}
{\spsC\ } & \tM    	& \tM   	&   \cM	&\tM    	&\cM   	&\tM    	&\cM  	&\cM  	&\tM        	&\tM  	&\cM  	&\cM        &\cM        \\\cline{1-14}
{\spwE\ } & \tM   	& \tM   	&  \cM	&\tM   	&\cM  	&\tM   	&\cM  	&\cM  	&\tM   	&\tM  	&\tM  	&\cM   	&\cM   	\\\cline{1-14}
{\spSE\ } &  \tM 	&  \tM 	&  \tM	&  \cM	&  \cM	&  \cM	&  \cM	&  \cM	&  \cM	&  \cM	&  \cM	&  \cM	&  \cM	        \\\cline{1-14}
{\spW\ } &\cM		&  \tM	& \tM		&\cM   	&  \tM	&   \cM 	&   \tM 	&\cM     	&   \cM 	&   \tM 	&\tM     	&   \tM 	&   \tM 	\\\cline{1-14}
{\spPP\ } &   \tM 	&\cM 	&   \tM	 &  \cM	  &\cM  	&   \cM	&   \cM	&   \cM	  &   \cM	&   \cM	&   \cM	  &   \cM	&   \tM	\\\cline{1-14}
{\spSI\ }& \cM & \cM  &  \tM   &   \tM & \cM & \cM & \tM  & \cM & \tM  & \cM  & \cM & \tM  & \tM  										\\\cline{1-14}
{\spSC\ }&  \tM 	&  \tM 	&  \tM	&  \cM	&  \cM	&  \cM	&  \tM	&  \cM	&  \cM	&  \tM	&  \tM	&  \tM	&  \tM		\\\cline{1-14}
{\spRC\ }   &\cM		&  \cM	& \tM		&\cM   	&  \cM	&   \cM 	&   \tM 	&\cM     	&   \cM 	&   \tM 	&\cM	     	&   \tM 	&   \tM \\\cline{1-14}	
{\spNC\ } &   \cM 	&\cM 	&   \tM	 &  \cM	  &\cM  	&   \cM	&   \tM	&   \cM	  &   \cM	&   \tM	&   \cM  	&   \tM	&   \tM	\\\cline{1-14}
{\spPI\ }	& \cM 	& \cM  	&  \cM   	&   \cM 	& \cM 	& \cM 	& \cM  	& \cM 	& \cM  	& \tM  	& \tM 	& \tM 	& (\cM) 	 \\\cline{1-14}
{\spCP\ } &   \tM 	&   \tM 	&   \tM 	& \tM  	& \tM   	&  \tM	&  \cM	&  \cM	&	\tM	&  \tM	&  \tM	&	\cM	&	\cM	\\\cline{1-14}
{\spSP\ }   &  \tM  	& \tM		& \tM		&\tM		&	\tM	& \tM		& \tM		& \cM	&	\tM	& \tM		& \tM		&	\tM	&	\tM	\\\cline{1-14}
${\spwC\ }$&\tM		&\tM		&\tM		&\tM		&\tM		&\tM		& \cM	& \cM	&  \tM	& \cM	& \tM		&  \cM	&  \cM	\\\cline{1-14}
{\spwSP\ } &   \tM 	&\tM 		&   \tM	 &  \tM	  &\tM  	&   \tM	&   \tM	&   \cM	  &   \tM	&   \cM	&   \tM	  &   \tM	&   \tM	\\\cline{1-14}
{\spsSP\ }& \tM 	& \tM  	&  \tM   	&   \tM 	& \cM 	& \tM 	& \tM  	& \cM 	& \tM  	& \tM  	& \cM 	& \cM  	& \tM  	\\\cline{1-14}
{\spUP\ } &   \tM 	 &   \tM 	 &   \tM 	 & \tM  	 & \tM   	 &  \tM	 &  \tM	 &  \cM	 &	\tM	 &  \tM	 &  \tM	 &	\tM	 &	\cM	\\\cline{1-14} 
${\spUI\ }$ &\cM	&\cM	&\tM		&\tM		&\cM	&\cM	& \tM	& \cM	&  \tM	& \cM	& \cM	&  \tM	&  \cM					\\\cline{1-14} 
{\spSIu\ } & \cM 	& \cM  	&  \tM   	&   \tM 	& \cM 	& \cM 	& \tM  	& \cM 	& \tM  	& \cM  	& \cM 	& \tM  	& \tM  	\\\cline{1-14} 
${\spE{\classP_{\hor}}}$	 & \cM	& \cM	& \cM	& \cM	& \cM	&  \cM	&  \cM	& \cM	& \cM	&  \cM	& \cM	& \cM	& \cM		\\\cline{1-14}		
${\spE{\classP_{\nor}}}$	&  \cM	&  \cM	&  \cM	& 	\tM	& \cM	&  \tM 	&  \tM 	&  -		& \tM		&  \tM 	&  \tM	& \tM		& \tM		\\\cline{1-14}
${\spE{\classP_{\dis}}}$	& -		& -		&  \cM	&  \cM	&  \cM	&  \tM	&  \tM	& -		&  \cM	&  \tM	& \tM		&  \tM	&  \tM	\\\cline{1-14}			
${\spE{\classP_{\ex}}}$ & -		& -		& -		& -		& -		&  \cM	&  \cM	& -	 	& -		&  \cM	& \cM	& \cM	& \cM	\\\cline{1-14}
\end{array}
$
\end{adjustbox}
\caption{Satisfaction of properties for known classes of forgetting operators. For class $\classF$ 
and property \spP,  
'\cM' represents that $\classF$ 
satisfies \spP, '\protect\tM' that $\classF$ does not satisfy \spP, and '-' that $\classF$ is not defined for the class $\classP$ in consideration.
}
\label{fig:properties} 
\end{table} 
 
\begin{theorem}\label{thm:propsOps}
All results in Table~\ref{fig:properties} hold. 
\end{theorem}

One first observation is that every class of operators obeys a different set of properties (apart from $\classF_{\FS}$ and $\classF_{\SE}$, which coincide, cf.\ Thm.~\ref{prop:EqualityFSFSE}). 
This is a strong indication that these properties play a role in characterizing the classes of operators. In fact, a precise characterization of some classes of operators in terms of the properties they satisfy sometimes exists \cite{WangZZZ12,WangZZZ14,DelgrandeW15}, although this not the case in general.

We now focus on analyzing specific properties and how they relate to the known classes of operators, following commonly the presentation of the properties from left to right according to the historical order established in Sect.~\ref{sec:props}.

Starting with \psC\ and \pwE, we know, by Thm.~\ref{prop:relations}, that any $\classF$ that is known to satisfy \pCP\ also satisfies these two.
Not surprisingly, $\classF_{\sem}$ also satifies both, even though it does not satisfy \pCP, since the proposal is based on the ideas behind these properties.
For the remaining classes, it is worth illustrating why $\classF_{\FS}$, $\classF_{\htF}$, and $\classF_{\SE}$ do not satisfy \psC\ by looking at the example where we forget about $a$ from $P=\{a\la \nf a\}$: all three classes require the result to be strongly equivalent to $\emptyset$, i.e., the forgetting operation introduces a new answer set.
Turning to \pwE, it requires that the results of forgetting about $p$ from $P=\{q \la \nf p, q\la \nf q\}$ and from $Q=\{q\la\}$ have the same answer sets, while the three classes $\classF_{\FS}$, $\classF_{\htF}$, and $\classF_{\SE}$ require that the results be strongly equivalent to $\f{P}{p}=\{q\la\nf q\}$ and $\f{Q}{p}=\{q\la\}$, respectively, which are obviously not equivalent. 
$\classF_{\FW}$ satisfies both properties (though not satisfying \pCP): in the previous two examples, $\bot$ must be returned in the former, while $\f{P}{p}$ includes $q\la$ in the latter.

The properties \pSE, \pW, and \pPP\ have received more attention in the literature, although focussing more on the properties not satisfied by previous approaches to motivate the introduction of a new one. As a result, several novel positive results are included in the table. It is perhaps worth pointing out that despite Wang et al.~\shortcite{WangZZZ14} having discussed that $\classF_{\FS}$ and $\classF_{\FW}$ do not satisfy \pPP, they did so using a counterexample -- 
Ex.~\ref{exampleHTvsFS} in this paper -- that is not really part of the language for which $\classF_{\FS}$ and $\classF_{\FW}$ are defined, since it relies on rules with double negation as missing consequences, which in our view seems to be an unfair argument. According to our uniformized notions, for the language for which they are defined, they satisfy \pPP.
In any case, we can observe that \pSE\ and \pPP\ are satisfied by most of the classes, basically only some of the early approaches do not satisfy these, and we can arguably conclude that these are vastly consensual properties of forgetting.
The same does not hold for \pW\ where the incompatibility result with \pwC\ (and thus \pCP), item \ref{i24} in Thm.~\ref{prop:relations}, affects the results.
\begin{example}\label{exampleFsasNotW}
Consider $P=\{a\la \nf b$, $b\la \nf c\}$ whose only answer set is $\{b\}$.
Thus, according to \pwC, a forgetting result must contain at least this answer set modulo the forgotten atoms.
For example, for $\fgt\in\classF_{\spF}$, $\f{P}{b}$ contains $a\la\nf\nf c$, whose only answer set is $\{\}$. 
However, this rule is not an HT-consequence of $P$, hence \pW\ is not satisfied.
\end{example}

\pSI\ has received less attention, yet often this non-trivial property is satisfied.
Wong \shortcite{Wong09} showed \pSI\ for strong and weak forgetting, but using $t$-equivalence instead of $HT$-equivalence, whose semantics differs.
This explains the negative result in the case of $\classF_{\FS}$.
The negative result for $\classF_{\SE}$ follows by correspondence to $\classF_{\FS}$, and for $\classF_{\smF}$ from forgetting about $b$ from $P$ as in Ex.~\ref{exampleFsasNotW}:
$\f{P}{b}\htequiv \emptyset$ for $\fgt\in\classF_{\smF}$, so adding $c\la$ results precisely in a program containing this fact.
If we add $c\la$ before forgetting, then the $HT$-models of the result of forgetting, ignoring all occurrences of $b$, correspond precisely to $\langle c,c\rangle,\langle c, ac\rangle$, and $\langle ac,ac \rangle$.
To preserve the answer sets of this modified program (there is only one -- $\{a,c\}$), only the last of these three HT-models can be considered.
Hence, $a\la$ and $c\la$ (or strongly equivalent rules) occur in the result of forgetting for any $\fgt\in\classF_{\smF}$, and \pSI\ does not hold.
Still, as pointed out in Sec.~\ref{sec:props}, \pSI\ is an important property as it allows one to focus forgetting on the rules that contain the atoms to be forgotten.

For the other three HT-related properties originally proposed by Wong \shortcite{Wong09}, \pSC, \pRC, and \pNC, the consensus is less obvious.
Considering first \pSC, we recall from \ref{i11}.\ and \ref{i12}.\ of Thm.~\ref{prop:relations} that it is implied by \pPP\ and \pW\ together, and in turn implies \pSE.
In fact, four of the five classes that satisfy \pSC, also satisfy \pPP\ and \pW.
Only $\classF_{\FW}$ does not satisfy \pW, and still satisfy \pSC\ (according to an first example presented by Wang et al.~\shortcite{WangZZZ14}): 
Consider $P=\{p\la \nf q; q\la \nf p\}$. 
For any $\fgt\in\classF_{\FW}$, forgetting about $p$ is strongly equivalent to $q\la$, which is not a consequence of $P$ itself.
Regarding the relation to \pSE, we can observe that there are several cases where \pSE\ holds, while \pSC\ does not, showing that both properties are relevant and indicating that \pSC\ can be seen as a stronger version of \pSE.
Still, given the broad consensus for \pSE, arguably \pSE\ seems preferable among the two.

Regarding \pRC\ and \pNC, we note that both are always satisfied for the same cases.
This could be based on some correlation, but none is known.
Arguably, this coincidence could be based on the fact that both are closely tied to the concrete
definitions of $\classF_{\FS}$ and $\classF_{\FW}$ along which they were introduced.
Actually, inspecting the definitions of both properties, one could consider that \pNC\ implies \pRC, i.e., that the rule which is an HT consequence in \pNC\ is somehow the rule $r'$ in \pRC, but this is not the case in general.
\begin{example}
Consider $\classF_{\strong}$ which satisfies \pNC\ and $P=\{a\la b\}$. 
We have that $\fgt_{\strong}(P,\{p\})=P$. 
Since $\fgt_{\strong}(P,\{p\})\htmodels a\la b$, we have, by \pNC, that $P\htmodels a\la b, \nf p$. 
But we also know that $\fgt_{\strong}(\{a\la b, \nf p\},\{p\})=\emptyset$, and, therefore $\fgt_{\strong}(\{a\la b, \nf p\},\{p\})\htmodels a\la b$ does not hold. This shows that there are cases where $r'$ of \pNC\ is not the rule for property \pRC.
\end{example}
Nevertheless,  item \ref{i13}.\ of Thm.~\ref{prop:relations} indicates that \pW\ implies \pNC, and there are several cases where the latter is satisfied and the former is not, indicating that \pW\ has a more restrictive condition than \pNC.
This is also corroborated by the two incompatibility results in Thm.~\ref{prop:relations} and the apparent coincidence of satisfaction for \pRC\ and \pNC\ in Table~\ref{fig:properties}, in the sense that \pW\ also has a more restrictive condition than \pRC.

Property \pPI\ is also widely accepted, i.e., a set of atoms can be forgotten in any order.
The only exceptions are the three classes closely tied to \pSP, $\classF_{\spF}$, $\classF_{\rF}$, and $\classF_{\mF}$, essentially, because it is not always possible to forget and preserve \pSP\ \cite{GoncalvesKL-ECAI16}.
In such situations, the three classes then provide approximations of forgetting while preserving \pSP, and, since the order in which atoms are forgotten may affect whether forgetting is possible (while preserving \pSP) \cite{GKL17}, the property does not hold for any of them.
Regarding $\classF_{\up}$, we note that a weaker version of \pPI\ was proven \cite{GoncalvesJKLW19}, showing that it is possible to iterate the operators of the class when applied in the context of modular logic programming.
Thus strictly speaking the exact result is open.
On the other hand, in the context of modular answer set programming, directed towards uniform equivalence, this result suffices, which explains the particular notation of the result in Table~\ref{fig:properties}.

The new negative results for \pCP\ and \pSP\ can be illustrated with forgetting about $b$ from $P=\{a\la \nf b$, $b\la\nf a\}$, i.e., the first two rules of Ex.~\ref{exampleHTvsFS}.
Since $\as{P}=\{\{a\},\{b\}\}$, the result must have two answer sets $\{a\}$ and $\emptyset$, which is not possible for disjunctive programs obtained from operators in $\classF_{\FS}$, $\classF_{\FW}$, and $\classF_{\SE}$.
The same example serves as counterexample for all negative results of \pwC, while positive results follow for classes satisfying \pCP\ from Thm.~\ref{prop:relations}.
Notably, the counterexample also applies to $\classF_{\SE}$, thus invalidating Thm.\ 2 in the paper of Delgrande and Wang~\shortcite{DelgrandeW15}.

Regarding \pSP, only $\classF_{\Sas}$ satisfies the property due to the way the class is defined.
However, an important note is in order.
Namely, unlike previously stated by Gon\c calves et al.~\shortcite{GoncalvesKL16,GKL17,GoncalvesJKLW19}, in Table~\ref{fig:properties}, all results are positive for $\classF_{\Sas}$.
The following consideration reveals the cause for this discrepancy.
Many of the negative results previously stated for $\classF_{\Sas}$ are based on counterexamples using the concrete operator defined in \cite{KnorrA14}, among them, for example, that $\classF_{\Sas}$ does not satisfy \pUP\ \cite{GoncalvesJKLW19}.
However, items \ref{i7}., \ref{i22}., and \ref{i23}.\ of Thm.~\ref{prop:relations} allow us to show that \pSP\ implies \pUP, seemingly a contradiction.
The reason why this is not a problem is revealed by a close inspection of the definition of class $\classF_{\Sas}$.
The operator defined by Knorr and Alferes \shortcite{KnorrA14} is actually not defined for a standard class of programs, nor for all programs in this class which is in conflict with the quantification applied in the definition.
This is complemented by the general result that forgetting and satisfying \pSP\ is not always possible for programs that include normal programs.
These observations lead to the following corollary, strongly restricting the class of programs to which operators in $\classF_{\Sas}$ can be applied.

\begin{corollary}
There is no forgetting operator $\fgt\in\classF_{\Sas}$ with $\classP_{\nor}\subseteq\classP(\fgt)$.
\end{corollary}
Thus, the operator defined by Knorr and Alferes \shortcite{KnorrA14} does not belong to $\classF_{\Sas}$ and all previously negative results for $\classF_{\Sas}$ are invalidated, allowing that this class indeed satisfies all the properties in the literature considered here, but only within the scope of Horn programs.

Regarding \pwSP\ and \psSP, these properties are essentially exclusive to the \pSP-related classes of forgetting, which is not surprising given item \ref{i16}.\ of Thm.~\ref{prop:relations}, with the exception of \psSP\ for $\classF_{\FW}$, as this class satisfies both \psC\ and \pSI, which implies satisfaction of \psSP\ by item \ref{i17}.\ of Thm.~\ref{prop:relations}.

Concerning \pUP\ this is indeed also a strong property, as it is essentially only satisfied by the class for which it was defined, besides $\classF_{\Sas}$, of course.
In fact, \pUP\ can be seen as an adaptation of \pSP\ to uniform equivalence.
Thus, despite the reduced number of classes that satisfy it, it is an important property as it shows that the conceptual idea of \pSP\ can be satisfied in the scope of general programs (not restricted to Horn), if one only varies facts for the sake of preserving the dependencies for the atoms not to be forgotten.
This is also well-aligned with a central idea of answer set programming, where the general specification of a problem is encoded as an answer set program which is combined with different set of facts, representing an instance of the problem one wants to solve.

With respect to \pUI, we observe that satisfaction is vastly sanctioned by the fact that it is implied by \pSI\ (cf.~\ref{i22} of Thm.~\ref{prop:relations}), with the exception of $\classF_{\up}$ for which the positive result is a result of \ref{i23}.\ of Thm.~\ref{prop:relations}.
We exemplify the negative result for the case of $\classF_{\FS}$. 
Consider forgetting about $p$ from $P=\{a\la \nf p,b; \ p\la \nf a;\  \la p,b \}$.  
This program has $a\la \nf\nf a,b$  as an HT-consequence, which is however not a disjunctive rule and thus not captured in the construction. So adding $b\la$ to the program makes $a$ true, and forgetting preserves that. 
If we forget first, that connection is lost and we cannot conclude $a$ by adding $b\la$ to the result of forgetting. 

Although $\pSIu$ is a weakening of \pSI, the results for $\pSIu$ are often a consequence of those for \pSI. The positive ones follow immediately from the fact that $\pSIu$ is implied by \pSI\ (cf.~\ref{i27} of Thm.~\ref{prop:relations}), and most of the counterexamples for the negative results are similar to those of \pSI. An interesting exception is the case of $\classF_{\mF}$, for which the negative result for \pSI\ can be easily justified by the satisfaction of \pCP, as these properties together are equivalent to \pSP. In the case of $\pSIu$ the same argument cannot be used, since this property together with \pCP\ is not enough to imply \pSP.
Rather an advanced counterexample can be found based on HT-models in which the result is different from that of $\classF_{\spF}$ and $\classF_{\rF}$, which both satisfy $\pSIu$ (for the details we refer to the appendix).   

Finally, the results on \pE{\classP} for different classes of programs $\classP$ reveal that all classes of operators are closed for $\classP_{\hor}$, and, in addition, each $\classF$ is closed for the maximal class of programs considered, but often not for intermediate ones, with the exception of $\classF_{\sem}$ and $\classF_{\FW}$.
Interestingly, the two known semantic operators in $\classF_{\sem}$ are not closed for $\classP_{\hor}$, while the syntactic one is, despite not being closed in general and thus not an operator according to \ref{def:forgOp}.
This can be remedied by defining an additional operator specific to Horn programs, which is a simplification of ${\sf forget_1}$ \cite{EiterW08} taking advantage of the fact that only one answer set exists. 
Note that '-' was used w.r.t. to the definitions of each $\classF$: the singleton classes $\classF_{\strong}$ and $\classF_{\weak}$ are precisely defined for normal programs; the intuition behind minimization embedded in $\classF_{\sem}$'s definition does not combine well with double negation; and, for $\classF_{\FS}$, $\classF_{\FW}$, and $\classF_{\SE}$, the consequence relation that is applied is defined for disjunctive rules.

\subsection{Classes of Operators}\label{subsec:classes}

The results in Table~\ref{fig:properties} provide us with valuable information to compare classes of operators, as well as some guidelines regarding the choice of a forgetting operator. 

The first concern is perhaps the required class of programs $\classP$.
If some class of operators is not closed or even not defined for $\classP$, then it is certainly not a good choice.
Of course, as already mentioned, nowadays, existing ASP solvers have no problem with accepting the full syntax of extended programs here considered, hence, there is no impediment in that regard.
Rather, if the application in question requires a certain class of programs, then this may discard certain classes of forgetting operators as possible choices.
Hence, many classes defined for extended programs may face difficulties if normal or disjunctive programs are required (cf.\ the occurrences of ``$\times$'' in Table~\ref{fig:properties}).
At the same time, $\classF_{\Sas}$ is clearly not suitable with its restriction to Horn programs, while $\classF_{\strong}$ and $\classF_{\weak}$ present considerable limitations if disjunctions are required,  since it is well-known that these cannot be represented in a strongly equivalent normal program; and $\classF_{\sem}$, $\classF_{\FS}$, and $\classF_{\FW}$ may require additional effort to represent double negation such as $a\la \nf\nf a$ by $a\la \nf aux$ and $aux \la \nf a$ using an additional auxiliar atom $aux$, where introducing new atoms to be able to forget others seems counterintuitive.

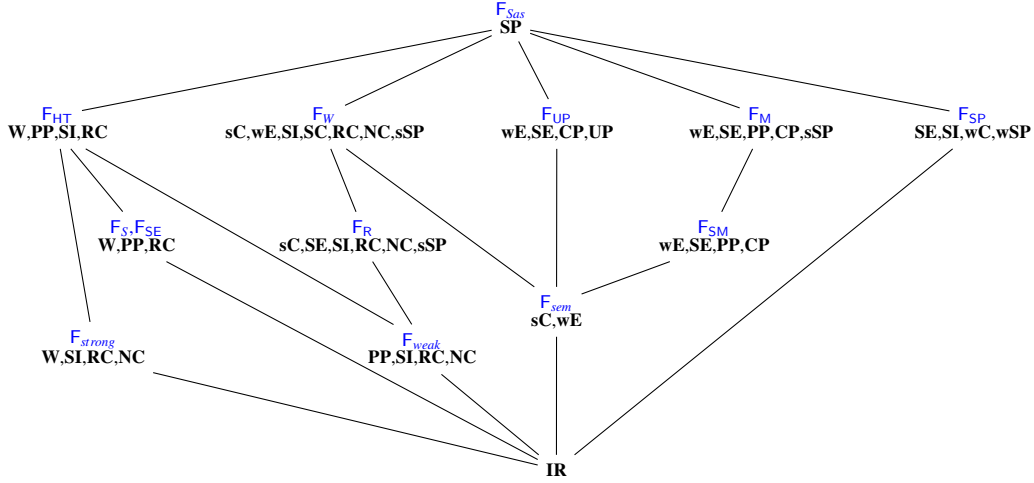
\begin{figure}[t]
\centering
\begin{tikzpicture}

\node(N0)     {\scriptsize{\spSP}};

\node(N1)       [below left = 1cm and 5cm of N0] {\scriptsize{\spW,\spPP,\spSI,\spRC}};
\node(N2)       [below left = 1cm and 0.8cm of N0] {\scriptsize{\spsC,\spwE,\spSI,\spSC,\spRC,\spNC,\spsSP}};
\node(N3)       [below right = 1cm  and -0.5cm of N0] {\scriptsize{\spwE,\spSE,\spCP,\spUP}};
\node(N4)       [below right = 1cm and 2cm of N0] {\scriptsize{\spwE,\spSE,\spPP,\spCP,\spsSP}};
\node(N5)       [below right = 1cm and 5cm of N0] {\scriptsize{\spSE,\spSI,\spwC,\spwSP}};

\node(N6)       [below left = 2.5cm and 4.1cm of N0] {\scriptsize{\spW,\spPP,\spRC}};
\node(N7)       [below left = 2.5cm and 0.5cm of N0] {\scriptsize{\spsC,\spSE,\spSI,\spRC,\spNC,\spsSP}};
\node(N8)       [below right = 2.5cm and 1.6cm of N0] {\scriptsize{\spwE,\spSE,\spPP,\spCP}};

\node(N9)       [below left = 4cm and 4.5cm of N0] {\scriptsize{\spW,\spSI,\spRC,\spNC}};
\node(N10)       [below left = 4cm and 0.1cm of N0] {\scriptsize{\spPP,\spSI,\spRC,\spNC}};
\node(N11)       [below right = 3.5cm and -0.1cm of N0] {\scriptsize{\spsC,\spwE}};

\node(N12)       [below right = 5.5cm and 0.1cm of N0]{\scriptsize{\spIR}};

\node(L0) [above=-0.2cm of N0]{\scriptsize\color{blue}$\classF_{\Sas}$};
\node(L1) [above=-0.2cm of N1]{\scriptsize\color{blue}$\classF_{\htF}$};
\node(L2) [above=-0.2cm of N2]{\scriptsize\color{blue}$\classF_{\FW}$};
\node(L3) [above=-0.2cm of N3]{\scriptsize\color{blue}$\classF_{\up}$};
\node(L4) [above=-0.2cm of N4]{\scriptsize\color{blue}$\classF_{\mF}$};
\node(L5) [above=-0.2cm of N5]{\scriptsize\color{blue}$\classF_{\spF}$};

\node(L6) [above=-0.2cm of N6]{\scriptsize\color{blue}$\classF_{\FS}$,$\classF_{\SE}$};
\node(L7) [above=-0.2cm of N7]{\scriptsize\color{blue}$\classF_{\rF}$};
\node(L8) [above=-0.2cm of N8]{\scriptsize\color{blue}$\classF_{\smF}$};

\node(L9) [above=-0.2cm of N9]{\scriptsize\color{blue}$\classF_{\strong}$};
\node(L10) [above=-0.2cm of N10]{\scriptsize\color{blue}$\classF_{\weak}$};
\node(L11) [above=-0.2cm of N11]{\scriptsize\color{blue}$\classF_{\sem}$};

\draw(N0) -- (L1);
\draw(N0) -- (L2);
\draw(N0) -- (L3);
\draw(N0) -- (L4);
\draw(N0) -- (L5);

\draw(N1) -- (L6);
\draw(N2) -- (L7);
\draw(N4) -- (L8);

\draw(N1) -- (L9);
\draw(N1) -- (L10);
\draw(N7) -- (L10);
\draw(N2) -- (L11);
\draw(N3) -- (L11);
\draw(N8) -- (L11);

\draw(N6) -- (N12);
\draw(N9) -- (N12);
\draw(N10) -- (N12);
\draw(N11) -- (N12);
\draw(N5) -- (N12);

\end{tikzpicture}
\caption{Sets of properties satisfied by known classes of forgetting operators, where each connection represents that the above set of properties implies the one below (with $\pSP$ and $\pW$ restricted to Horn programs).}
\label{fig:KnownInstances}
\end{figure}

With these considerations on \pE{\classP} in mind, we can analyze the remaining properties.
To that end, Fig.~\ref{fig:KnownInstances} presents a lattice of inclusions between the sets of properties satisfied by each known class of forgetting operators, taking into consideration the following:
\begin{itemize}
\item Properties on existence, i.e., \pE{\classP}, as well as \pPI\ are not considered in  this lattice. The reasons are that existence can be easily handled orthogonally, and the same is true for the few cases where \pPI\ does not hold. Moreover, this allows obtaining more interesting observations that would otherwise be obfuscated.
\item For the sake of readability, not all properties are made visible in each case. Rather, taking advantage of the results obtained in Thm.~\ref{prop:relations}, only the necessary ones are presented, those that are implied are left implicit. E.g., for $\classF_{\FW}$, which satisfies both \pSC\ and \pSE, only \pSC\ is shown, while \pSE\ is left implicit.
\item As $\classF_{\Sas}$ satisfies all the properties, it represents the top element of the lattice, in a certain sense the ideal case (even though this turned out to only be possible for Horn programs). To complement this, property \pIR\, which is true for any notion of forgetting, has been chosen as the bottom element.
\end{itemize}

Fig.~\ref{fig:KnownInstances} makes it apparent that there is one kind of property that divides the classes into two groups, namely whether some kind of relation between the answer sets of the original program and those of its result of forgetting holds or not.
The classes for which this is the case are $\classF_{\sem}$, $\classF_{\FW}$, $\classF_{\smF}$, $\classF_{\Sas}$, $\classF_{\spF}$, $\classF_{\rF}$, $\classF_{\mF}$, and $\classF_{\up}$ by either satisfying \psC\ or \pwC\ or a property that implies these.

Among them, putting $\classF_{Sas}$ aside, as it satisfies all the properties which turns out to not provide considerable insights, the two classes that satisfy \pCP\ and \pPP\ turn out to be separated by only one property as follows (using semi-formal notation):\footnote{Such equations are in fact not to be read as precise characterizations of classes, but rather a form of visualization of differences.}
\[
\classF_{\smF} + \psSP\ \rightsquigarrow \classF_{\mF}
\]
Hence, the difference between these two classes resides in the preservation of answer sets, no matter which rules are added (over the remaining atoms). 

Another such close relationship in terms of satisfied properties can be established between $\classF_{\FW}$ and $\classF_{\rF}$ as follows:

\[
\classF_{\rF} + \pSC\ + \pwE \rightsquigarrow \classF_{\FW}
\]
Thus, the difference between these two classes lies in strengthening property \pSE\ (using \pSC) and adding preserving equivalence while forgetting.

We can even take this one step further and state the following:
\[
\classF_{\rF} + \classF_{\sem} + \pSC\  \rightsquigarrow \classF_{\FW}
\]

However, note that picking an operator of $\classF_{\rF}$ and somehow enforcing \pSC\ and \pwE\ will not provide an operator of $\classF_{\FW}$. 
It can be shown that the effect of forgetting $V$ from $P$ for $\fgt\in \classF_{\FW}$ yields a result that replaces all $v\in V$ as if they were false, independently of the actual rules in $P$.
E.g., forgetting about $p$ from $p\la$ would yield $\bot$, which is not aligned with the original idea of forgetting, i.e., removing all $v\in V$ without affecting other derivations, nor with the idea of class $\classF_{\rF}$ in particular, and neither \pSC\ nor \pwE\ will change that.
In general, an operator of a superclass does not necessarily belong to a subclass in the hierarchy of Fig.~\ref{fig:KnownInstances}.
Still, there are cases where this is true, e.g., $\fgt\in\classF_{\up}$ naturally satisfies the defining condition of class $\classF_{\sem}$, and thus belongs to the class.

Also note that the three classes related to \pSP\, $\classF_{\spF}$, $\classF_{\rF}$, $\classF_{\mF}$, place themselves quite differently in this picture. 
Recall that these three classes, in the face of the result that it is not always possible to forget while satisfying \pSP\ \cite{GoncalvesKL-ECAI16}, are considered relaxations of the equivalence, stating that \pSP\ is composed of \psC, \pwC, and \pSI\ \cite{GoncalvesKLW20}.
Each of the three classes then basically is relaxed by dropping one of these three properties.
Given that these three properties imply different other properties (cf.\ Thm.~\ref{prop:relations}), this relation becomes less apparent in our lattice representation.
Still, some observations can be made.
$\classF_{\mF}$, which satisfies both \psC\ and \pwC, aligns well with approaches that preserve answer sets.
$\classF_{\rF}$, which satisfies both \psC\ and \pSI, fits within the classes that satisfy \pSI.
Finally, $\classF_{\spF}$, which satisfies both \pwC\ and \pSI, does not relate to any class other than $\classF_{\Sas}$, essentially because a) it is the only class that satisfies \pwC\ and not \psC, and b) it satisfies a set of properties distinct from any of the classes that satisfy \pCP.

Among the four classes that do not support preservation of answer sets, $\classF_{\strong}$ and $\classF_{\weak}$ are closely related due to their similar definition.
Both coincide on satisfying \pSI, a consequence of their syntactic definition that only manipulates rules containing the atoms to be forgotten, and differ on \pW\ and \pPP, a consequence of the different treatment of negated occurrences of the atoms to be forgotten.

For the third node without this preservation support, $\classF_{\FS}/\classF_{\SE}$, there is also a close proximity to $\classF_{\FW}$ based on their definition, yet, their characterizations differ substantially. 
Both satisfy \pSE\ and \pPP, but differ on six other properties, which means that in this case the variation in the definition has a much more profound effect on the set of satisfied properties.
At the same time, we already know that there is a close relation to the remaining class $\classF_{\htF}$ as witnessed in Thm.~\ref{prop:SEFequalsHTF}.
This is matched by a close correspondence in terms of satisfied properties.
\[
\classF_{\FS}/\classF_{\SE} + \pSI \rightsquigarrow \classF_{\htF} 
\]
Here, \pSI\ plays a distinguishing role.
Notably, this clarifies the apparent mismatch of the characterizations for $\classF_{\htF}$ and $\classF_{\SE}$ in terms of satisfied properties, both claiming that it is precisely given by \pIR, \pW, \pPP, and \pNP.
This is indeed true for each of them for the maximal class of programs considered, but, intuitively, restricting $\classF_{\htF}$ to $\classP_{\dis}$ cancels \pSI.

$\classF_{\htF}$ is also closely connected to $\classF_{\smF}$, as the latter restricts the HT-models of the result s.t.\ \pCP\ holds.
It turns out that this not only cancels \pW\ (see 1.\ of Thm.~\ref{prop:relations}), but also \pSI\ and four further properties related to HT-consequences.

Now, in the spirit of describing classes of forgetting operators, for several classes, a defining characterization in terms of satisfied properties has been introduced in the literature.
But providing operators that precisely satisfy only a certain set of properties can also be used to show that some sets of properties do not suffice to characterize a class:
\begin{itemize}
\item \pW, \pSI\ : delete all rules with atoms to be forgotten.
\end{itemize}
This operator matches the properties satisfied by Strong Forgetting, but it clearly does not fit into the class (even if we ignored that we defined the class $\classF_{\strong}$ as a singleton), since, by the way deletion is applied, the idea of (WGPPE) is lost.

We mention examples of further operators that can be defined in a similar style, providing evidence that certain sets of properties alone are probably of little interest as they are satisfied by absurd operators. 
In fact, several of them do not even correspond to the definition of a forgetting operator, as no semantic relations between the remaining atoms are preserved.
 
\begin{itemize}
\item \pIR\ : delete all rules; then add some arbitrary rules over the remaining alphabet (after forgetting);\footnote{The term ``arbitrary'' is used freely to represent some deterministic set, e.g., the first elements of some specific ordering.}

\item \pW\ : delete all rules with atoms to be forgotten and an arbitrary 50\% of the remaining rules;

\item \pPP\ : perform weak forgetting; in the resulting program, pick an arbitrary set of rules and turn them into facts by removing their body;

\item \pSE\ : compute all answer sets; remove all atoms to be forgotten from them; create a set of facts that represents the intersection of all such reduced answer sets;

\item \pSI\ : perform the WGPPE replacement step of strong and weak forgetting; in the resulting program, arbitrarily delete rules with negative occurrences (in the body) of atoms to be forgotten, or just remove their bodies, and delete all rules with positive occurrences (in the body or head) of atoms to be forgotten;

\item \pSE,\pW\ : delete the entire program; 

\item \pSE,\pPP\ : add, as facts, to the result of performing SE-forgetting, the atoms that belong to some
answer set of the original program and for which there is some rule 
of the original program that contains some negative and no positive occurrence of forgotten atoms.

\end{itemize}

Thus, often a meaningful choice of a class of forgetting operators requires looking at more than one property in combination with the rationale behind their definition, in particular for cases such as $\classF_{\sem}$ where the number of satisfied properties is comparably small.

Finally, while the results in Thm.~\ref{thm:propsOps} for \pE{\classP} naturally differentiate between the classes of programs $\classP$ considered, the remaining properties are stated for the most general class of programs for which the class of operators is defined.
This begs the question whether restricting $\classP$ would affect the results shown in Fig.~\ref{fig:properties}, which we will now address to gain further insights.

We first consider Horn programs, for which we already know that all classes of operators are closed.
From Thm.~\ref{prop:coincideHorn}, we know that several classes that satisfy different sets of properties in the general case actually coincide.
As expected, these classes all satisfy the same set of properties when restricted to Horn programs.
The reason can be traced back to the incompatibility result \ref{i24}.\ in Thm.~\ref{prop:relations}, only stated for program classes above $\classP_{\hor}$, i.e., it does not apply here.

\begin{theorem}\label{thm:allProperties}
For Horn programs, the following holds:
\begin{itemize}

\item $\classF_{\strong}$, $\classF_{\weak}$, $\classF_{\FS}$, $\classF_{\htF}$, $\classF_{\smF}$, $\classF_{\Sas}$, $\classF_{\SE}$, $\classF_{\spF}$, $\classF_{\rF}$, $\classF_{\mF}$, and $\classF_{\up}$ satisfy \pW, \pRC, \pSP, and \pPI;

\item $\classF_{\sem}$ satisfies  \pCP\ and \pPI;

\item $\classF_{\FW}$ satisfies the same properties as in the general case.

\end{itemize}
\end{theorem}
\noindent
Actually, only the minimally necessary properties are mentioned, the remaining can be obtained from Thm.~\ref{prop:relations}.
This means, in particular, that the first group of classes satisfies all the presented properties (with \pE{\classP} limited to \pE{\hor}, of course), and, that reducing to Horn programs does not change the set of properties satisfied by $\classF_{\FW}$.

For normal programs, it turns out that the introduction of negation in the body immediately makes the result coincide with the general one for all the classes (except for $\classF_{Sas}$ which is not defined for such classes).
This is witnessed by the fact that all counterexamples are normal programs, in particular those mentioned in the previous section.

\subsection{On desirability of classes of operators}\label{subsec:desirable}

The plethora of results presented so far spread out over the previous sections, and the many differentiating details discussed with respect to these make it difficult to assess which classes of operators are in the end more important.
To aid the reader in that regard, we present some summarizing considerations, guided mainly by the definition of forgetting operators, i.e., the idea that the semantic relations between the remaining atoms be preserved.

We start by identifying several classes of forgetting operators which in our view are less important, and can be dismissed for the remaining discussion.
We note that in listing these, we follow the chronological order only, without any preference associated to this order.

\begin{itemize}
	\item $\classF_{\strong}$ and $\classF_{\weak}$ can be dismissed, as their semantic consequence relation is non-standard (T-equivalence), resulting in forgetting results that do not even align with answer sets. Moreover, they are only defined for normal programs which severely limits the application.
	
	\item $\classF_{\sem}$ can be dismissed, as the characterization based on answer sets only is semantically too weak, not even aligning with expected results for Horn programs in general. It is true that one can find concrete operators that overcome part of these problems, but not in general.
	
	\item $\classF_{\FW}$ can be dismissed because, even though the class satisfies many properties, it cannot coincide with intended results even for Horn programs, i.e., the semantic relations preserved over the remaining atoms are not aligned with our expectations.
	
	\item $\classF_{\Sas}$ can be dismissed, as it is only defined for Horn programs, which does not fit the intention of forgetting in ASP, where default negation is fundamental. The interest in the approach lies rather in the introduction of the idea of \pSP\ itself.
	
\end{itemize}

The remaining seven classes, $\classF_{\FS}$, $\classF_{\htF}$, $\classF_{\smF}$, $\classF_{\spF}$, $\classF_{\rF}$, $\classF_{\mF}$, and $\classF_{\up}$, all coincide for Horn programs, so further considerations are needed.
If we look for a consensus among the satisfied properties, then we note that \pPP\ and \pSE\ are basically satisfied by all of them (with the exception of $\classF_{\up}$ which is justified by the fact that this approach focusses on uniform equivalence).
However, these two properties alone do not suffice because, as we have seen, corresponding absurd operators can be defined.
To facilitate a comparison, we can separate them according to how semantic relations between the remaining atoms are preserved.

\paragraph{Preserving strong equivalence}
This is the strongest form of preservation of semantic relations, as it aims to approximate satisfying \pSP, and the classes $\classF_{\spF}$, $\classF_{\rF}$, and $\classF_{\mF}$ correspond to this.
A preference among these three strongly depends on which of the individual distinguishing characteristics one prefers.
This relies mainly on the satisfied properties, as the computational complexity and the classes of programs to which these can be applied do coincide. 
These differences have been spelled out with much detail by Gon\c calves et al.~\shortcite{GoncalvesKLW20} and we refer the reader to this paper.

\paragraph{Preserving uniform equivalence}
This is a weaker form of preserving the dependencies compared to the previous one, and aligned with \pUP.
The class $\classF_{\up}$ belongs to this category.
The main benefit over the previous ones is that \pUP\ is indeed satisfied and that individual atoms can be forgotten in any order without leading to a different result. It is also well-aligned with ASP when we only want to vary the instance data, i.e., the facts. 

\paragraph{Preserving equivalence}
This is again weaker than the previous ones, and aligned with \pCP.
Only $\classF_{\smF}$ fits this category among the remaining.
While the preservation of relations over the remaining atoms is weaker, as noted in the previous section in particular w.r.t.\ $\classF_{\mF}$, it comes in exchange with a lower computational complexity.

\paragraph{Preserving consequences}
These approaches preserve the semantic consequences over the remaining atoms, including the classes $\classF_{\FS}$ and $\classF_{\htF}$.
While a comparison with the previous ones w.r.t.\ the strength of preserving semantic relations from the original program is not straightforward, we argue that, since the definition of $\classF_{\smF}$ imposes a further restriction on that of $\classF_{\htF}$, preserving consequences is weaker.
Among the two classes, as shown by several technical results in this paper, $\classF_{\htF}$ is preferable.

While this exposition indicates an order of preference among these classes, it is not strict, and still depends on the concrete intended usage.
To aid in that regard, we next discuss with more detail the relation to uniform interpolation, before we consider applications.

\subsection{Forgetting and interpolation}\label{subsec:interpol}

A strong notion of interpolation, called \emph{uniform interpolation}, is well-known to be closely related to forgetting \cite{ZZ-AI09,LutzW11,GabbayPV11}.
Gabbay et al. \shortcite{GabbayPV11} introduced the notion for the case of logic programs, which we here adapt to make it more concise.
\begin{definition}
A class of logic programs $\classP$ is said to have the uniform interpolation property with respect to a consequence relation $\vdash$, if for every program $P\in \classP$ and set $V\subseteq\sign$ of atoms, there exists a program $P^*\in\classP$ with $\sign(P^*)\subseteq \sign\setm V$, such that:
\begin{itemize}
 
\item[$(i)$] $P\vdash P^*$;

\item[$(ii)$] $P^*\vdash R$, for every $R\in\classP$ with $\sign(R)\subseteq \sign\setm V$ such that $P\vdash R$.

\end{itemize}
In this case, $P^*$ is said to be a \emph{uniform interpolant} of $P$ with respect to $V$.  
\end{definition}

We say that a forgetting operator $\fgt$ over a class $\classP$ of logic programs can be used to obtain uniform interpolants, if $\f{P}{V}$ is a uniform interpolant of $P$ with respect to $V$, for every $P\in \classP$ and $V\subseteq \sign\setm V$.
Gabbay et al. \shortcite{GabbayPV11} considered the non-monotonic skeptical consequence relation $\vsim$ for logic programs, defined as $P\vsim P'$ if $M\models P'$, for every $M\in\as{P}$.
Using a forgetting operator in $\classF_{\sem}$, as defined by Eiter and Wang~\shortcite{EiterW08}, Gabbay et al. showed that, with respect to the consequence relation $\vsim$, the class of disjunctive programs satisfies uniform interpolation for programs $R$ composed of facts. 
Later, Wang et al.~\shortcite{WangWZ13} extended this result by showing that, with respect to the consequence relation $\vsim$, the class of all logic programs satisfies uniform interpolation. 
Besides knowing that the class $\classP_e$ of all logic programs satisfies uniform interpolation, it is also worth determining which approaches of forgetting can be used to obtain such uniform interpolants. 
It turns out that satisfying \pCP\ is a sufficient condition to obtain uniform interpolants with respect to $\vsim$, and in fact the two inclusions of \pCP, i.e., \pwC\ and \psC, imply, respectively, conditions $(i)$ and $(ii)$ of the definition of uniform interpolation.  

\begin{theorem}\label{thm:CPimpliesASInterpolation}
If a class $\classF$ of operators over a class $\classP$ of logic programs satisfies property \pCP, then every operator of that class can be used to obtain uniform interpolants w.r.t.~$\vsim$.
\end{theorem}

Given the above result, we know that every forgetting operator satisfying \pCP\ can be used to obtain uniform interpolants, namely the classes $\classF_{\smF}$, $\classF_{\mF}$, and $\classF_{\up}$.
Class $\classF_{\Sas}$ also satisfies \pCP, but this case is not worth considering, because the class is only defined for Horn programs.
From the remaining classes that do not satisfy \pCP, only $\classF_{\rF}$ satisfies the conditions of uniform interpolation.   
\begin{theorem}\label{thm:classesSatASInterpolation}
Every forgetting operator of the classes $\classF_{\smF}$, $\classF_{\mF}$, $\classF_{\up}$, and $\classF_{\rF}$ can be used to obtain uniform interpolants with respect to $\vsim$. 
 \end{theorem}

None of the other classes of forgetting operators can be used to obtain uniform interpolants with respect to $\vsim$, and we now present, for each of these classes, a counterexample showing that one of the conditions for uniform interpolation is not satisfied.
As shown by Gabbay et al. \shortcite{GabbayPV11}, the class $\classF_{\sem}$ satisfies the conditions of uniform interpolation when $R$ is a set of facts. Although condition $(ii)$ is satisfied in general, we can see that this is not the case for condition $(i)$. Consider the program $P=\{a\la p, a\la b, p\la \nf b,b\la\nf p\}$ over $\sign=\{a,b,c,p\}$, where $\as{P}=\{\{a,p\},\{a,b\}\}$. Then, since the only restriction that any $\fgt \in \classF_{\sem}$ has to satisfy is $\as{\f{P}{p}}=\{\{a\}\}$, we can consider $\f{P}{p}=\{a\la, c\la b\}$. In this case $P\not\vsim \f{P}{p}$, showing that condition $(i)$ of uniform interpolation is not satisfied.

In the case of $\classF_{\weak}$, consider the program $P=\{a\la \nf b, b\la \nf a\}$, for which $\as{P}=\{\{a\},\{b\}\}$. In this case, $\fgt_{\weak}(P,a)=\{b\la\}$. But since $\{a\}\not\models \fgt_{\weak}(P,a)$, we have that $P\not\vsim \fgt_{\weak}(P,a)$, and therefore condition $(i)$ of uniform interpolation is not satisfied. 
For the classes $\classF_{\strong}$ and $\classF_{\FS}$, consider the program $P=\{b\la \nf a\}$, where $\as{P}=\{\{b\}\}$. 
In this case, for $\fgt\in (\classF_{\strong}\cup\classF_{\FS})$, we have that $\fgt(P,a)\htequiv\{\}$, and therefore $ \as{\fgt(P,a)}=\{\{\}\}$. 
If we take $R=\{b\la\}$, then $P\vsim R$, but $\fgt(P,a)\not\vsim R$, which means that condition $(ii)$ of uniform interpolation is not satisfied.
 
For $\classF_{\FW}$, consider the program $P=\{b\la, a\la b\}$, where $\as{P}=\{\{a,b\}\}$. In this case, for $\fgt\in \classF_{\FW}$, we have $\fgt(P,a)=\{b\la, \bot\la b\}$, and therefore $\as{\fgt(P,a)}=\{\}$. But since $\{a,p\}\not\models \fgt(P,a)$, we have that $P\not\vsim \fgt(P,a)$, and therefore condition $(i)$ of uniform interpolation is not satisfied. 

If we now consider $\classF_{\htF}$, we can take a program $P$ such that $\HT(P)=\{\tuple{ab,ab},\tuple{b,ab}\}$. Therefore, $\as{P}=\{\}$. In this case, for $\fgt\in \classF_{\htF}$, we have $\HT(\fgt(P,a))=\{\tuple{b,b}\}$, and therefore $\as{\fgt(P,a)}=\{\{b\}\}$. If we take $R=\{\bot\la \nf b\}$, then $P\vsim R$, but $\fgt(P,a)\not\vsim R$, which means that condition $(ii)$ of uniform interpolation is not satisfied. 

Finally, in the case of $\classF_{\spF}$, consider the program $P=\{a\la p, b\la \nf p, p\la \nf\nf p\}$, for which $\as{P}=\{\{a,p\},\{b\}\}$. In this case, for $\fgt\in \classF_{\spF}$, we have $\as{\fgt(P,p)}=\{\{a\},\{b\},\{a,b\}\}$. If we take $R=\{\bot\la a, b\}$, then $P\vsim R$, but $\fgt(P,a)\not\vsim R$, which means that condition $(ii)$ of uniform interpolation is not satisfied. 

All these classes of operators that cannot be used to obtain uniform interpolants share the fact that property \pCP\ is not satisfied. Although this may indicate that \pCP\ could be a necessary condition for uniform interpolation with respect to $\vsim$, 
a simple counterexample shows that this is not the case, and that, therefore, \pCP\ is in fact strictly stronger than uniform interpolation. 

\begin{example}
Consider the program $P=\{a\la\nf b,b\la\nf a, c\la a, c\la b\}$, where $\as{P}=\{\{a,c\},\{b,c\}\}$,  and a forgetting operator $\fgt$ such that $\f{P}{a}=\{c\la\}$. Then, it is clear that $\fgt$ cannot satisfy \pCP. Nevertheless, it satisfies the two conditions of uniform interpolation, and it could therefore, be taken as a uniform interpolant for $P$. 
\end{example}
This also reveals that uniform interpolation with respect to the skeptical consequence $\vsim$ is not well-aligned with forgetting, as it does not impose a strong connection between the answer sets of the original program and those of the result of forgetting. This is in part due to the fact that the consequence $\vsim$ only imposes the preservation of skeptical consequence.

Hence, instead of using the weak skeptical consequence, $\vsim$, as originally considered by Gabbay et al. \shortcite{GabbayPV11}, HT-consequence, $\htmodels$, is arguably a more suitable consequence relation for uniform interpolation with respect to logic programs. 
In this case, conditions $(i)$ and $(ii)$ of uniform interpolation with respect to HT-consequence match two of the properties considered for forgetting, namely \pW\ and \pPP, respectively. 

\begin{theorem}\label{thm:WPPcharacterizInterpolation}
A class $\classF$ of forgetting operators can be used to obtain uniform interpolants w.r.t.\ $\htmodels$ iff $\classF$ satisfies both \pW\ and \pPP.
\end{theorem}

The above characterization, together with the results of Thm.~\ref{thm:propsOps}, allow us to conclude exactly what classes of operators can be used to obtain uniform interpolants with respect to $\htmodels$.

\begin{theorem}\label{thm:opsSatInterpolationHT}
 Every forgetting operator of the classes $\classF_{\htF}$ and $\classF_{\FS}$ can be used to obtain uniform interpolants with respect to $\htmodels$. 
\end{theorem}

The classes $\classF_{\htF}$ and $\classF_{\FS}$ are therefore closely aligned with uniform interpolation, as both satisfy \pW\ and \pPP. 
In fact, as stated in Thm. (10) of~\cite{WangZZZ14}, the class $\classF_{\htF}$ is precisely characterized by \pW\ and \pPP, which also means that it precisely characterizes uniform interpolation with respect to $\htmodels$. 
As pointed out by Wang et al., this result also shows that the class of all logic programs has the uniform interpolation property with respect to HT-consequence. 
Note that, when considering the class of all logic programs, we should consider $\classF_{\htF}$ to obtain uniform interpolants, since $\classF_{\FS}$ is only defined for disjunctive programs. 

The above results and discussion also show that, contrarily to forgetting in the realm of monotonic logics, e.g. classical logic or description logics, where uniform interpolation can serve as a guideline for what forgetting should be, in our view, in non-monotonic frameworks such as ASP this cannot be the case: 
the relation $\vsim$ seems too weak to capture forgetting and $\htmodels$ imposes \pW, a monotonic property which is incompatible with the preservation of answer sets when forgetting.  

\subsection{Applications}\label{subsec:apps}

In this section, we will discuss some of the applications of forgetting mentioned in the literature and provide indications as to what classes of operators are, from our point of view, most suitable for each of them.

\paragraph{Conflict Resolution}
Forgetting has been considered as a means of conflict resolution for inconsistencies by weakening pieces of information to restore consistency in propositional logic, also termed recovery of preferences \cite{LangM02,LangM10}.
This has been adapted to a multi-agent setting using answer set programming \cite{ZhangF06,EiterW08,DelgrandeW15}, where a set of programs represent the knowledge or preferences of individual agents.
If these programs together are inconsistent, then a compromise is a sequence of sets of atoms -- one set per program indicating that these are to be forgotten from the corresponding program -- such that the resulting programs together are consistent, admitting an agreement, i.e., one or several answer sets.
Zhang and Foo \shortcite{ZhangF06} defined compromises (termed preferred solutions) that are minimal w.r.t.\ the amount of atoms forgotten, and Eiter and Wang \shortcite{EiterW08} argued that, in such a setting, answer sets indeed result in minimal agreements (unlike in propositional logic where non-minimal agreements may be obtained).
Finally, Delgrande and Wang \shortcite{DelgrandeW15} emphasized that such forgetting results should clearly keep as much of the program as is while forgetting, in particular not affecting rules over the remaining atoms. 
Thus, in our view, an approach that preserves as much as possible the semantics over the remaining atoms is recommended according to our considerations in Sect.~\ref{subsec:desirable}, in particular, one that preserves these semantic relations when other rules are added from the programs of the other agents, i.e., when using one of the classes $\classF_{\spF}$, $\classF_{\rF}$, and $\classF_{\mF}$.

\paragraph{Query Answering}
As argued by Delgrande and Wang \shortcite{DelgrandeW15}, in query answering, if one can determine what is relevant to a query, then the irrelevant part of the knowledge base/logic program can be forgotten, allowing for more efficient querying.
Similar ideas have been expressed in the context of forgetting in Description Logics for using only a small fraction of the ontology in an application \cite{KonevWW09}.
Conceptually common to these ideas is that one wants to use a simplified version of the knowledge base in question ``as is'', i.e., without having to incorporate additional information during the reasoning process.
This is indeed closely related to uniform interpolation.
Therefore, in our view, the most suitable solutions can be found among the approaches that can be used to obtain uniform interpolants with respect to $\htmodels$, i.e., $\classF_{\htF}$ and $\classF_{\FS}$, among which, according to our considerations in Sect.~\ref{subsec:interpol}, we prefer to recommend $\classF_{\htF}$, since it is more general and satisfies \pSI\ which allows us to simplify the process of computing forgetting results.

\paragraph{Modular Reuse}
While similar in spirit to the ideas presented with respect to query answering, a crucial difference for modular reuse is that a simplified program is created via forgetting that, unlike the former, is still subject to interaction with rules from other programs.
The answer to the question what approach is preferred then largely depends on how this interaction is realized.
If arbitrary rules can be added or interact with the program, as, e.g., outlined for the case of conflict resolution where a joint program is composed, then the same recommmendations apply as outlined for conflict resolution.
If however the interaction follows a truly modular approach with well-defined input-output interfaces between modules \cite{JanhunenOTW09}, where interaction is limited to atoms, then a particular preference can be given to $\classF_{\up}$ whose characteristics are well-suited for this setting.

\paragraph{Hiding/Privacy}
The principal idea of hiding is that certain parts/terms of a program are not intended to be public.
This usage is also aligned with the notion of privacy and the data protection regulations such as the GDPR \cite{eu:gdpr}.
Here, the recommendation on what class of forgetting operators to use depends on the intended usage.
If we aim to merely create a publicly viewable version of a program, then any approach that can be used to obtain uniform interpolants is suitable.
Thus, as spelled out for Query Answering, $\classF_{\htF}$ would be most suitable.
However, since this class does not come with a known syntactic operator, among the two, $\classF_{\FS}$ is preferable in this context.
Alternatively, if the resulting program is meant to be a public version that is being used, possibly together with other programs, then the recommendations from Modular Reuse are more suitable.

\paragraph{Embeddings in Forgetting}
Forgetting has been considered in the literature as a way to capture other problems with the aim to facilitate their study and comparison, as well as being able to re-use existing algorithms.
For example, Zhang and Foo \shortcite{ZhangF06} have studied how different approaches to updating logic programs can be embedded into their framework of solving conflicts which is built on forgetting, allowing then to analyze these different frameworks.
Similarly, Eiter and Wang \shortcite{EiterW08} showed how inheritance logic programs \cite{BuccafurriFL02} can be captured using forgetting, which in turn could be used to capture updates in logic programs.
In these cases, picking a suitable approach depends to a considerable extent on the target, so a more detailed inspection on the approaches and what properties they satisfy is recommended (cf.~Sect.~6.1).

Summing up, in our view, those classes of operators that are closely related with \pSP\ and \pUP, i.e.,  $\classF_{\spF}$, $\classF_{\rF}$, $\classF_{\mF}$, and $\classF_{\up}$, are in general the most useful, except for those applications that are closer related to uniform interpolation, in which case $\classF_{\htF}$ would be more suitable.

\section{Conclusions}\label{sec:concl}

The landscape of forgetting in ASP comprises many operators and classes of operators defined to obey some subset of a large number of \emph{desirable} properties proposed in the literature, while lacking a systematic account of all these results and their relations, making it all too difficult to get a clear understanding of the state-of-the-art, and even to choose the most adequate operator for some specific application.

This paper aimed at addressing this problem, presenting a systematic study of forgetting in ASP, including a thorough investigation of both properties and existing classes of forgetting operators, going well beyond a survey of the state-of-the-art as many novel results were included, thus achieving a truly comprehensive picture of the landscape. 

In more detail, after providing a uniform definition of forgetting, we presented a detailed account on properties of forgetting found in the literature, that allowed us to establish existing relations between these properties independently of specific classes of forgetting operators.

We then recalled the classes of forgetting operators proposed in the literature in a systematic and uniform way, which allowed us to obtain several relations between them, including that some classes of forgetting operators coincide for restricted classes of programs, and that two of them even coincide in general.
This was strengthened by complementary considerations on existing concrete forgetting operators and results on computational complexity so far spread out over the literature.

We also provided a complete study showing which properties are satisfied by which class of operators. 
This allowed a thorough discussion and comparison of the existing properties, their impact on comparisons between classes of forgetting operators, as well as considerations on unsuitable operators of forgetting, all contributing to guidelines for the choice of a concrete forgetting operator. 

Regarding such choice, it clearly depends on the application in mind, and we discussed several applications and indicated options of forgetting operators to consider.
As a brief summary, the following criteria are important in our view:
\begin{itemize}
\item Preservation of semantic relations: Arguably, \pSP\ best captures the notion of forgetting, but as this is in general impossible for classes of programs beyond Horn, approximations are in order, which requires a more detailed look into the satisfied properties.
\item Concrete Operators: Preferably operators should provide results that are as similar as possible to the given program with the aim to preserve the declarative nature of ASP.
\item Complexity: Even though forgetting is known to be computationally expensive, a lower computational complexity is preferable. 
\item Program Class: Depending on the application, certain operators may not be suitable.
\end{itemize}
As our study shows, optimizing all criteria simultaneously, is not suitable.
E.g., the classes that provide closer approximations of \pSP\ are computationally more expensive, and simple concrete operators may only be defined for a restricted class of programs.
Still, we believe that our results together with the considerations on suitable options for certain use cases provide substantial material to help making a choice that balances these criteria.

One important open issue is establishing further concrete operators that are efficient (within the theoretical limitations imposed) and preferably provide forgetting results that are as similar as possible to the original program to preserve declarative nature of the programs, in the line of work by Berthold et al.~\shortcite{BertholdGKL19}.
Other avenues for future research include investigating the potential connections between forgetting and the work on abstraction in the context of Answer Set Programs~\cite{SaribaturE18,SaribaturSE19,SaribaturE21,SaribaturES21}.
Alternatively, we may further pursue forgetting over extensions of the syntax of logic programs, in the line of the work by Aguado et al.~\shortcite{AguadoCFPPV19}, where an extension of the syntax of programs by a new connective, called fork, allows Strong Persistence to always hold (cf.\ \cite{BertholdGKL19b}).
Also interesting is to study forgetting with semantics other than ASP, such as the FLP-semantics \cite{Truszczynski10} following the work by Wang et al.~\shortcite{WangZZZ14}, or the well-founded semantics \cite{GelderRS91} as considered by Knorr and Alferes \shortcite{AlferesKW13,KnorrA14}.
Finally, one ambitious open problem is the generalization to non-ground programs with variables to be able to forget from ASP programs expressed in first-order terms allowing then to forget predicates or constants. 

\paragraph{Acknowledgments}
We thank the anonymous reviewers for their helpful comments.
This work was partially supported by FCT project FORGET ({PTDC}/{CCI-INF}/{32219}/{2017}) and by FCT project NOVA LINCS ({UIDB}/{04516}/{2020}).

\bibliographystyle{acmtrans}
\bibliography{bib}

\newpage
\pagenumbering{arabic}
\section*{Appendix}


In this section, we prove the results stated in the paper. 
For Theorem~\ref{thm:propsOps}, we divide it into several insightful results, grouped by the classes of operators, which we present before the theorem. These intermediate results are stated as theorems (Theorems 11-23), and to visually distinguish them from the theorems stated in the paper, we do not use bold face to write the theorem keyword and number.
For the sake of simplicity and in order to highlight the important results, we prove some more technical lemmas only subsequently. 

\begin{theorem*}{prop:relations}
The following relations hold for all $\classF$:
 
\begin{enumerate}


\item \pW\ is equivalent to \pNP;

\item \pSP\ implies \pSE;

\item \pCP\ and \pSI\ together are equivalent to \pSP; 

\item  \psC\ and \pwC\ together are equivalent to \pCP;

\item \pCP\ implies \pwE;

\item \pSE\ and \pSI\ together imply \pPP;

\item \pwSP\ and \psSP\ together are equivalent to \pSP;

\item \psC\ and \pSI\ together imply \psSP;

\item \pwC\ and \pSI\ together imply \pwSP;

\item  \pW\ and \pPP\ together imply \pSC;

\item \pSC\ implies \pSE;

\item \pW\ implies \pNC;

\item \pwC\ is incompatible with \pW\ for $\classF$ over $\classP$ such that $\classP_{\nor}\subseteq \classP$;

\item \pwC\ and \pUI\ together are incompatible with \pRC\ for $\classF$ over $\classP$ such that $\classP_{\nor}\subseteq \classP$.


\item \pSI\ implies \pUI;

\item \pCP\ and \pUI\ together are equivalent to \pUP;

\item \pSI\ implies $\pSIu$;

\item \pUP\ is incompatible with $\pSIu$.

\end{enumerate}

\end{theorem*}
\begin{proof}
 The first two results were already proven in the literature, so we focus on the remainder. 
 

To show~\ref{i7}., first, let $\classF$ be a class of operators satisfying \pSP.  Let $\fgt\in \classF$, $V\subseteq \sign$, and $P$ a program.
The fact that $\classF$ satisfies \pCP\ follows immediately from \pSP, by taking $R=\emptyset$. 
The proof that \pSP\ implies \pSI\ involves reasoning with subsignatures. 
For a program $P'$ over $\sign\setminus V$, we denote by $\HT_{\parallel V}(P)$ the set of HT-models of $P$ over signature $\sign\setminus V$.
Let $R\in \classP$ with $\sign(R)\subseteq \sign\setminus V$. To show \pSI\ we aim to prove that $\f{P}{V}\cup R \htequiv \f{P\cup R}{V}$, i.e., $\HT(\f{P}{V}\cup R)=\HT( \f{P\cup R}{V})$.
We first prove that $\f{P}{V}\cup R \htequiv \f{P\cup R}{V}$ for the restricted signature $\sign\setminus V$, i.e., $\HT_{\parallel V}(\f{P}{V}\cup R)=\HT_{\parallel V}( \f{P\cup R}{V})$.
Let $R'$ be a program such that $\sign(R')\subseteq \sign\setminus V$. Then, by \pSP, we have $\as{\f{P}{V}\cup R\cup R'}= \as{P\cup R\cup R'}_{\parallel V}$. Also by \pSP, we have $\as{\f{P\cup R}{V}\cup R'}= \as{P\cup R\cup R'}_{\parallel V}$. Therefore,  $\as{\f{P}{V}\cup R\cup R'}=\as{\f{P\cup R}{V}\cup R'}$. This means that $\f{P}{V}\cup R \htequiv \f{P\cup R}{V}$ for the restricted signature $\sign\setminus V$, i.e., $\HT_{\parallel V}(\f{P}{V}\cup R)=\HT_{\parallel V}( \f{P\cup R}{V})$.
Now let $M$ be an HT-interpretation over $\sign$. Then, since $\f{P}{V}\cup R$ is a program over $\sign\setminus V$, we have that $M\htmodels \f{P}{V}\cup R$ iff $M_{\parallel V}\htmodels \f{P}{V}\cup R$. Since $\HT_{\parallel V}(\f{P}{V}\cup R)=\HT_{\parallel V}( \f{P\cup R}{V})$, we have that $M_{\parallel V}\htmodels \f{P}{V}\cup R$ iff $M_{\parallel V}\htmodels  \f{P\cup R}{V}$. Since $\f{P\cup R}{V}$ is a program over $\sign\setminus V$, we know that $M_{\parallel V}\htmodels  \f{P\cup R}{V}$ iff $M\htmodels  \f{P\cup R}{V}$. Therefore, $\HT(\f{P}{V}\cup R)=\HT( \f{P\cup R}{V})$. 

Now let $\classF$ be a class of operators satisfying \pCP\ and \pSI. Let $\fgt\in \classF$, $V\subseteq \sign$, $P$ a program, and $R$ a program with $\sign(R)\subseteq \sign\setminus V$.
Using \pSI\ we have $\as{\f{P}{V}\cup R}=\as{\f{P\cup R}{V}}$. Using \pCP\ we have $\as{\f{P\cup R}{V}}=\as{P\cup R}_{\parallel V}$. Putting these together we have $\as{\f{P}{V}\cup R}=\as{P\cup R}_{\parallel V}$. Therefore, \pSP\ holds.

Result ~\ref{i8}. follows immediately from the fact that \psC\ and \pwC\ correspond to the two inclusions on the equality condition in \pCP.

To show~\ref{i9}., let $\classF$ be a class of operators satisfying \pCP. Let $\fgt\in \classF$, $V\subseteq \sign$, and $P_1,P_2$ be two programs such that $\as{P_1}=\as{P_2}$. By \pCP, we have the equalities:  $\as{\f{P_1}{V}}=\as{P_1}\setminus V=\as{P_2}\setminus V=\as{\f{P_2}{V}}$. Thus, $\as{\f{P_1}{V}}=\as{\f{P_2}{V}}$.

To show~\ref{i10}., let $\classF$ be a class of operators satisfying \pSE\ and \pSI.
 Let $\fgt\in \classF$, $V\subseteq \sign$, and $P$ a program.
 Consider $R=\{r:P\htmodels r \text{ and } r \text{ does not contain variables from } V\}$. Then clearly $\f{P}{V}\cup R\htmodels R$. By \pSI, we have that $\f{P}{V}\cup R\htequiv \f{P\cup R}{V}$. We can then conclude that $\f{P\cup R}{V}\htmodels R$. Now, by \pSE\ and the fact that $P\cup R\htequiv P$, we can conclude that $\f{P\cup R}{V}\htequiv \f{P}{V}$. Therefore, $\f{P}{V}\htmodels R$, i.e., $\f{P}{V}\htmodels r$ for every rule $r$ such that $P\htmodels r$ and $r$ does not contain $V$. Thus \pPP\ is satisfied.
 
Result \ref{i16}.\ can be shown straightforwardly due to the fact that the conditions of \psSP\ and \pwSP\ provide the two directions of the condition of \pSP.

To show \ref{i17}., let $\fgt\in \classF$, $P\in \classP(\fgt)$ and $V\subseteq \sign$ such that $\classF$ satisfies \psC\ and \pSI. 
Consider $M\in \as{\f{P}{V}\cup R}$.
By \pSI, we have that $M\in \as{\f{P\cup R}{V}}$.
By \psC, we obtain $M\in \as{P\cup R}_{\parallel V}$ which finishes the proof.

To show \ref{i18}., let $\fgt\in \classF$, $P\in \classP(\fgt)$ and $V\subseteq \sign$ such that $\classF$ satisfies \pwC\ and \pSI. 
Consider $M\in \as{P\cup R}_{\parallel V}$.
By \pwC, we obtain $M\in\as{\f{P\cup R}{V}}$.
Then, by \pSI, we have $M\in\as{\f{P}{V}\cup R}$ which finishes the proof.
 
Results \ref{i11}.--\ref{i13}. have been shown by Gon\c calves et al.~\shortcite{GKL17}.

To show \ref{i24}.\ and \ref{i25}., we rely on the following example which fits the class of programs required.
Consider $P=\{a\la p; p\la \nf \nf p\}$ from which we want to forget about $p$.
Note that though the program uses double negation, we can easily replace $p\la \nf\nf p$ by two rules $p\la \nf q$ and $q\la \nf p$ and forget both $p$ and $q$.
To ease the presentation, we rely on $P$.
The idea is now to show that in both cases, it is not possible to satisfy both (sets of) properties simultaneously.
Note that $P$ has two answer sets, $\emptyset$ and $\{a,p\}$.

Consider first \ref{i24}., and suppose both \pwC\ and \pW\ are satisfied.
By \pwC, we have that $\as{\f{P}{\{p\}}}\supseteq \{\emptyset,\{a\}\}$.
But there is only one rule over $a$ that has (at least) these two answer sets: $a\la \nf \nf a$.
Now, $P$ has an HT-model $\langle \emptyset,a\rangle$, while a forgetting result $\f{P}{\{p\}}$ cannot have this HT-model.
We obtain a contradiction to \pW\ being satisfied.

Now consider \ref{i25}., and suppose that \pwC, \pUI, and \pRC\ are satisfied.
Again, by \pwC, we have that $\as{\f{P}{\{p\}}}\supseteq \{\emptyset,\{a\}\}$, and there is only one rule over $a$ that has (at least) these two answer sets: $a\la \nf \nf a$. 
Let this rule be $r$ in \pRC.
Clearly, $\f{P}{\{p\}}\htmodels r$.
Consider $r'$ such that $P\htmodels r'$.
Note that thus $r'$ cannot be $r$ itself.
By \pUI, we have $\f{\{r'\}}{\{p\}}\cup \{a\la\} \Sequiv \f{\{r'\}\cup \{a\la\}}{\{p\}}$.
We consider two cases.
First, $\f{\{r'\}}{\{p\}}$ has no HT-model $\langle \emptyset,a\rangle$.
Then, it cannot have an HT-model $\langle a,a\rangle$.
We derive a contradiction to the condition imposed by \pUI.
Second, $\f{\{r'\}}{\{p\}}$ has an HT-model $\langle \emptyset,a\rangle$ (as well as $\langle a,a\rangle$).
But this is a contradiction to $\f{P}{\{p\}}\htmodels r$, which finishes the argument.

Then, \ref{i22}.\ is a consequence of the definition of the respective properties where the uniform property, \pUI, is just a special case of \pSI, and the proof of \ref{i23}.\ is a precise adaptation of that of \ref{i7}.\ replacing all used programs $R$ by sets of facts.
Finally, \ref{i27} is straightforward by the definitions of \pSI\ and $\pSIu$, and \ref{i26}.\ has been shown by Gon\c calves et al.~\shortcite{GoncalvesJKL21}.
\end{proof}

\begin{theorem*}{prop:coincideHorn}
For all Horn programs $P$, every $V\subseteq \sign(P)$, and all forgetting operators $\fgt_1, \fgt_2$ in the classes $\classF_{\strong}$, $\classF_{\weak}$, $\classF_{\FS}$, $\classF_{\htF}$, $\classF_{\smF}$, $\classF_{\Sas}$, $\classF_{\SE}$, $\classF_{\spF}$, $\classF_{\AltNaive}$, $\classF_{\TAltNaive}$, and $\classF_{\up}$, it holds that $\fgt_1(P,V)\htequiv\fgt_2(P,V)$.
\end{theorem*}
\begin{proof}
We will base our proof on Theorem 10 by Wang et al.~\shortcite{WangZZZ14}, a representation result for HT-Forgetting. This theorem implies that, whenever $\classF_{\htF}$ is closed for a class of programs, then it coincides with any class of forgetting operators that is closed for the same class and satisfies \pW\ and  \pPP\ for that class. Our aim now is to prove that every class of forgetting operators mentioned in the statement of this theorem  coincides with $\classF_{\htF}$. 
Just for the sake of simplify we will use the results in Theorem~\ref{thm:propsOps}. We should stress, nevertheless, that there is no circular dependence between these two results.

 In order to prove that a class $\classF$ coincides with $\classF_{\htF}$ on the class of Horn programs, we need to prove that $\classF_{\htF}$ is closed for the class of Horn programs, and: (i) $\classF$ is closed for the class of Horn programs, (ii) $\classF$ satisfies \pW\ and \pPP\ on the class of Horn programs.
 
The fact that $\classF_{\htF}$ is closed for the class of Horn programs is precisely Theorem 8 \cite{WangZZZ14}. So we now prove that each other class of operators satisfies (i) and (ii).

First, it is straightforward from the definitions of $\classF_{\strong}$ and $\classF_{\weak}$ that they coincide on the class of Horn programs. It is also clear that they are closed for this class. Since $\classF_{\strong}$ satisfies \pW\ (Theorem~\ref{thm:propsOps}), $\classF_{\weak}$ satisfies \pPP\ (Theorem~\ref{thm:propsOps}), and they coincide on the class of Horn programs, they both satisfy \pPP\ and \pW\ on that class.

Thm.~\ref{prop:SEFequalsHTF} states that $\classF_{\SE}$ and $\classF_{\FS}$ coincide with $\classF_{\htF}$ on the class of disjunctive programs, whenever the result of $\classF_{\htF}$ is a disjunctive program. Since $\classF_{\htF}$ is closed for the class of Horn programs, it follows immediately that $\classF_{\SE}$ and $\classF_{\FS}$ coincide with  $\classF_{\htF}$ when restricted to Horn programs. 

It follows easily from the algorithm presented by Knorr and Alferes~\shortcite{KnorrA14} that for $\fgt_{\Sas}\in\classF_{\Sas}$, $V\subseteq \sign$, and $P$ a Horn program $\fgt_{\Sas}(P,V)\htequiv \fgt_{\strong}(P,V)$. As a consequence, $\classF_{\Sas}$, just as $\classF_{\strong}$, coincides with $\classF_{\htF}$ on the class of Horn programs.

Since $\classF_{\htF}$ coincides with $\classF_{\Sas}$ for Horn programs, we have that $\classF_{\htF}$ satisfies \pCP\ for Horn programs. Therefore, by definition, $\classF_{\smF}$ coincides with $\classF_{\htF}$ for Horn programs.

In the case of $\classF_{\spF}$, $\classF_{\AltNaive}$, and $\classF_{\TAltNaive}$, the result is a consequence of Proposition 6~\cite{GoncalvesKLW20}, which shows that $\classF_{\spF}$ coincides with $\classF_{\htF}$ when restricted to Horn programs, and Proposition 21~\cite{GoncalvesKLW20}, which shows that the three classes $\classF_{\spF}$, $\classF_{\AltNaive}$, and $\classF_{\TAltNaive}$ coincide when restricted to Horn programs.

Finally, in the case of $\classF_{\up}$, Proposition 1~\cite{GoncalvesJKLW19} shows that $\classF_{\up}$ coincides with $\classF_{\htF}$ when restricted to Horn programs.
\end{proof}

\begin{theorem*}{prop:EqualityFSFSE}
Consider the class of disjunctive programs.
Then, $\classF_{\FS}$ and $\classF_{\SE}$ coincide.
\end{theorem*}
\begin{proof}
 First recall that, on the one hand, given a disjunctive program $P$ and $V\subseteq \sign$, $P_\FS(P,V)$ is defined by removing from $\Cn(P,a)=\{r\mid r \text{ disjunctive}, P\htmodels r$, $\sign(r)\subseteq \sign(P)\}$ all rules in which atoms from $V$ occur. On the other hand, given a disjunctive program $P$ and $V\subseteq \sign$, for any $\fgt_{\SE}\in \classF_{\SE}$, we have that $\fgt_{\SE}(P,V)$ is equivalent to $\Cn_\sign(P)\cap \lang_{\sign(P)\setminus V}$, where $\Cn_\sign(P)=\{r\in\lang_\sign\mid r \text{ disjunctive}, P\vdash_s r\}$. Since, as Wong~\shortcite{Wong09} showed, the consequence $\vdash_s$ is sound and complete with respect to $\htmodels$, we have that $\Cn(P,a)=\Cn_\sign(P)\cap \lang_{\sign(P)}$. Therefore, $\Cn(P,a)\cap \lang_{\sign(P)\setminus V}= \Cn_\sign(P)\cap\lang_{\sign(P)\setminus V}$. This means that $\fgt\in \classF_{\FS}$ iff $\fgt(P,V)\htequiv \Cn(P,a)\cap \lang_{\sign(P)\setminus V}$ iff $\fgt(P,V)\htequiv \Cn_\sign(P)\lang_{\sign(P)\setminus V}$ iff $\fgt \in \classF_{\SE}$. Therefore, $\classF_{\FS}=\classF_{\SE}$.
\end{proof}

\begin{theorem*}{prop:SEFequalsHTF}
Let $P$ be a disjunctive program, $V\subseteq \sign(P)$, $\fgt_{\FS}\in\classF_{\FS}$, $\fgt_{\htF}\in\classF_{\htF}$, and $\fgt_{\SE}\in\classF_{\SE}$.
Then, $\fgt_{\FS}(P,V)\htequiv\fgt_{\htF}(P,V)\htequiv\fgt_{\SE}(P,V)$ whenever $\fgt_{\htF}(P,V)$ is strongly equivalent to a disjunctive program.
\end{theorem*}
\begin{proof} Let $P$ be a disjunctive program, $V\subseteq \sign(P)$, such that $\fgt_{\htF}(P,V)$ is strongly equivalent to a disjunctive program.
We only prove that $\fgt_{\htF}(P,V)\htequiv\fgt_{\SE}(P,V)$ since Theorem~\ref{prop:EqualityFSFSE} already states that $\fgt_{\FS}(P,V)\htequiv\fgt_{\SE}(P,V)$.
 Since both $\classF_{\htF}$ and $\classF_{\SE}$ satisfy \pW, we have that $(i)$ $P\htmodels \fgt_{\htF}(P,V)$ and $(ii)$ $P\htmodels \fgt_{\SE}(P,V)$. Since $\classF_{\SE}$ satisfies \pPP\ (restricted to disjunctives programs) and $\fgt_{\htF}(P,V)$ is equivalent to a disjunctive program, we can use $(i)$ to conclude that $\fgt_{\SE}(P,V)\htmodels \fgt_{\htF}(P,V)$. Since $\classF_{\htF}$ satisfies \pPP\ we can use $(ii)$ to conclude that $\fgt_{\htF}(P,V)\htmodels \fgt_{\SE}(P,V)$.
Therefore, $\fgt_{\SE}(P,V)\htequiv \fgt_{\htF}(P,V)$.
\end{proof}


We now present several intermediate results, each presenting the set of properties satisfied and not satisfied by each class of operators considered in the paper, that we will combine to prove Theorem~\ref{thm:propsOps}.

\begin{theorem}\label{prop:Strongsat}
$\classF_{\strong} $ satisfies \pW, \pSI, \pRC, \pNC, \pPI, \pUI, $\pSIu$,  \pE{\classP_{\hor}}, \pE{\classP_{\nor}}, but not \psC, \pwE, \pSE, \pPP, \pSC, \pCP, \pSP, \pwC, \pwSP, \psSP\ and \pUP.
\end{theorem}
\begin{proof}
To prove \pW, let $P$ be a normal logic program over a signature $\sign$, and $v\subseteq \sign$.
In the first step of the definition of $\fgt_{\strong}(P,V)$ we obtain an intermediate program $P'$ by adding to $P$, for each $v\in V$, the rules $a \la B_1,B_2, \nf C_1, \nf C_2$ such that $a\la B_1,v, \nf C_1$ and $v\la B_2, \nf C_2$ are in $P$. By Lemma~\ref{lemma:InclusionOfModels} every $HT$ model of both $a\la B_1,v, \nf C_1$ and $v\la B_2, \nf C_2$ is also a model of $a \la B_1,B_2, \nf C_1, \nf C_2$. Thus, $\HT(P)\subseteq \HT(P')$. Therefore, $\HT(P\cup P')=\HT(P)\cap \HT(P')=\HT(P)$, i.e., $P\cup P'$ is strongly equivalent to $P$.
In the second and last step of the definition of $\fgt_{\strong}(P,V)$, all rules from $P\cup P'$ containing some $v\in V$ are eliminated, thus obtaining $\fgt_{\strong}(P,V)$. 
Since  $\fgt_{\strong}(P,V)\subseteq P\cup P'$, we have that $\HT(P)=\HT(P\cup P')\subseteq \HT(\fgt_{\strong}(P,V))$.
Note that we proved the result for forgetting one variable. The general result then follows from (F6) \cite{Wong09}, which was shown to hold for both Strong and Weak Forgetting.
 
For \pSI, let $P$ be a normal logic program over a signature $\sign$ and let $V\subseteq \sign$.
Let $P_{\parallel V}$ be the set of rules of $P$ that do not mention any $v\in V$.
It follows from the definition of Strong Forgetting that $P_{\parallel V}\subseteq \fgt_{\strong}(P,V)$.
 Let $R$ be a normal logic program over $\sign\setminus V$.
We now prove that  $\fgt_{\strong}(P\cup R,V)=\fgt_{\strong}(P,V)\cup R$.
Let us prove the two inclusions.

For the left to right inclusion, let $r\in \fgt_{\strong}(P\cup R,V)$. Then, we have two cases:
\begin{description}
 \item[a)] $r\in P_{\parallel V}\cup R$. Then using the above observation we have that $r\in \fgt_{\strong}(P,V)\cup R$.
 \item[b)]  $r=A\la B,B',\nf C, \nf C'$ such that there exist $r_1,r_2\in P\cup R$ and $v\in V$ such that $r_1=A\la B,v,\nf C$ and $r_2=v\la B', \nf C'$. In this case, $r_1,r_2\in P$ since $R$ does not contain rules with atoms from $V$. Therefore, we can immediately conclude that $r\in \fgt_{\strong}(P,V)$, and then $r\in \fgt_{\strong}(P,V)\cup R$.
\end{description}

For the converse inclusion, let $r\in \fgt_{\strong}(P,V)\cup R$. Then, we have two cases:
\begin{description}
 \item[c)] $r\in R$. Then using the above observation we have that $r\in \fgt_{\strong}(P\cup R,V)$.
 \item[d)]  $r\in \fgt_{\strong}(P,V)$. If $r\in P_{\parallel V}$, then using the above observation we have $r\in \fgt_{\strong}(P\cup R,V)$. Otherwise, if $r=A\la B,B',\nf C, \nf C'$ such that there exist $r_1,r_2\in P$ and $v\in V$ such that $r_1=A\la B,v,\nf C$ and $r_2=v\la B', \nf C'$. In this case, also $r_1,r_2\in P\cup R$, and we can immediately conclude that $r\in \fgt_{\strong}(P\cup R,V)$.
\end{description}
 
 The fact that $\classF_{\strong} $ satisfies \pRC, \pNC\ and \pPI\ was proved by Gon\c calves et al.~\shortcite{GKL17}.

The fact that $\classF_{\strong} $ satisfies \pUI\ follows from the already proved fact that $\classF_{\strong} $ satisfies \pSI, and item \ref{i22} of Thm.~\ref{prop:relations}.

The fact that $\classF_{\strong} $ satisfies $\pSIu$ follows from the already proved fact that $\classF_{\strong} $ satisfies \pSI, and item \ref{i27} of Thm.~\ref{prop:relations}.

The fact that  \pE{\classP_{\hor}} is satisfied follows from the simple observation that, when restricted to Horn programs, $\fgt_{\strong}$ and  $\fgt_{\weak}$ coincide and the result is by definition always Horn.

Finally, the definition of the concrete operator $\fgt_{\strong}$ by Zhang and Foo~\shortcite{ZhangF06} implicitly implies that $\classF_{\strong}$ satisfy \pE{\nor}.

For the negative results, Eiter and Wang \shortcite{EiterW08} provided counterexamples showing that $\classF_{\strong}$ does not satisfy \psC\ and \pwE;
Wong \shortcite{Wong09} provided counterexamples showing that $\classF_{\strong}$ does not satisfy \pSE;
Wang et al.~\shortcite{WangZZZ14} provided counterexamples showing that $\classF_{\strong}$ does not satisfy \pPP, and Gon\c calves et al.~\shortcite{GKL17} provided a counterexample to show that $\classF_{\strong}$ does not satisfy \pSC; 
 Wang et al.~\shortcite{WangWZ13}  presented a counterexample showing that $\classF_{\strong}$ does not satisfy \pCP; and Knorr and Alferes \shortcite{KnorrA14} provided a counterexample showing that $\classF_{\strong}$ does not satisfy \pSP.

For \pwC, consider forgetting about $b$ from $P=\{a\la \nf b$, $b\la\nf a\}$.
Since $\as{P}=\{\{a\},\{b\}\}$, to satisfy \pwC, the result must have the two answer sets $\{a\}$ and $\emptyset$, which is not possible for disjunctive programs.
Therefore, $\classF_{\strong}$ does not satisfy \pwC.

Gon\c calves et al.~\shortcite{GoncalvesKLW20} showed that $\classF_{\strong}$ does not satisfy \pwSP\ and \psSP, and that it does not satisfy \pUP~\shortcite{GoncalvesJKLW19}.
\end{proof}

\begin{theorem}\label{prop:WakSat}
$ \classF_{\weak} $ satisfies \pPP, \pSI,  \pRC,	\pNC,  \pPI, \pUI, $\pSIu$, \pE{\classP_{\hor}}, {\pE{\classP_{\nor}}}, but not 
\psC, \pwE, \pSE, \pW, \pSC, \pCP,	\pSP,	 \pwC, \pwSP,	 \psSP\ and \pUP.
\end{theorem}
\begin{proof}
For \pPP\ consider the inferential system given by Wong \shortcite{Wong08}, which is sound and complete with respect to HT-consequence for disjunctive programs.
Suppose that $P\htmodels r$, where $\sign(r)\subseteq \sign\setm V$. Now recall that every rule of $P$ that does not contain the atoms to be forgotten is in the result of forgetting.
The rule (WGPPE) is the only inference rule in Wong's system that allows to derive a rule without the atoms to be forgotten from rules that have those atoms. Since, by definition of Weak Forgetting, the result of applying (WGPPE) belongs to the result of forgetting, we have that $r$ must also be a consequence of the result of forgetting. 

For \pSI, let $P$ be a normal logic program over a signature $\sign$ and let $V\subseteq \sign$.
Let $P_{\parallel V}$ be the set of rules of $P$ that do not mention any $v\in V$.
It follows from the definition of Weak Forgetting that $P_{\parallel V}\subseteq \fgt_{\weak}(P,V)$.
 Let $R$ be a normal logic program over $\sign\setminus V$.
We now prove that  $\fgt_{\weak}(P\cup R,V)=\fgt_{\weak}(P,V)\cup R$.
Let us prove the two inclusions.

For the left to right inclusion, let $r\in \fgt_{\weak}(P\cup R,V)$. Then, we have three cases:
\begin{description}
 \item[a)] $r\in P_{\parallel V}\cup R$. Then using the above observation we have that $r\in \fgt_{\weak}(P,V)\cup R$.
 \item[b)]  $r=A\la B,B',\nf C, \nf C'$ such that there exist $v\in V$, $r_1,r_2\in P\cup R$ such that $r_1=A\la B,v,\nf C$ and $r_2=v\la B', \nf C'$. In this case, $r_1,r_2\in P$ since $R$ does not contain any $v\in V$. Therefore, we can immediately conclude that $r\in \fgt_{\weak}(P,V)$, and then $r\in \fgt_{\weak}(P,V)\cup R$.
 \item[c)]  $r=A\la B, \nf C$ such that there exists $v\in V$ and $r_1=A\la B, \nf v, \nf C\in P\cup R$. Then, $r_1\in P$ since $R$ does not contain any  $v\in V$. Therefore, we can conclude that $r\in \fgt_{\weak}(P,V)$, and then $r\in \fgt_{\weak}(P,V)\cup R$.
\end{description}

For the converse inclusion, let $r\in \fgt_{\weak}(P,V)\cup R$. Then, we have two cases:
\begin{description}
 \item[e)] $r\in R$. Then using the above observation we have that $r\in \fgt_{\weak}(P\cup R,V)$.
 \item[f)]  $r\in \fgt_{\weak}(P,V)$. 
 
\begin{itemize}
 \item[i)] If $r\in P_{\parallel V}$, then using the above observation we have $r\in \fgt_{\weak}(P\cup R,V)$.
 \item[ii)] If $r=A\la B,B',\nf C, \nf C'$ such that there exist $v\in V$, and $r_1,r_2\in P$ such that $r_1=A\la B,v,\nf C$ and $r_2=v\la B', \nf C'$. In this case, also $r_1,r_2\in P\cup R$, and we can immediately conclude that $r\in \fgt_{\weak}(P\cup R,V)$. 
 \item[iii)] If $r=A\la B, \nf C$ such that there exist $v\in V$, and $r_1=A\la B, \nf v, \nf C\in P$. Then, $r_1\in P\cup R$, and we can conclude that $r\in \fgt_{\weak}(P,V)$. Therefore, $r\in \fgt_{\weak}(P,V)\cup R$.
\end{itemize}
\end{description}

 The fact that $\classF_{\weak} $ satisfies \pRC, \pNC\ and \pPI\ was shown by Gon\c calves et al.~\shortcite{GKL17}.

In the case of  \pUI\ the result follows from the already proved fact that $\classF_{\weak} $ satisfies \pSI, and item \ref{i22} of Thm.~\ref{prop:relations}.

The fact that $\classF_{\weak} $ satisfies $\pSIu$ follows from the already proved fact that $\classF_{\weak} $ satisfies \pSI, and item \ref{i27} of Thm.~\ref{prop:relations}.

The fact that  \pE{\classP_{\hor}} is satisfied follows from the simple observation that, when restricted to Horn programs, $\fgt_{\strong}$ and  $\fgt_{\weak}$ coincide and the result is by definition always Horn.

Finally, the definition of the concrete operator $\fgt_{\weak}$ \cite{ZhangF06} implicitly implies that $\classF_{\weak}$ satisfy \pE{\nor}.

For the negative results, Eiter and Wang \shortcite{EiterW08} provided counterexamples showing that $\classF_{\weak}$ does not satisfy \psC\ and \pwE;
Wong \shortcite{Wong09} provides counterexamples showing that $\classF_{\weak}$ does not satisfy \pSE;
 \cite{WangZZZ14} provided a counterexample showing that $\classF_{\weak}$ does not satisfy \pW;
 the fact that $\classF_{\weak} $ does not satisfy \pSC\ was proved by Gon\c calves et al.~\shortcite{GKL17};
  Wang et al.~\shortcite{WangWZ13} presented a counterexample showing that $\classF_{\weak}$ does not satisfy \pCP; and Knorr and Alferes \shortcite{KnorrA14} provided a counterexample showing that $\classF_{\weak}$ does not satisfy \pSP.

For \pwC, consider forgetting about $b$ from $P=\{a\la \nf b$, $b\la\nf a\}$.
Since $\as{P}=\{\{a\},\{b\}\}$, to satisfy \pwC, the result must have the two answer sets $\{a\}$ and $\emptyset$, which is not possible for disjunctive programs.
Therefore, $\classF_{\weak}$ does not satisfy \pwC.

Gon\c calves et al.~\shortcite{GoncalvesKLW20} showed that $\classF_{\weak}$ does not satisfy \pwSP\ and \psSP, and that it does not satisfy \pUP \cite{GoncalvesJKLW19}.
\end{proof}

\begin{theorem}\label{prop:SemSat}
$ \classF_{\sem} $ satisfies \psC, \pwE,  \pPI,  { \pE{\classP_{\hor}}}, {\pE{\classP_{\nor}}}, {\pE{\classP_{\dis}}}, but not 
\pSE, \pW, \pPP, \pSI, \pSC, \pRC, \pNC, \pCP, \pSP, \pwC, \pwSP, \psSP, \pUP, \pUI\ and $\pSIu$.

\end{theorem}
\begin{proof}
 Eiter and Wang \shortcite{EiterW08} proved that $\classF_{\sem}$ satisfies \psC~(Proposition 6) and \pwE~(Proposition 4), and, by the definitions of concrete operators, it is implicit that $\classF_{\sem}$ satisfies { \pE{\classP_{\nor}}}\ and { \pE{\classP_{\dis}}}.
 The fact that $\classF_{\sem} $ satisfies \pPI\ was proved by Gon\c calves et al.~\shortcite{GKL17}.
 Finally, in the case of { \pE{\classP_{\hor}}} the result follows from the observation that a variant of ${\sf forget_1}$ \cite{EiterW08} can be simplified which constructs a single program consisting of facts given that there does exist only at most one answer set in this case.
 

Regarding negative results, counterexamples have been presented in the literature for \pSE~\cite{EiterW08}, for \pW~\cite{WangWZ13}, for \pPP~\cite{WangZZZ14}, and for \pSI~\cite{Wong09}.
Also, the fact that $\classF_{\sem} $ does not satisfy \pSC, \pRC, and \pNC\ was proved by Gon\c calves et al.~\shortcite{GKL17}. 
Wang et al.~\shortcite{WangWZ13} showed that $\classF_{\sem} $ does not satisfy \pCP, 
and Knorr and Alferes \shortcite{KnorrA14} that it does not satisfy \pSP.
The counterexample for \pwC\ is similar to other cases, just by considering forgetting about $b$ from $P=\{a\la \nf b$, $b\la\nf a\}$.
Since $\as{P}=\{\{a\},\{b\}\}$, to satisfy \pwC, the result must have the two answer sets $\{a\}$ and $\emptyset$, which is not possible for disjunctive programs.
Therefore, $\classF_{\sem}$ does not satisfy \pwC.
Gon\c calves et al.~\shortcite{GoncalvesKLW20} showed that $\classF_{\sem}$ does not satisfy \pwSP\ and \psSP, and that it does not satisfy \pUP~\cite{GoncalvesJKLW19}.
For the case of \pUI, consider forgetting about $b$ from $P=\{a\la b\}$. In this case, an operator $\fgt$ of  $\classF_{\sem}$  is such that $\fgt(P,b)=\{\}$, but adding $b\la$ before and after forgetting gives rise to results with different sets of answer sets.
Finally, for the case of $\pSIu$, consider $P=\{a\la b\}$ over the signature $\sign=\{a,b,c\}$, and the program $R=\{b\la\}$ over $\sign\setm \{c\}$.
Since $\as{P}=\{\{\}\}$ and $\as{P\cup R}=\{\{a,b\}\}$, then, according to the definition of ${\sf forget_1}\in \classF_{\sem}$ in \cite{EiterW08}, we have that 
${\sf forget_1}(P,\{c\})=\{\}$ and ${\sf forget_1}(P\cup R,\{c\})=\{a\la, b\la\}$. We can now easily see that $\as{{\sf forget_1}(P,\{c\})\cup R}=\{\{b\}\}\neq\{\{a,b\}\}=\as{{\sf forget_1}(P\cup R,\{c\})}$, and therefore  ${\sf forget_1}(P,\{c\})\cup R\not \equiv_u {\sf forget_1}(P\cup R,\{c\})$.
\end{proof}

\begin{theorem}\label{prop:FSsat}
$ \classF_{\FS} $ satisfies  \pSE, \pW, \pPP,  \pSC, \pRC, \pNC, \pPI,  { \pE{\classP_{\hor}}},  {\pE{\classP_{\dis}}}, but not 
\psC, \pwE, \pSI, \pCP, \pSP, \pwC, \pwSP, \psSP, \pUP, \pUI, $\pSIu$\ and {\pE{\classP_{\nor}}}.
\end{theorem}
\begin{proof}
 
In the case of \pSE, Wong~\shortcite{Wong09} proved that $\classF_{\FS}$ satisfies \pSE~(Lemma 4.27), and it is implicit that it satisfies { \pE{\classP_{\dis}}}.
 Delgrande and Wang \shortcite{DelgrandeW15} proved that $\classF_{\SE}$ satisfies \pSE, \pW\ and \pPP~(Proposition 1). Therefore, by Thm.~\ref{prop:EqualityFSFSE}, the results also hold for $ \classF_{\FS} $.
The fact that $ \classF_{\FS} $ satisfies \pSC, \pRC, \pNC\ and \pPI\ was proved by Gon\c calves et al.~\shortcite{GKL17}. 
Finally, the fact that { \pE{\classP_{\hor}}} holds follows from Theorem 8~\cite{WangZZZ14}, which states that $\classF_{\htF}$ is closed for the class of Horn programs, and Thm.~\ref{prop:SEFequalsHTF}, which implies that the operators from $\classF_{\FS}$ are equivalent with those of $\classF_{\htF}$ for Horn programs.

 For the negative results, in the case of \psC, consider forgetting about $a$ from $P=\{a\la \nf a\}$. The result should be strongly equivalent to $\emptyset$, i.e., the forgetting operation introduces a new answer set.
Turning to \pwE, this property requires that the results of forgetting about $p$ from $P=\{q \la \nf p, q\la \nf q\}$ and from $Q=\{q\la\}$ have the same answer-sets, while $\classF_{\FS}$ requires that the results be strongly equivalent to $\f{P}{p}=\{q\la\nf q\}$ and $\f{Q}{p}=\{q\la\}$, which are obviously not equivalent.

Wong~\shortcite{Wong09} provided a counterexample showing that $\classF_{\FS}$ does not satisfy \pSI.
The fact that $ \classF_{\FS} $ does not satisfy \pCP\ and \pSP\ follows immediately from the fact that it does not satisfy \psC, and items 3. and 4. of Thm.~\ref{prop:relations}.
The counterexample for \pwC\ is similar to other cases, just by considering forgetting about $b$ from $P=\{a\la \nf b$, $b\la\nf a\}$.
Since $\as{P}=\{\{a\},\{b\}\}$, to satisfy \pwC, the result must have the two answer sets $\{a\}$ and $\emptyset$, which is not possible for disjunctive programs.
Therefore, $\classF_{\FS}$ does not satisfy \pwC.
Gon\c calves et al.~\shortcite{GoncalvesKLW20} showed that $\classF_{\FS}$ does not satisfy \pwSP\ and \psSP, and that it does not satisfy \pUP~\cite{GoncalvesJKLW19}.

To prove that $ \classF_{\FS} $ does not satisfy \pUI, consider forgetting about $p$ from the program $P=\{a\la \nf p,b,\ p\la \nf a,\ \la p,b \}$. 
We have that $P\cup \{b\la\}\htmodels a\la$, and since $ \classF_{\FS} $ satisfies \pPP, we have that  $\fgt(P\cup \{b\la\},p)\htmodels a\la$.
Since $P\not\htmodels a\la b$, we have that  $\fgt(P,p)\not\htmodels a\la b$. Therefore $\fgt(P,p)\cup\{b\la\}\not\htmodels a\la$. 
 
 For the case of $\pSIu$, consider the same program as for \pUI, i.e.,  $P=\{a\la \nf p,b,\ p\la \nf a,\ \la p,b \}$. Again, we have that $P\cup \{b\la\}\htmodels a\la$, and since $ \classF_{\FS} $ satisfies \pPP, we have that  $\fgt(P\cup \{b\la\},p)\htmodels a\la$. This means that $\HT(\fgt(P\cup \{b\la\},p\})\subseteq \HT(\{a\la\})=\{\tuple{a,a},\tuple{ab,ab},\tuple{a,ab}\}$. Since, by using the same argument as before, it is also the case that $\fgt(P\cup \{b\la\},p)\htmodels b\la$, we can conclude that $\HT(\fgt(P\cup \{b\la\},p))\subseteq \HT(\{a\la\})\cap \HT(\{b\la\})=\{\tuple{ab,ab}\}$. Since $\tuple{ab,ab}\in \HT(P\cup \{b\la \})$ and $\classF_{\FS}$ satisfies \pW, we have that $\tuple{ab,ab}\in \HT(\fgt(P\cup \{b\la\},p))$, and therefore $\HT(\fgt(P\cup \{b\la\},p))=\{\tuple{ab,ab}\}$. We can then conclude that $\as{\fgt(P\cup \{b\la\},p)}=\{\{a,b\}\}$. On the other hand, we know that $\HT(\fgt(P,p)\cup\{b\la\})\subseteq \HT(\{b\la\})$. Since $\fgt(P,p)\cup\{b\la\}\not\htmodels \{a \la\}$, we have that $\HT(\fgt(P,p)\cup\{b\la\})\not\subseteq \HT(\{a\la\})$.
 We therefore have two alternative possibilities: a) $\tuple{b,b}\in \HT(\fgt(P,p)\cup\{b\la\})$ or b) $\tuple{b,ab}\in \HT(\fgt(P,p)\cup\{b\la\})$. If a) is the case, then $\{b\}\in \as{\fgt(P,p)\cup\{b\la\}}$. If b) is the case we have that $\{a,b\}\notin \as{\fgt(P,p)\cup\{b\la\}}$. In both cases we can conclude that $\as{\fgt(P,p)\cup\{b\la\}}\neq \as{\fgt(P\cup \{b\la\},p)}=\{\{a,b\}\}$. Therefore, $\fgt(P,p)\cup\{b\la\}\not\equiv_{u} \fgt(P\cup \{b\la\},p)$, showing that $\classF_{\sem}$ does not satisfy $\pSIu$.
 
Finally, for the negative result of {\pE{\classP_{\nor}}}, we prove for $\classF_{\SE}$. The corresponding result for $\classF_{\FS}$ follows immediately from Theorem~\ref{prop:EqualityFSFSE}.
 Let $P= \{h_1 \la \nf a_1,\  h_2 \la \nf a_2,\  h_3\la a_1, a_2 \}$ be a normal program.
Using Theorem 3~\cite{DelgrandeW15}, the result of forgetting about $a_1$ from $P$ is equivalent to $P'=\{h_1\vee h_2 \la \nf h_3,\ h_2 \la \nf a_2\}$.
 Using the characterization result for a set of HT-models of normal programs~\cite{EiterFTW04} it not difficult to see that $P'$ is not strongly equivalent to any normal program: just note that both $\tuple{\{h_1\},\{h_1, h_2, a_2\}}$ and  $\tuple{\{h_2\},\{h_1, h_2, a_2\}}$ are HT-models of $P'$, but the so-called Here-intersection $\tuple{\{h_1\}\cap \{h_2\}, \{h_1, h_2, a_2\}}=\tuple{\emptyset, \{h_1, h_2, a_2\}}$ is not.
 \end{proof}

\begin{theorem}\label{prop:FWsat}
$ \classF_{\FW} $ satisfies \psC, \pwE, \pSE,  \pPP, \pSI, \pSC, \pRC, \pNC, \pPI,  \psSP,  \pUI, $\pSIu$, {\pE{\classP_{\hor}}}, {\pE{\classP_{\nor}}}, {\pE{\classP_{\dis}}}, but not
\pW, \pCP, \pSP, \pwC, \pwSP\ and \pUP.
\end{theorem}
\begin{proof}
In the case of \psC, the result follows immediately from Lemma~\ref{lemma:FWEqualityAS}.
For \pwE,  let $\fgt\in \classF_{\FW}$, $P_1, P_2$ disjunctive programs over $\sign$, and $V\subseteq \sign$. 
 Suppose $\as{P_1}=\as{P_2}$. We aim to prove that $\as{\fgt(P_1,V)}=\as{\fgt(P_2,V)}$. The result follows from Lemma~\ref{lemma:FWEqualityAS} and the assumption $\as{P_1}=\as{P_2}$, since  
$ \as{\fgt(P_1,V)}  = \{X\in \as{P_1}: V\cap X=\emptyset\} = \{X\in \as{P_2}: V\cap X=\emptyset\}=\as{\fgt(P_2,V)}$.

Wong~\shortcite{Wong09} proved that $\classF_{\FW}$ satisfies \pSE~(Lemma 4.27), and it is implicit that it satisfies { \pE{\classP_{\dis}}}.
 
 The fact that \pPP\ holds follows easily from the definition of $\classF_{\FW}$. Note that, by definition, $r\in \fgt_{\FW}(P,v)$ for every disjunctive rule $r$ not containing $v$ such that $P\htmodels r$. Therefore, for every disjunctive rule $r$ not containing $v$ and such that $P\htmodels r$ we have that $\fgt_{\FW}(P,v)\htmodels r$.

For \pSI, let $\fgt\in \classF_{\FW}$, $P$ be a program over a signature $\sign$, $V\subseteq \sign$, and $R$ a program over $\sign\setminus V$.
 We aim to prove that $\fgt(P\cup R,V)\htequiv \fgt(P,V)\cup R$. 
Consider the following sequence of equalities: 
\begin{align*}
\HT_{\parallel V}(\fgt(P\cup R,V)) & ={\HT}((P\cup R)\cup \{\la v :v\in V\})\\
& = {\HT}((P\cup \{\la v:v\in V\}) \cup R)\\
& = {\HT}((P\cup \{\la v:v\in V\})) \cap {\HT}(R)\\
& = \HT_{\parallel V}(\fgt(P,V)) \cap {\HT}(R)\\
& = \HT_{\parallel V}(\fgt(P,V) \cup R)
 \end{align*}
  The first and fourth equalities follow from Lemma~\ref{cor:EquilityFW}. The second and the third are well-known properties of HT-models. The fifth equality follows from the fact that $R$ does not contain any $v\in V$ and Lemma~\ref{lemma:intersectionVersusUnionOfVExtensions}. 
 We then have that $\HT_{\parallel V}(\fgt(P\cup R,V))=\HT_{\parallel V}(\fgt(P,V) \cup R)$.
 Since both $\fgt(P\cup R,V)$ and $\fgt(P,V)\cup R$ do not contain any $v\in V$, the above equality immediately entails that $\HT(\fgt(P\cup R,V))=\HT(\fgt(P,V)\cup R)$. 

The fact that $ \classF_{\FW} $ satisfies \pSC, \pRC, \pNC\ and \pPI\ was proved by Gon\c calves et al.~\shortcite{GKL17}.  
 
 Property \psSP\ follows from Thm.~\ref{prop:relations} and the fact that $\classF_{\FW}$ satisfies \psC\ and \pSI.
 
 In the case of  \pUI\ the result follows from the already proved fact that $\classF_{\FW} $ satisfies \pSI, and item \ref{i22} of Thm.~\ref{prop:relations}.
 
The fact that $\classF_{\FW} $ satisfies $\pSIu$ follows from the already proved fact that $\classF_{\FW} $ satisfies \pSI, and item \ref{i27} of Thm.~\ref{prop:relations}.
 
 For { \pE{\classP_{\hor}}}, let $\fgt\in \classF_{\FW}$ and $P$ a Horn program. Using Lemma~\ref{cor:EquilityFW} we have that $\HT_{\parallel V}(\fgt(P,V))=\HT(P\cup\{\la v: v\in V\})$.
Since $P\cup\{\la v:v\in V\}$ is a normal program, it can be easily shown that $\HT(\fgt(P,V))=(\HT_{\parallel V}(\fgt(P,V)))_{\dagger V}$ satisfies all conditions characterizing a class of HT-models of a Horn program \cite{WangZZZ14}, taking into account that $\HT(P\cup\{\la v: v\in V\})$ satisfies these conditions.

For { \pE{\classP_{\nor}}}, let $\fgt\in \classF_{\FW}$ and $P$ a normal program. Using Lemma~\ref{lemma:EquilityFW} we have that $\HT_{\parallel V}(\fgt(P,v))=\HT(P\cup\{\la v\})$.
Since $P\cup\{\la v\}$ is a normal program, it can be easily shown that $\HT(\fgt(P,v))=(\HT_{\parallel V}(\fgt(P,v)))_{\dagger v}$ satisfies all conditions characterizing a class of HT-models of a normal program \cite{EiterFTW04}, taking into account that $\HT(P\cup\{\la v\})$ satisfies these conditions.

 For the negative results, Wang et al.~\shortcite{WangZZZ12} provided counterexamples showing that $\classF_{\FW}$ does not satisfy $\pW$.
 The negative results for \pCP\ and \pSP\ can be illustrated with forgetting about $b$ from $P=\{a\la \nf b$, $b\la\nf a\}$.
Since $\as{P}=\{\{a\},\{b\}\}$, the result must have two answer sets $\{a\}$ and $\emptyset$, which is not possible for disjunctive programs obtained from operators in $\classF_{\FW}$.
 
 The counterexample for \pwC\ is similar to other cases, just by considering forgetting about $b$ from $P=\{a\la \nf b$, $b\la\nf a\}$.
Since $\as{P}=\{\{a\},\{b\}\}$, to satisfy \pwC, the result must have the two answer sets $\{a\}$ and $\emptyset$, which is not possible for disjunctive programs.
Therefore, $\classF_{\FW}$ does not satisfy \pwC, which also implies that it does not satisfy \pwSP\ by Thm.~\ref{prop:relations}.
\end{proof}

\begin{theorem}\label{prop:HTsat}
$ \classF_{\htF} $ satisfies  \pSE, \pW, \pPP, \pSI, \pSC, \pRC, \pNC, \pPI,  \pUI, $\pSIu$,{ \pE{\classP_{\hor}}},  {\pE{\classP_{\ex}}}, but not
\psC, \pwE, \pCP, \pSP, \pwC, \pwSP, \psSP, \pUP, {\pE{\classP_{\nor}}}\ and {\pE{\classP_{\dis}}}.
\end{theorem}
\begin{proof}
 Wang et al.~\shortcite{WangZZZ12} proved that $\classF_{\htF}$ satisfies \pSE~(Proposition 3), \pW, \pPP~(Theorem 3), { \pE{\classP_{\hor}}}~(Theorem 2), and { \pE{\classP_{\ex}}}~(Theorem 1).

For \pSI, let $\fgt\in \classF_{\htF}$, $P$ be a program over a signature $\sign$, $V\subseteq \sign$, and $R$ a program over $\sign\setminus V$.
We aim to prove that $\fgt({P},{V})\cup R \htequiv \fgt({P\cup R},{V})$, which is the same as $\HT(\fgt({P},{V})\cup R) = \HT(\fgt({P\cup R},{V}))$.
Consider the following sequence of equalities.
\begin{align*}
\HT(\fgt({P},{V})\cup R) & = \HT(\fgt({P},{V}))\cap \HT(R)\\
& = ({\HT}_{\parallel V}(P))_{\dagger V} \cap ({\HT}_{\parallel V}(R))_{\dagger V}\\
& = ({\HT}_{\parallel V}(P) \cap {\HT}_{\parallel V}(R))_{\dagger V}\\
& = ({\HT}_{\parallel V}(P\cup R))_{\dagger V}\\
& = \HT( \fgt({P\cup R},{V}))
\end{align*}
The first equality follows from a well-known property of $HT$-models. The second equality follows from the definition of $HT$-forgetting. The third equality follows from Lemma~\ref{lemma:intersectionOfVExtensions}, while the fourth equality follows from Lemma~\ref{lemma:intersectionVersusUnionOfVExtensions}. Finally, the last equality also follows from the definition of $HT$-forgetting.

 The fact that $ \classF_{\htF} $ satisfies \pSC, \pRC, \pNC\ and \pPI\ was proved by Gon\c calves et al.~\shortcite{GKL17}.  
 In the case of  \pUI\ the result follows from the already proved fact that $\classF_{\htF} $ satisfies \pSI, and item \ref{i22} of Thm.~\ref{prop:relations}.
 
The fact that $\classF_{\htF} $ satisfies $\pSIu$ follows from the already proved fact that $\classF_{\htF} $ satisfies \pSI, and item \ref{i27} of Thm.~\ref{prop:relations}.

  For the negative results, in the case of \psC, consider forgetting about $a$ from $P=\{a\la \nf a\}$. The result should be strongly equivalent to $\emptyset$, i.e., the forgetting operation introduces a new answer-set.
Turning to \pwE, this property requires that the results of forgetting about $p$ from $P=\{q \la \nf p, q\la \nf q\}$ and from $Q=\{q\la\}$ have the same answer sets, while $\classF_{\htF}$ requires that the results be strongly equivalent to $\f{P}{p}=\{q\la\nf q\}$ and $\f{Q}{p}=\{q\la\}$, which are obviously not equivalent.

The fact that $ \classF_{\htF} $ does not satisfy \pCP\ and \pSP\ follows immediately from the fact that it does not satisfy \psC, and items 3. and 4. of Thm.~\ref{prop:relations}.
 
 For \pwC, consider forgetting about $b$ from $P=\{a\la \nf b$, $b\la\nf a\}$.
Since $\as{P}=\{\{a\},\{b\}\}$, to satisfy \pwC, the result must have the two answer sets $\{a\}$ and $\emptyset$, but the result of $\fgt(P,b)$ for $\fgt\in \classF_{\htF}$ is equivalent to the empty program.
Therefore, the answer set $\{a\}$ is not preserved, and thus $\classF_{\htF}$ does not satisfy \pwC.

The fact that $ \classF_{\htF} $ does not satisfy \pwSP\ and \psSP\ follows immediately from the fact that it does not satisfy \psC\ and \pwC, and items 8. and 9. of Thm.~\ref{prop:relations}.

Gon\c{c}alves et al.~\shortcite{GoncalvesJKLW19}  showed that $\classF_{\htF}$ does not satisfy \pUP.
 Wang \shortcite{WangZZZ12} provided counterexamples showing that $\classF_{\htF}$ does not satisfy {\pE{\classP_{\nor}}}\ and {\pE{\classP_{\dis}}}.
\end{proof}

\begin{theorem}\label{prop:SMsat}
$ \classF_{\smF} $ satisfies \psC, \pwE, \pSE,  \pPP,  \pPI, \pCP,  \pwC,  { \pE{\classP_{\hor}}},  {\pE{\classP_{\ex}}}, but not 
\pW, \pSI, \pSC, \pRC, \pNC, \pSP, \pwSP, \psSP, \pUP, \pUI, $\pSIu$, {\pE{\classP_{\nor}}}\ and {\pE{\classP_{\dis}}}.
\end{theorem}
\begin{proof}
By definition, $ \classF_{\smF} $ satisfies \pCP. From this and Thm.~\ref{prop:relations} it follows easily that $ \classF_{\smF} $ satisfies \psC, \pwC\ and \pwE.
Wang et al.~\shortcite{WangWZ13} have shown that $\classF_{\smF}$ satisfies \pSE, \pPP, { \pE{\classP_{\ex}}}~(Proposition 1), and { \pE{\classP_{\hor}}}~(Theorem 3), and the fact that $ \classF_{\smF} $ satisfies \pPI\ was proved by Gon\c calves et al.~\shortcite{GKL17}.

For the negative results, Wang et al.~\shortcite{WangWZ13} pointed out that $\classF_{\smF}$ does not satisfy \pW.
In the case of \pSI, consider forgetting about $b$ from $P=\{a\la \nf b$, $b\la \nf c\}$.
In this case we have that $\f{P}{b}\htequiv \emptyset$ for $\fgt\in\classF_{\smF}$, so adding $c\la$ results precisely in a program containing this fact.
If we add $c\la$ before forgetting, then the $HT$-models of the result of forgetting, ignoring all occurrences of $b$, correspond precisely to $\langle\{c\},\{c\}\rangle,\langle\{c\},\{a,c\}\rangle$, and $\langle\{a,c\},\{a,c\}\rangle$.
To preserve the answer sets, only the last of these three can be considered.
Hence, $a\la$ and $c\la$ (or strongly equivalent rules) occur in the result of forgetting for any $\fgt\in\classF_{\smF}$, and \pSI\ does not hold.
Since this counterexample only adds a fact before and after forgetting, it also shows that \pUI\ does not hold. 

The fact that $ \classF_{\smF} $ does not satisfy \pSC, \pRC, \pNC\ was proved by Gon\c calves et al.~\shortcite{GKL17}.  
Knorr and Alferes \shortcite{KnorrA14} provided a counterexample showing that $\classF_{\smF}$ does not satisfy \pSP. 
 
Gon\c calves et al.~\shortcite{GoncalvesKLW20} showed that $\classF_{\smF}$ does not satisfy \pwSP\ and \psSP, and that it does not satisfy \pUP\ \shortcite{GoncalvesJKLW19}.
For the case of $\pSIu$, consider a program $P=\{a\la \nf b, b\la \nf c\}$. We can easily check that for any $\fgt\in \classF_{\smF}$, we have that $\fgt(P,b)\htequiv \{\}$.
We can then conclude that $\as{\fgt(P,b)\cup \{c\la\}}=\{\{c\}\}$. On the other hand, since $\classF_{\smF}$ satisfies \pCP, we have that $\as{\fgt(P\cup \{c\la\},b)}=\as{P\cup \{c\la\}}_{\parallel \{b\}}=\{\{a,c\}\}$. But then, $\as{\fgt(P,b)\cup \{c\la\}}=\{\{c\}\}\neq \{\{a,c\}\}=\as{\fgt(P\cup \{c\la\},b)}$, and we can conclude that $\fgt(P,b)\cup \{c\la\}\not\equiv_{u}\fgt(P\cup \{c\la\},b)$, showing that $\classF_{\smF}$ does not satisfy $\pSIu$.

Wang et al.~\shortcite{WangWZ13} provided counterexamples showing that $\classF_{\smF}$ does not satisfy { \pE{\classP_{\nor}}} and { \pE{\classP_{\dis}}}.
\end{proof}

\begin{theorem}\label{prop:SaSsat}
$ \classF_{\Sas} $ satisfies 
\psC, \pwE, \pSE, \pW, \pPP, \pSI, \pSC, \pRC, \pNC, \pPI, \pCP, \pSP, \pwC, \pwSP, \psSP, \pUP, \pUI, $\pSIu$, and { \pE{\classP_{\hor}}}.
\end{theorem}
\begin{proof}
By definition of the class, \pSP\ is satisfied. From this fact it easily follows from the results in Thm.~\ref{prop:relations} that $ \classF_{\Sas}$ satisfies 
 \psC, \pwE, \pSE, \pPP, \pSI, \pCP, \pSP, \pwC, \pwSP, \psSP, \pUP, \pUI, $\pSIu$.

For \pW, \pSC, \pRC, \pNC\ and \pPI, first note that Thm. 1 \cite{GoncalvesKLW20} states that there is no operator over a class of programs that contains normal programs and that satisfies \pSP. Therefore, every operator in $ \classF_{\Sas}$ is necessarily defined over $\classP_{\hor}$. 
Thm.~\ref{prop:coincideHorn} states that when restricted to Horn programs the result of operators in $ \classF_{\Sas} $ is strongly equivalent to the result of operators in the class $ \classF_{\FS} $.
Since Thm.~\ref{prop:FSsat} states that $ \classF_{\FS} $ satisfies \pW, \pSC, \pRC, \pNC, \pPI\ and { \pE{\classP_{\hor}}}, so does $ \classF_{\Sas}$.
\end{proof}

\begin{theorem}\label{prop:SEsat}
$ \classF_{\SE} $ satisfies  \pSE, \pW, \pPP,  \pSC, \pRC, \pNC, \pPI,  { \pE{\classP_{\hor}}},  {\pE{\classP_{\dis}}}, but not
\psC, \pwE, \pSI, \pCP, \pSP, \pwC, \pwSP, \psSP, \pUP, \pUI, $\pSIu$ and {\pE{\classP_{\nor}}}.
\end{theorem}
\begin{proof}
The results follow from Thm.~\ref{prop:EqualityFSFSE} and Thm.~\ref{prop:FSsat}.
\end{proof}

\begin{theorem}\label{prop:SPsat}
$ \classF_{\spF} $ satisfies  \pSE,  \pPP, \pSI, \pwC, \pwSP,  \pUI, $\pSIu$, {\pE{\classP_{\hor}}},  {\pE{\classP_{\ex}}}, but not
\psC, \pwE, \pW, \pSC, \pRC, \pNC, \pPI, \pCP, \pSP, \psSP, \pUP, {\pE{\classP_{\nor}}}\ and {\pE{\classP_{\dis}}}.
\end{theorem}
\begin{proof}
Gon\c calves et al.~\shortcite{GoncalvesKLW20} showed that $\classF_{\spF}$ satisfies \pSE, \pPP, \pSI, \pwC, \pwSP, { \pE{\classP_{\hor}}}\ and  {\pE{\classP_{\ex}}}.
Property \pUI\ follows from item 15 of Thm.~\ref{prop:relations} and the fact that  $\classF_{\spF}$ satisfies \pSI.

The fact that $\classF_{\spF} $ satisfies $\pSIu$ follows from the already proved fact that $\classF_{\spF} $ satisfies \pSI, and item \ref{i27} of Thm.~\ref{prop:relations}.

The negative results for \psC, \pwE, \pW, \pCP, \pSP, \psSP, {\pE{\classP_{\nor}}}\ and {\pE{\classP_{\dis}}} were proven by Gon\c calves et al.~\shortcite{GoncalvesKLW20}.

For \pSC, consider the program $P=\{a\la \nf p,\ p\la \nf a\}$. We have that $P\htmodels a\la \nf p$, but $\fgt(P,\{p\})\not\htmodels \fgt(\{a\la\nf p\},p)$, since $\fgt(P,\{p\})$ is strongly equivalent to $\{a\la \nf\nf a\}$ and  
$\fgt(\{a\la\nf p\},\{p\})$ is strongly equivalent to $\{a\la\}$.

The fact that $ \classF_{\spF}$ does not satisfy \pRC\ follows from item 14. of Thm.~\ref{prop:relations} and the fact that $ \classF_{\spF} $ satisfies \pwC\ and \pUI.
 
 For \pNC\ consider the program $P=\{a\la p; p\la \nf\nf p\}$. Then, for $\fgt\in \classF_{\spF}$ we have that $\f{P}{\{p\}}\htequiv\{a\la \nf\nf a\}$. Therefore, $\f{P}{\{p\}}\htmodels a\la \nf\nf a$, but it is not the case that $P\htmodels a\la \nf\nf a,\nf p$.

The fact that $ \classF_{\spF} $ does not satisfy \pPI\ follows from the result given in~\cite{GKL17} about the impossibility of iteration while satisfying \pSP. 
The counterexample for \pUP\ was presented by Gon\c calves et al.~\shortcite{GoncalvesJKLW19}. 
\end{proof}

\begin{theorem}\label{prop:FRsat}
$ \classF_{\rF} $ satisfies \psC,  \pSE,  \pPP, \pSI,  \pRC, \pNC,  \psSP,  \pUI, $\pSIu$,{ \pE{\classP_{\hor}}},  {\pE{\classP_{\ex}}}, but not 
\pwE, \pW, \pSC, \pPI, \pCP, \pSP, \pwC, \pwSP, \pUP, {\pE{\classP_{\nor}}}\ and {\pE{\classP_{\dis}}}.
\end{theorem}
\begin{proof}
Gon\c calves et al.~\shortcite{GoncalvesKLW20} showed that $\classF_{\rF}$ satisfies \psC,  \pSE,  \pPP, \pSI, \psSP, { \pE{\classP_{\hor}}}\ and  {\pE{\classP_{\ex}}}.
 
To show that $\classF_{\rF}$ satisfies \pNC, let $r=A\la B, \nf C, \nf\nf D$ be a rule such that $\fgt(P,V)\htmodels r$. Then, $\HT(\fgt(P,V))\subseteq \HT( r)$.
Let $r'=A\la B, \nf C, \nf V, \nf\nf D$. Then it is clear that $\HT(r)\subseteq \HT(r')$, and as a consequence we can conclude that $\HT(\fgt(P,V))\subseteq \HT(r')$.
Consider two cases: first let $\tuple{X,Y}\in \HT(P)$ such that $Y\cap V\neq \emptyset$. In this case, $\tuple{X,Y}\in \HT(r')$ since $Y\models r'$ and $\{r'\}^Y=\emptyset$.
 Now take $\tuple{X,Y}\in \HT(P)$ such that $Y\cap V= \emptyset$. By definition of $ \classF_{\rF}$, we have that $\tuple{X,Y} \in \HT(\fgt(P,V))$, and therefore $\tuple{X,Y}\in \HT(r')$.
 In both cases, $\tuple{X,Y}\in \HT(r')$, and therefore $P\htmodels r'$.
 
 To prove \pRC, let $r=A\la B, \nf C, \nf\nf D$ be a rule such that $\fgt(P,V)\htmodels r$. Let $r'=A\la B, \nf C, \nf V, \nf\nf D$. 
 We can easily prove that if $\tuple{X,Y}\in \HT(r')$ and $Y\cap V=\emptyset$ then $\tuple{X,Y}\in \HT(r)$.
 Now, since $\classF_{\rF}$ satisfies \pNC, we can conclude that $P\htmodels r'$.
 We aim to prove that $\fgt(\{r'\},V)\htmodels r$. For that, let $\tuple{X,Y}\in \HT(\fgt(\{r'\},V))$. Then, there exists $A\in Rel^Y_{\tuple{\{r'\},V}}$ such that $\tuple{X\cup X',Y\cup A}\in \HT(r')$, where $X'\subseteq A$.
 We now prove that $A=\emptyset$. Suppose not. Then, since $\{r'\}^{Y\cup A}=\{\}$, we have that $\tuple{Y,Y\cup A}$ is also a model of $r'$, which contradicts the fact that $A\in \Rel^Y_{\tuple{\{r'\},V}}$.
Since $A$ must be empty we can conclude that $\tuple{X,Y}\in \HT(r')$. Since $Y\cap V=\emptyset$, the above consideration entails that $\tuple{X,Y}\in \HT(r)$, which completes the proof.
 
 Property \pUI\ follows from item 15 of Thm.~\ref{prop:relations} and the fact that $\classF_{\rF}$ satisfies \pSI.

 Property $\pSIu$\ follows from item 17 of Thm.~\ref{prop:relations} and the fact that $\classF_{\rF}$ satisfies \pSI.
 
The negative results for  \pwE, \pW, \pPI, \pCP, \pSP, \pwC, \pwSP, {\pE{\classP_{\nor}}}\ and {\pE{\classP_{\dis}}} are due to Gon\c calves et al.~\shortcite{GoncalvesKLW20}.
 
 For \pSC, consider the program $P=\{a\la \nf p,\ p\la \nf a\}$. We have that $P\htmodels a\la \nf p$, but $\fgt(P,\{p\})\not\htmodels \fgt(\{a\la\nf p\},p)$, since $\HT(\fgt(P,\{p\}))$ contains $\tuple{\emptyset,\emptyset}$, therefore $\fgt(P,\{p\})\not\htmodels a\la$, but $\fgt(\{a\la\nf p\},\{p\})$ is strongly equivalent to $\{a\la\}$.

The fact that $ \classF_{\rF} $ does not satisfy \pPI\ follows from the result about the impossibility of iteration while satisfying \pSP\ \cite{GKL17}.

The counterexample for \pUP\ was given by Gon\c calves et al.~\shortcite{GoncalvesJKLW19}. 
\end{proof}

\begin{theorem}\label{prop:FMsat}
$ \classF_{\mF} $ satisfies \psC, \pwE, \pSE,  \pPP,  \pCP,  \pwC, \psSP, { \pE{\classP_{\hor}}},  {\pE{\classP_{\ex}}}, but not 
\pW, \pSI, \pSC, \pRC, \pNC, \pPI, \pSP, \pwSP,  \pUP, \pUI, $\pSIu$, {\pE{\classP_{\nor}}} and {\pE{\classP_{\dis}}}.
\end{theorem}
\begin{proof}
 Gon\c calves et al.~\shortcite{GoncalvesKLW20} showed that $\classF_{\mF}$ satisfies \psC, \pwE, \pSE,  \pPP,  \pCP,  \pwC, \psSP, { \pE{\classP_{\hor}}},  {\pE{\classP_{\ex}}}, as well as the negative results for \pW, \pSI, \pSP, \pwSP, {\pE{\classP_{\nor}}}\ and {\pE{\classP_{\dis}}}.
 
In the case of \pSC, consider the program $P=\{a\la \nf p,\ p\la \nf a\}$. We have that $P\htmodels a\la \nf p$, but $\fgt(P,\{p\})\not\htmodels \fgt(\{a\la\nf p\},p)$, since $\fgt(P,\{p\})$ is strongly equivalent to $\{a\la \nf\nf a\}$ and  
$\fgt(\{a\la\nf p\},\{p\})$ is strongly equivalent to $\{a\la\}$.

In the case of \pRC\ consider forgetting about $p$ from the program $P=\{a\la p,\ p\la \nf\nf p\}$. 
We have that $\fgt(P,\{p\})$ is strongly equivalent to $\{a\la \nf \nf a\}$.
Nevertheless, there is no rule $r'$ over $a$ and $p$ such that $P\htmodels r'$ and $\fgt(\{r'\},\{p\})\htmodels a\la \nf\nf a$.

 For \pNC\ consider the program $P=\{a\la p; p\la \nf\nf p\}$. Then, for $\fgt\in \classF_{\mF}$ we have that $\f{P}{\{p\}}\htequiv\{a\la \nf\nf a\}$. Therefore, $\f{P}{\{p\}}\htmodels a\la \nf\nf a$, but it is not the case that $P\htmodels a\la \nf\nf a,\nf p$.
 
 The fact that $ \classF_{\mF} $ does not satisfy \pPI\ follows from the result about the impossibility of iteration while satisfying \pSP\ \cite{GKL17}.
  The counterexample for \pUP\ was presented by Gon\c calves et al.~\shortcite{GoncalvesJKLW19}. 

 The negative result for \pUI\ follows from item 16. of Thm.~\ref{prop:relations} and the fact that  $ \classF_{\mF} $ satisfies \pCP\ but not \pUP.
 
 In the case of $\pSIu$, consider a program $P$ over $\sign=\{a,b,p\}$ such that $\HT(P)=\{\tuple{ab,ab},$ $\tuple{a,ab}, \tuple{abp,abp}$, $\tuple{b,abp}, \tuple{a,a}, \tuple{\emptyset,a}, \tuple{ap,ap}\}$.
By definition, for $\fgt\in \classF_{\mF}$, we have that forgetting $p$ is such that $\HT(\fgt(P,p))=$$\{\tuple{ab,ab},$ $\tuple{b,ab},\tuple{a,ab},\tuple{a,a}\}$.
Consider $R$ over $\sign\setm\{p\}$ such that $\HT(R)=\{\tuple{ab,ab}$, $\tuple{b,ab}\}$. 
Then, $\HT(P\cup R)=\{\tuple{ab,ab}, \tuple{abp,abp}$, $\tuple{b,abp}\}$. Again by definition, for $\fgt\in \classF_{\mF}$, we have that 
$\HT(\fgt(P\cup R,p))=\{\tuple{ab,ab}\}$. On the other hand, $\HT(\fgt(P,p)\cup R)=\{\tuple{ab,ab},\tuple{b,ab}\}$. 
This then implies that $\as{\fgt(P\cup R,p)}=\{\{a,b\}\}\neq \{\}=\as{\fgt(P,p)\cup R}$, which means that $\fgt(P\cup R,p)\not\equiv_{u}\fgt(P,p)\cup R$, thus showing that $\classF_{\mF}$ does not satisfy $\pSIu$. 
\end{proof}

\begin{theorem}\label{prop:UPsat}
$ \classF_{\up} $ satisfies \psC, \pwE, \pSE,   \pPI, \pCP,  \pwC,  \pUP, \pUI, { \pE{\classP_{\hor}}},  {\pE{\classP_{\ex}}}, but not
\pW, \pPP, \pSI, \pSC, \pRC, \pNC, \pSP, \pwSP, \psSP, $\pSIu$, {\pE{\classP_{\nor}}}\ and {\pE{\classP_{\dis}}}.
\end{theorem}
\begin{proof}
Gon\c calves et al.~\shortcite{GoncalvesJKLW19} showed that $ \classF_{\up} $ satisfies \psC, \pwE, \pSE,   \pPI, \pCP,  \pwC,  \pUP, \pUI  { \pE{\classP_{\hor}}}, {\pE{\classP_{\ex}}}, and that it does not satisfy \pW, \pPP, \pSI, \pSP, {\pE{\classP_{\nor}}}\ and {\pE{\classP_{\dis}}}. Also, a weaker version of \pPI\ was proved, showing that it is possible to iterate the operators when applied in the context of modular logic programming. Gon\c calves et al.~\shortcite{GoncalvesJKL21} proved that $ \classF_{\up} $ does not satisfy $\pSIu$.

In the case of \pSC, consider the program $P=\{a\la \nf p,\ p\la \nf a\}$. We have that $P\htmodels a\la \nf p$, but $\fgt(P,\{p\})\not\htmodels \fgt(\{a\la\nf p\},p)$, since $\fgt(P,\{p\})$ is strongly equivalent to $\{a\la \nf\nf a\}$ and  $\fgt(\{a\la\nf p\},\{p\})$ is strongly equivalent to $\{a\la\}$.

The negative result for \pRC\ follows from item 14. of Thm.~\ref{prop:relations} and the fact that $ \classF_{\up} $ satisfies \pwC\ and \pUI.

 For \pNC\ consider the program $P=\{a\la p; p\la \nf\nf p\}$. Then, for $\fgt\in \classF_{\up}$ we have that $\f{P}{\{p\}}\htequiv\{a\la \nf\nf a\}$. Therefore, $\f{P}{\{p\}}\htmodels a\la \nf\nf a$, but it is not the case that $P\htmodels a\la \nf\nf a,\nf p$.

For \pwSP, consider $P$ such that $\HT(P)=\{\tuple{ab,ab}, \tuple{a,ab}, \tuple{b,abp}, \tuple{abp,abp}, \tuple{a,abp}\}$. 
Then, by definition $\HT(\fgt(P,\{p\}))=\{\tuple{ab,ab}, \tuple{a,ab}, \tuple{\emptyset,ab}\}$.
Consider a program $R$ over $\{a,b\}$ such that $\HT(R)=\{\tuple{ab, ab},\tuple{\emptyset, ab} \}$ (over $\{a,b\}$).
Then, $\HT(P\cup R)=\HT(P)\cap \HT(R)=\{\tuple{abp, abp}, \tuple{ab, ab}\}$, thus $\{a,b\}\in \as{P\cup R}_{\parallel_V}$, but $\{a,b\}\notin \as{\fgt(P,V)\cup R}$.

For \psSP, consider $P$ such that $\HT(P)=\{\tuple{ab,ab}, \tuple{a,ab}, \tuple{b,abp}, \tuple{abp,abp}\}$.  
Then, by definition $\HT(\fgt(P,\{p\}))=\{\tuple{ab,ab}, \tuple{\emptyset,ab}\}$.
Consider a program $R$ over $\{a,b\}$ s.t. $\HT(R)=\{\tuple{ab, ab},\tuple{a, ab}, \tuple{b, ab} \}$ (over $\{a,b\}$). 
Then $\fgt(P,\{p\})\cup R$ has an answer set $\{a,b\}$, but this is not an answer set of $\fgt(P\cup R,\{p\})$. 
\end{proof}


\begin{theorem*}{thm:propsOps}
All results in Table~\ref{fig:properties} hold. 
\end{theorem*}
\begin{proof}
The result is an immediate consequence of Thms.~\ref{prop:Strongsat} to \ref{prop:UPsat}.
\end{proof}

\begin{theorem*}{thm:allProperties}
For Horn programs, the following holds:
\begin{itemize}

\item $\classF_{\strong}$, $\classF_{\weak}$, $\classF_{\FS}$, $\classF_{\htF}$, $\classF_{\smF}$, $\classF_{\Sas}$, $\classF_{\SE}$, $\classF_{\spF}$, $\classF_{\rF}$, $\classF_{\mF}$, and $\classF_{\up}$ satisfy \pW, \pRC, \pSP, and \pPI;

\item $\classF_{\sem}$ satisfies  \pCP\ and \pPI;

\item $\classF_{\FW}$ satisfies the same properties as in the general case.

\end{itemize}
\end{theorem*}
\begin{proof}
 The fact that $\classF_{\strong}$, $\classF_{\weak}$, $\classF_{\FS}$, $\classF_{\htF}$, $\classF_{\smF}$, $\classF_{\Sas}$, $\classF_{\SE}$, $\classF_{\spF}$, $\classF_{\rF}$, $\classF_{\mF}$, and $\classF_{\up}$, when restricted to Horn programs, satisfy \pW, \pRC, \pSP, and \pPI\ follows easily from Thm.~\ref{prop:coincideHorn}, and the fact that $\classF_{\Sas}$ satisfies all these properties.
 
 The fact that $\classF_{\sem}$, when restricted to Horn programs, additionally satisfies \pCP\  follows easily from the fact that any Horn program has at most one answer set, i.e., $\as{P}=\{A\}$ or $\as{P}=\{\}$. Then, for every $\fgt_{\sem}\in \classF_{\sem}$, by definition we have $\as{\fgt_{\sem}(P,V)} = \Min{\as{P}_{\parallel V}}$. Since $\as{P}$ has at most one element we have $\as{\fgt_{\sem}(P,V)}=\{A_{\parallel V}\}=\as{P}_{\parallel V}$ or $\as{\fgt_{\sem}(P,V)}=\{\}=\as{P}_{\parallel V}$. Therefore, $\classF_{\sem}$ satisfies \pCP\ when restricted to Horn programs.
 \end{proof}

\begin{theorem*}{thm:CPimpliesASInterpolation}
If a class $\classF$ of operators over a class $\classP$ of logic programs satisfies property \pCP, then every operator of that class can be used to obtain uniform interpolants w.r.t.~$\vsim$.
\end{theorem*}

\begin{proof}
 For that, assume that $\fgt$ satifies \pCP, i.e, given $P$ and $V\subseteq\sign$, we have that $\as{\f{P}{V}}=\as{P}_{\parallel V}$. Condition $(i)$ of uniform interpolation follows easily from the fact that every $M\in \as{P}$ is such that $M_{\parallel V}\in \as{\f{P}{V}}$, and therefore $M\models \f{P}{V}$. For condition $(ii)$, we let $R$ be a program such that $P\vdash R$ and $\sign(R)\subseteq \sign\setm V$, and we aim to conclude that $\f{P}{V}\vsim R$. Let $M\in \as{\f{P}{V}}$. Then, since we are assuming that $\as{\f{P}{V}}=\as{P}_{\parallel V}$, we know that there is $M^*\in \as{P}$ such that $M=M^*\setm V$. Given that $P\vdash R$, we can conclude that $M^*\models R$, and since $R$ is a program over $\sign\setm V$, we can conclude that $M\models R$.
\end{proof}

\begin{theorem*}{thm:classesSatASInterpolation}
Any forgetting operator of the classes $\classF_{\smF}$, $\classF_{\mF}$, $\classF_{\up}$, and $\classF_{\rF}$ can be used to obtain uniform interpolants with respect to $\vsim$. 
 \end{theorem*}
\begin{proof}
The case of $\classF_{\smF}$, $\classF_{\mF}$, and $\classF_{\up}$ follows from Thm.\ref{thm:CPimpliesASInterpolation} and the fact that these classes satisfy \pCP. 

In the case of $\classF_{\rF}$ let $P$ be a program and $V\subseteq \sign$. We start by considering $M\in \as{P}$. Taking into account the definition of $\classF_{\rF}$, it is easy to see that, for every $\fgt \in\classF_{\rF}$, we have that $\tuple{M\setm V, M\setm V}\in \HT(\fgt(P,V))$. This implies that $M\setm V\models \fgt(P,V)$, and since $\fgt(P,V)$ is over $\sign\setm V$, we also have that $M\models \fgt(P,V)$, meaning that condition $(i)$ of uniform interpolation is satisfied. Condition $(ii)$ is a consequence of $\classF_{\rF}$ satisfying \psC, i.e., $\as{\fgt(P,V)}\subseteq \as{P}_{\parallel V}$. 
\end{proof}

\begin{theorem*}{thm:WPPcharacterizInterpolation}
A class $\classF$ of forgetting operators can be used to obtain uniform interpolants w.r.t.~$\htmodels$ iff $\classF$ satisfies both \pW\ and \pPP.
\end{theorem*}
\begin{proof}
 The result follows immediately, since the conditions $(i)$ and $(ii)$ of uniform interpolation precisely coincide with \pW\ and \pPP, respectively.
\end{proof}

\begin{theorem*}{thm:opsSatInterpolationHT}
 Any forgetting operator of the classes $\classF_{\htF}$ and $\classF_{\FS}$ can be used to obtain uniform interpolants with respect to $\htmodels$. 
\end{theorem*}
\begin{proof}
 The proof follows immediately from Thm.~\ref{thm:WPPcharacterizInterpolation} and the fact that $\classF_{\htF}$ and $\classF_{\FS}$ satisfy \pW\ and \pPP.
\end{proof}



\begin{lemma}\label{lemma:EquilityFW}
 Let $P$ be a disjunctive program over a signature $\sign$ and $v\in \sign$.
Then, for every $\fgt\in \classF_{\FW}$ we have  \[\HT_{\parallel \{v\}}(\fgt(P,v))=\HT(P\cup\{\la v\}).\]
\end{lemma}
\begin{proof}
 Let $P$ be a disjunctive program over a signature $\sign$, and $v\in \sign$.
 We prove the equality for $\fgt_{\FW}$ and the result extends to every $\fgt\in \classF_{\FW}$ since these only differ from $\fgt_{\FW}$ up to strong equivalence.
 We prove both inclusions of the equality $\HT_{\parallel \{v\}}(\fgt_{\FW}(P,v))=\HT(P\cup\{\la v\})$.
 In what follows let $A,B,C\subseteq \sign\setminus \{v\}$.

 We start with the left to right inclusion.
 Let $I=\tuple{X,Y}$ such that $I\in \HT_{\parallel \{v\}}(\fgt_{\FW}(P,v))$. Since $v$ does not occur in $\fgt(P,v)$ we have that $I\in \HT(\fgt_{\FW}(P,v))$.
Since $I$ does not contain $v$, then $I\htmodels \la v$.
We now show that $I\htmodels r$, for every $r\in P$.
 If $r\in P$ such that it does not contain $v$, then clearly $r\in \fgt(P,v)$, and so $I\htmodels r$.
 If $r\in P$ and it contains $v$, then we have the following cases:
 
\begin{itemize}
\item If $r$ is a tautology, then clearly $I\htmodels r$; 
 
\item If $v\in \rbody{r}$, then, since $I$ does not contain $v$, trivially $I\htmodels r$;

\item If $r=A,v\la B, \nf v, \nf C$ then the rule $r^*=A\la B, \nf C\in \fgt_{\FW}(P,v)$. Suppose $I\not\htmodels r$. Then we have two cases:

	\begin{itemize}
 	\item $Y\not\models r$, which is an absurd since $v\notin Y$ and $I\htmodels r^*$;
	
	\item $Y\cap C=\emptyset$ and $X\not\models A,v\la B$. This is an absurd since $v\notin X$,  ${r^*}^{Y}=A,v\la B$, and $X\models {r^*}^{Y}$.

	\end{itemize}
  
\item  If $r=A,v\la B, \nf C$ then $r^*=A\la B, \nf C\in \fgt_{\FW}(P,v)$. Suppose $I\not\htmodels r$. Then we have two cases:
  \begin{itemize}
 	\item $Y\not\models r$, which is an absurd since $v\notin Y$ and $I\htmodels r^*$;
	
	\item $Y\cap C=\emptyset$ and $X\not\models A,v\la B$. This is an absurd since $v\notin X$, ${r^*}^{Y}=A\la B$, and $X\models {r^*}^{Y}$.
 
\end{itemize}
 
\item If $r=A\la B, \nf v, \nf C$ then $r^*=A\la B, \nf C\in \fgt_{\FW}(P,v)$. Suppose $I\not\htmodels r$. Then we have two cases:

	\begin{itemize}
 	\item $Y\not\models r$, which is an absurd since $v\notin Y$ and $I\htmodels r^*$;
	
	\item $Y\cap C=\emptyset$ and $X\not\models A\la B$. This is an absurd since in this case ${r^*}^{Y}=A\la B$, and $X\models {r^*}^{Y}$.

	\end{itemize}
 \end{itemize} 

We now prove the right to left inclusion.
Let $I=\tuple{X,Y}\in \HT(P\cup\{\la v\})$. Since $I\htmodels \la v$ then clearly $I$ does not contain $v$.
We prove that for every $r\in \fgt_{\FW}(P,v)$ we have that $I\htmodels r$. This immediately implies that $\HT(P\cup\{\la v\})\subseteq \HT_{\parallel \{v\}}(\fgt(P,v))$.
Let $r\in \fgt_{\FW}(P,v)$. We consider the following cases.

\begin{itemize}
 
 \item $r\in \Cn(P,v)$. In this case, since $I\in \HT(P\cup\{\la v\})$, we have that $I\in \HT(P)$, and therefore $I\htmodels r$;
 
 \item If $r\notin \Cn(P,v)$ then we have the following three possibilities:

\begin{itemize}
 		\item $r=A\la B, \nf C$ such that  $r^*=A,v\la B, \nf v, \nf C\in \Cn(P,v)$. Suppose $I\not\htmodels r$. Then we have two cases:

	\begin{itemize}
 	\item $Y\not\models r$, which is an absurd since $v\notin Y$ and $I\htmodels r^*$;
	
	\item $Y\cap C=\emptyset$ and $X\not\models A\la B$. This is an absurd since $v\notin X$,  ${r^*}^{Y}=A,v\la B$, and $X\models {r^*}^{Y}$.

	\end{itemize}	
		
\end{itemize}
 
 \begin{itemize}
 		\item $r=A\la B, \nf C$ and $r^*=A,v\la B, \nf C\in \Cn(P,v)$. Suppose $I\not\htmodels r$. Then we have two cases:

	\begin{itemize}
 	\item $Y\not\models r$, which is an absurd since $v\notin Y$ and $I\htmodels r^*$;
	
	\item $Y\cap C=\emptyset$ and $X\not\models A\la B$. This is an absurd since $v\notin X$,  ${r^*}^{Y}=A,v\la B$, and $X\models {r^*}^{Y}$.

	\end{itemize}	
		
\end{itemize}

 \begin{itemize}
 		\item $r=A\la B, \nf C$ and $r^*=A\la B, \nf v, \nf C\in \Cn(P,v)$. Suppose $I\not\htmodels r$. Then we have two cases:

	\begin{itemize}
 	\item $Y\not\models r$, which is an absurd since $v\notin Y$ and $I\htmodels r^*$;
	
	\item $Y\cap C=\emptyset$ and $X\not\models A\la B$. This is an absurd since ${r^*}^{Y}=A\la B$, and $X\models {r^*}^{Y}$.

	\end{itemize}	
		
	\end{itemize}

\end{itemize}
\end{proof}

\begin{lemma}\label{cor:EquilityFW}
Let $P$ be a disjunctive program over a signature $\sign$ and $V\subseteq \sign$.
Then, for every $\fgt\in \classF_{\FW}$ we have  \[\HT_{\parallel V}(\fgt(P,V))=\HT(P\cup\{\la v: v\in V\}).\]
\end{lemma}
\begin{proof}
 The result follows easily by induction on the number of elements of $V$, using the definition of $\fgt_{\FW}(P,V)$ and Lemma~\ref{lemma:FWEqualityAS}.
\end{proof}

\begin{lemma}\label{lemma:FWEqualityAS}
 Let $P$ be a disjunctive program over a signature $\sign$ and $V\subseteq \sign$. Then, for every $\fgt\in \classF_{\FW}$
 \[\as{\fgt(P,V)}=\{X\in \as{P}: V\cap X=\emptyset\}.\]
 \end{lemma}
 
\begin{proof}
Let $\fgt\in \classF_{\FW}$. We will prove both inclusions. 
 
 Let us start with the left to right inclusion. Let $X\in \as{\fgt(P,V)}$. Then $\tuple{X,X}\in \HT(\fgt(P,V))$ and there is no $X'\subset X$ such that $\tuple{X',X}\in \HT(\fgt(P,V))$. We can conclude that $V\cap X=\emptyset$, since otherwise $\tuple{X\setminus V,X}\in \HT(\fgt(P,V))$, which contradicts $X\in \as{\fgt(P,V)}$.

Since $\tuple{X,X}\in \HT_{\parallel V}(\fgt(P,V))$, by Lemma~\ref{cor:EquilityFW},  we have that $\tuple{X,X}\in \HT(P\cup\{\la v:v\in V\})$, and therefore, $\tuple{X,X}\in \HT(P)$. We need to prove that there is no $X'\subset X$ such that $\tuple{X',X}\in \HT(P)$. Suppose there is. Then $\tuple{X',X}\in \HT(P\cup\{\la v:v\in V\})$$=\HT_{\parallel V}(\fgt(P,V))$, thus contradicting the fact that $X\in \as{\fgt(P,V)}$.
 
Now let us prove the right to left inclusion. Let $X\in \as{P}$ such that $V\cap X=\emptyset$. Then $\tuple{X,X}\in \HT(P)$ and there is no $X'\subset X$ such that $\tuple{X',X}\in \HT(P)$. Since $V\cap X=\emptyset$ we have $\tuple{X',X}\in \HT(P\cup\{\la v:v\in V\})=\HT_{\parallel V}(\fgt(P,V))$. We just need to prove that there is no $X'\subset X$ such that $\tuple{X',X}\in \HT(P\cup\{\la v: v\in V\})$. Suppose there is. Then, $\tuple{X',X}\in \HT(P)$, which contradicts $X\in \as{P}$.
 \end{proof}

\begin{lemma}\label{lemma:equivalenceOfVExtensions}
 Let $\sign$ be a signature and $V\subseteq\sign$. Let $I=\tuple{X,Y}$ an $HT$-interpretation over $\sign$ that does not contain any $v\in V$ and $I^*$ such that $I^*\sim_{V}I$.
 Then, for every formula $\varphi$ over $\sign$ not containing any $v\in V$, we have that $I\htmodels\varphi$ iff $I^*\htmodels\varphi$.
\end{lemma}
\begin{proof}
 We prove the result by induction on the structure of the formula $\varphi$.
For the base case suppose $\varphi$ is a propositional atom $p\in\sign$, thus $p\notin V$. Then, $I\htmodels p$ iff $p\in X$ iff $p\in X'$ with $X'\sim_V X$ (since $p\notin V$) iff  $I^*\htmodels p$.
 
For the induction step we only consider the case $\varphi=\varphi_1\supset\varphi_2$. The other cases are straightforward.
We have that $I\htmodels \varphi$ iff (i) $Y\htmodels \varphi_1\supset\varphi_2$ and (ii) $I\htmodels \varphi_2$ whenever $I\htmodels \varphi_1$. From a well known result from classical logic, we have that $Y\htmodels \varphi_1\supset\varphi_2$ iff $Y\cup V\htmodels \varphi_1\supset\varphi_2$ (since no $v\in V$ occurs in $\varphi_1$ nor in $\varphi_2$). The equivalence of condition (ii) with the correspondent one for $I^*$ follows easily using induction hypothesis. 
\end{proof}

\begin{lemma}\label{lemma:modelsOfNonVPrograms}
 Let $\sign$ be a signature and $V\subseteq\sign$. For every formula $\varphi$ over $\sign$ not containing any $v\in V$ we have that $\HT(\varphi)=(\HT_{\parallel V}(\varphi))_{\dagger V}$ 
\end{lemma}
\begin{proof}
Let us prove both inclusions. For the left to right inclusion, assume that $I\in \HT(\varphi)$. Then, $I_{\parallel V}\in \HT_{\parallel V}(\varphi)$. Therefore, $I\in (\HT_{\parallel V}(\varphi))_{\dagger V}$.

For the reverse inclusion, let $I \in (\HT_{\parallel V}(\varphi))_{\dagger V}$. Then, $I_{\parallel V} \in \HT_{\parallel V}(\varphi)$. Then, there exists $I^*\in \HT(\varphi)$ such that $I^*\sim_V I_{\parallel \{v\}}$. Using Lemma~\ref{lemma:equivalenceOfVExtensions} we have that $I_{\parallel V}\in \HT(\varphi)$. Since $I\sim_V I_{\parallel \{v\}}$,  we can conclude, again using Lemma~\ref{lemma:equivalenceOfVExtensions}, that $I\in \HT(\varphi)$.
\end{proof}

\begin{lemma}\label{lemma:intersectionVersusUnionOfVExtensions}
  Let $\sign$ be a signature, $v\in \sign$, and $P_1, P_2$ programs over $\sign$, such that $P_2$ does not contain $v$. Then, 
\[(\HT_{\parallel V}(P_1\cup P_2))_{\dagger V}= (\HT_{\parallel V}(P_1)\cap \HT_{\parallel V}(P_2))_{\dagger V}\]
\end{lemma}

\begin{proof}
We prove both inclusions. Let us start with the left to right inclusion. Let $I\in (\HT_{\parallel V}(P_1\cup P_2))_{\dagger V}$. Then, $I_{\parallel V}\in \HT_{\parallel V}(P_1\cup P_2)$. Therefore, there exists $I'\sim_V I_{\parallel V}$ such that $I'\in {\HT}(P_1\cup P_2)$. From this we can conclude that $I'\in \HT(P_1)$ and $I'\in \HT(P_2)$, and thus $I_{\parallel V}\in \HT_{\parallel V}(P_1)$ and $I_{\parallel V}\in (P_2)$. Therefore, we have that 
$I_{\parallel V}\in \HT_{\parallel V}(P_1)\cap \HT_{\parallel V}(P_2)$, thus $I \in (\HT_{\parallel V}(P_1)\cap \HT_{\parallel V}(P_2))_{\dagger V}$.

For the reverse inclusion consider that $I \in (\HT_{\parallel V}(P_1)\cap \HT_{\parallel V}(P_2))_{\dagger V}$. Then, $I_{\parallel V}\in \HT_{\parallel V}(P_1)\cap \HT_{\parallel V}(P_2)$, and thus $I_{\parallel V}\in \HT_{\parallel V}(P_1)$ and $I_{\parallel V}\in \HT_{\parallel V}(P_2)$. Therefore, there exists $I^*\sim_V I_{\parallel V}$ such that $I^*\in \HT(P_1)$.
Since $\HT(P_2)=(\HT_{\parallel V}(P_2))_{\dagger V}$ (by Lemma~\ref{lemma:modelsOfNonVPrograms} and since no $v\in V$ occurs in $P_2$), $I_{\parallel V}\in \HT_{\parallel V}(P_2)$ and since $I^*\sim_V I_{\parallel V}$, we can conclude that $I^*\in \HT(P_2)$. Therefore, $I^*\in \HT(P_1\cup P_2)$, and then $I_{\parallel V}\in \HT_{\parallel V}(P_1\cup P_2)$. We can then conclude that $I\in (\HT_{\parallel V}(P_1\cup P_2))_{\dagger V}$.
\end{proof}

\begin{lemma}\label{lemma:InclusionOfModels}
Let $\sign$ be a signature and $v\in \sign$. Let $B_1$, $B_2$, $C_1$, $C_2$, $\{a\}$ sets of atoms over $\sign\setminus \{v\}$.
Consider the rules 
\begin{itemize}
 \item[] $r_1=a\la B_1, v,\nf C_1$
\item[] $r_2=v\la B_2, \nf C_2$
 \item[] $r=a\la B_1, B_2, \nf C_1, \nf C_2$
\end{itemize}
Then,
 \[{\HT}(\{r_1\}\cup \{r_2\})\subseteq  {\HT}(\{r\})\]
\end{lemma}
\begin{proof}
 Let $I=\tuple{X,Y}\in {\HT}(\{r_1\}\cup \{r_2\})$. Then, we have 
  \begin{description}
\item[$i_1)$] \hspace{0.15cm}$Y\models a\la B_1, v, \nf C_1$, 
\item[$ii_1)$] $X\models a\la B_1, v$ whenever $C_1\cap Y=\emptyset$,
\item[$i_2)$]\hspace{0.08cm} $Y\models v\la B_2, \nf C_2$,
\item[$ii_2)$] $X\models v\la B_2$ whenever $C_2\cap Y=\emptyset$.
\end{description}

Our aim is to prove that $I \in {\HT}(\{r\})$, i.e., 
 \begin{description}
\item[$i)$]\hspace{0.07cm} $Y\models a\la B_1, B_2, \nf C_1, \nf C_2$ and 
\item[$ii)$] $X\models a\la B_1, B_2$ whenever $(C_1\cup C_2)\cap Y=\emptyset$.
\end{description}
 If $(C_1\cup C_2)\cap Y\neq\emptyset$ then $i)$ and $ii)$ trivially hold. So, suppose that $(C_1\cup C_2)\cap Y=\emptyset$. If $B_1\cup B_2\not \subseteq Y$, then again $i)$ and $ii)$ trivially hold. So assume that $B_1\cup B_2 \subseteq Y$. Then from $i_1)$ and $i_2)$ we can conclude that $Y\models a$, and so $i)$ holds. If $B_1\cup B_2\not \subseteq X$ then immediately $ii)$ holds. So assume that $B_1\cup B_2 \subseteq X$. Then $ii_1)$ and $ii_2)$ allow us to conclude that $X\models a$, and so $ii)$ holds.
 \end{proof}

\begin{lemma}\label{lemma:intersectionOfVExtensions}
  Let $\sign$ be a signature, $V\subseteq \sign$, and $P_1, P_2$ programs over $\sign$. Then,
 $(\HT_{\parallel V}(P_1))_{\dagger V}\cap (\HT_{\parallel V}(P_2))_{\dagger V}$ 
 $= (\HT_{\parallel V}(P_1)\cap \HT_{\parallel V}(P_2))_{\dagger V}$.
\end{lemma}

\begin{proof}
Let $I$ be an $HT$-interpretation over $\sign$. Consider the following sequence of equivalent sentences:
\begin{description}
 \item[] $I\in (\HT_{\parallel  V}(P_1))_{\dagger V}\cap (\HT_{\parallel  V}(P_2))_{\dagger V}$ iff 
\item[] $I \in (\HT_{\parallel  V}(P_i))_{\dagger V}$, for each $i\in \{1,2\}$ iff
\item[]  $I_{\parallel  V} \in \HT_{\parallel  V}(P_i)$ for each $i\in \{1,2\}$ iff
\item[] $I_{\parallel  V}\in (\HT_{\parallel  V}(P_1)\cap \HT_{\parallel  V}(P_2))$ iff
\item[] $I\in (\HT_{\parallel  V}(P_1)\cap \HT_{\parallel V}(P_2))_{\dagger V}$. 
\end{description}
\end{proof}


\label{lastpage}
\end{document}